\documentclass[11pt]{article}
\usepackage[T1]{fontenc}
\usepackage[utf8]{inputenc}
\usepackage[margin=1.25in]{geometry}
\usepackage{setspace}
\usepackage{amsmath,amssymb,amsthm,mathtools,bm}
\usepackage{microtype,graphicx,booktabs,array,enumitem,float,listings,xcolor,pgfplots}
\pgfplotsset{compat=1.18}

\providecommand{\halmos}{\hfill\ensuremath{\square}}
\usepackage[authoryear,round]{natbib}
\usepackage{authblk}

\usepackage[colorlinks=true,linkcolor=blue!70!black,citecolor=blue!70!black,urlcolor=blue!70!black]{hyperref}
\usepackage[nameinlink,capitalize,noabbrev]{cleveref}
\crefname{assumption}{Assumption}{Assumptions}

\newcommand{\rev}[1]{#1}
\newenvironment{revblock}{}{}
\definecolor{MSRevisionBlue}{RGB}{0,0,0}
\newfloat{algorithm}{tbp}{loa}\floatname{algorithm}{Algorithm}
\floatstyle{ruled}\restylefloat{algorithm}
\newenvironment{algorithmic}[1][]{\begin{list}{}{\leftmargin=1em\rightmargin=1em}}{\end{list}}
\newcommand{\STATE}{\item}\newcommand{\REQUIRE}{\item\textbf{Require. }}
\newcommand{\LOOP}{\item\textbf{Loop. }}
\newcommand{\ENDLOOP}{\item\textbf{End loop}}\newcommand{\FOR}[1]{\item\textbf{For #1}}
\newcommand{\ENDFOR}{\item\textbf{End for}}\newcommand{\IF}[1]{\item\textbf{If #1}}
\newcommand{\ENDIF}{\item\textbf{End if}}

\allowdisplaybreaks
\newtheorem{theorem}{Theorem}
\providecommand{\theHtheorem}{\arabic{theorem}}
\newtheorem{lemma}{Lemma}
\providecommand{\theHlemma}{\arabic{lemma}}
\newtheorem{proposition}{Proposition}
\providecommand{\theHproposition}{\arabic{proposition}}

\providecommand{\theHcorollary}{\arabic{corollary}}

\newtheorem{assumption}{Assumption}
\providecommand{\theHassumption}{\arabic{assumption}}
\newtheorem{definition}{Definition}
\providecommand{\theHdefinition}{\arabic{definition}}
\newtheorem{remark}{Remark}
\providecommand{\theHremark}{\arabic{remark}}

\providecommand{\theHexample}{\arabic{example}}

\providecommand{\argmax}{\operatorname*{arg\,max}}

\DeclareMathOperator{\E}{\mathbb{E}}
\DeclareMathOperator{\Var}{Var}

\newcommand{\Prob}{\mathbb{P}}
\newcommand{\Pbb}{\mathbb{P}} %
\DeclareMathOperator{\kl}{KL}

\DeclareMathOperator{\clip}{clip}

\newcommand{\N}{\mathbb{N}}
\newcommand{\cX}{\mathcal{X}}  %
\newcommand{\cD}{\mathcal{D}}  %

\newcommand{\1}{\mathbf{1}}

\title{\Large Designing Service Systems from Textual Evidence}
\author[1]{Ruicheng Ao\thanks{E-mail: \texttt{aorc@mit.edu}, \texttt{chenhy@mit.edu}, \texttt{siyangao@cityu.edu.hk}, \texttt{hanweili@cityu.edu.hk}, \texttt{dslevi@mit.edu}. Authors are listed in alphabetical order.}}
\author[1]{Hongyu Chen}
\author[2]{Siyang Gao}
\author[3]{Hanwei Li}
\author[1,4,5]{David Simchi-Levi}
\affil[1]{Institute for Data, Systems, and Society, Massachusetts Institute of Technology}
\affil[2]{School of Data Science, City University of Hong Kong}
\affil[3]{Department of Decision Analytics and Operations, City University of Hong Kong}
\affil[4]{Department of Civil and Environmental Engineering, Massachusetts Institute of Technology}
\affil[5]{Operations Research Center, Massachusetts Institute of Technology}
\date{November 4, 2026}
\hypersetup{pdftitle={Designing Service Systems from Textual Evidence},pdfauthor={Ruicheng Ao, Hongyu Chen, Siyang Gao, Hanwei Li, David Simchi-Levi}}
\begin{document}
\maketitle
\begin{abstract}
In service system design, it is often the case that the main evidence of system performance appears in textual form, such as customer support transcripts, complaint narratives, and compliance review reports. With the increasing use of large language models (LLMs), these textual trajectories can be converted into low-cost proxy scores that support comparison and optimization of candidate service configurations. However, LLM-based evaluations can exhibit systematic bias and require human audits for correction. To this end, this research studies how to identify the best service configuration with high confidence while minimizing the use of expensive human audits. We formulate the problem as fixed-confidence best arm identification \rev{with an always-observed, potentially biased proxy and selectively acquired human-verified outcomes. Moreover, several configurations can run on one case. A human audit can provide a case label that scores all their outputs. We show that sharing this reference and deciding whether to audit after observing proxy scores can each strictly reduce asymptotic identification cost. The selected configurations determine the available proxy scores and verified outcomes, so the proposed algorithm allocates configuration evaluations and subsequent audits jointly. For a specified finite family of data-generating models containing the true law and distinguishing different best configurations, its expected cost asymptotically attains the information-theoretic lower bound. For audits of individual outputs, we develop prediction-powered confidence sequences valid under adaptive auditing and delayed feedback. Numerical experiments on synthetic examples and service datasets show that the proposed algorithms require substantially less human review while maintaining reliable selection.}
\end{abstract}

\section{Introduction}\label{sec:intro}

In many service operations applications, managers often face a fundamental design problem of choosing among alternative system designs to optimize service quality and customer experience. Applications include call center routing and staffing, customer-support escalation policies, content-moderation review workflows, healthcare triage protocols, and compliance screening rules \citep{gans2003telephone, aksin2007modern, topol2019high, bates2003improving}. In each case, the manager must compare competing system configurations under uncertainty and choose the one that delivers the best service in practice.

Existing work on learning-based service system design and
performance optimization typically assumes access to random performance samples
that are available as computable scalars. In bandit-based and Bayesian optimization methods \citep{bubeck2012regret, frazier2018tutorial}, one iteratively evaluates configurations on the real system and collects scalar-valued samples, while in ranking and selection and search-based simulation optimization methods \citep{kim2001fully, chen2000, fu2002optimization}, one first builds a simulation model and then collects
scalar-valued samples from the simulator for different configurations. These methods are effective and have been widely studied, but their methodological foundations are built on performance samples that can be directly aggregated, compared, and optimized.

However, in many real service systems, the primary evidence of performance is not a readily computable scalar. Instead, it appears as textual trajectories, such as ticket descriptions and resolution logs, agent-customer transcripts, customer complaints, compliance review reports, or medical encounter notes. These records faithfully reflect what the system did and contain important information for evaluating the system performance. However, since they are unstructured text, they are difficult to incorporate into standard design and optimization algorithms. In practice, one can ask human experts to review and compare these textual trajectories, but this process is often highly inefficient. For example, a call center supervisor may be able to review only a few dozen calls per day out of the thousands the center handles, and a content moderation team may assess a few hundred posts out of the millions that may be generated over a comparable period. When a firm must evaluate hundreds of system configurations or more, possibly under a short time horizon, relying on human review of textual trajectories alone quickly becomes infeasible. As a result, service systems whose performance evidence is predominantly textual often cannot be effectively designed at large scale.

In the meantime, the wide adoption of large language models (LLMs) offers a potential solution to this gap. As automated readers of natural language, LLMs can process text that describes
service system performance and convert it into standardized scores. That is, they can serve as an
\emph{LLM-as-a-judge} that maps a text-based performance evaluation instance into a proxy scalar score. This mapping makes it possible to conduct service system design in settings where system performance is documented in language.

In this research, we study this setting. The manager must choose among a finite set of service configurations when the available performance samples are textual trajectories. Here, a configuration can include service operations design choices, such as staffing rules, queue discipline and priority weights, routing and transfer logic, etc. At the same time, since the system performance evaluation involves LLM-based scoring, a configuration can also include LLM-related choices, such as prompt templates, reasoning-depth settings, tool access, and the underlying language model.

\begin{revblock}
In this service system design problem, several configurations can be evaluated on one common instance. We call
this set a bundle, selected before drawing the instance from an offline pool.
Each configuration produces an output and a proxy score. An audit acquires
human-verified information, called a reference, such as a case label or an
output-specific judgment. The verified outcome is the configuration score
obtained from it. A case label, such as the correct intent of a request, can
evaluate all outputs on that case. The manager then pays for each output but
only once for the reference. The proxy scores indicate how useful an audit
will be before it is requested. The two decisions therefore need joint
consideration. When an audit evaluates only one response, we instead consider
one configuration at a time.
\end{revblock}

Although it is conceptually straightforward to use an LLM judge to convert text trajectories into scores and apply standard design algorithms to compare configurations and identify the best one, the use of LLMs raises new and nontrivial questions.

First, we cannot rely on the LLM judge alone. LLM judges make evaluations cheap and scalable, but their scores can exhibit systematic biases that vary across configurations and evaluation instances \citep{wang2023large, stureborg2024large, panickssery2024llm}. A judge scoring customer service quality may favor verbose responses over concise ones, irrespective of correctness. A compliance-oriented judge may systematically reward conservative phrasing. More importantly, these biases can be configuration dependent. The judge might over-rate certain routing behaviors (e.g., faster handoffs) even when they increase downstream workload, or under-rate concise triage notes that are clinically correct but stylistically atypical. When such biases differ across the configurations being compared, simply collecting more proxy scores does not resolve the problem, and can even amplify confidence in the wrong choice. It can be proved that the proxy scores alone \rev{are} not sufficient to distinguish between competing configurations, so that even with arbitrarily many LLM judge evaluations, a proxy-only procedure cannot be guaranteed to identify the true best configuration (\Cref{thm:mf-failure}, Part~1). To handle it, we still adopt human expert review, but use it in a selective-audit manner to address the issue described above. We treat audited expert outcomes as the verified outcome of interest, obtain a proxy score for every evaluated text trajectory, and selectively acquire human audits only for the scored cases that are necessary for valid correction and reliable selection.

Second, selective human audits require correction for sample selection. Under selective human audits, audited outcomes are not a random sample of all proxy scores, but are selected precisely from scores that look uncertain or otherwise salient. As a result, naive estimators that simply average audited outcomes, or that audit when the LLM is uncertain and then treat those audited outcomes as representative, can remain biased, even with infinitely many human audits, because the audit decision depends on samples that are predictive of the outcome (\Cref{thm:mf-failure}, Part~2). Therefore, the design challenge is not merely to allocate a fixed audit budget, but to develop principled rules that decide when to request human audits and how to combine audited and unaudited evaluations so that inference remains valid. Our approach addresses this by correcting selective auditing via an inverse propensity weighting (IPW) estimator that remains unbiased under adaptive audit decisions, and using these corrected estimates to drive a selection algorithm.

We formulate this service system design problem under the fixed-confidence best arm identification (BAI) \citep{even2006action, kaufmann2016complexity} framework with \(K\) service configurations (arms) under comparison\footnote{Hereafter, we use the terms ``arm'' and ``configuration'' interchangeably, and primarily use ``arm'' when presenting the BAI formulation and algorithms}. Each time we evaluate an arm on a text-based instance, we observe a proxy score \(F\in[0,1]\) at low cost (e.g., \$0.01 per API call), which summarizes the textual trajectory into a scalar signal but may be biased in ways that depend on both the configuration and the instance. We may additionally request a human audit \rev{that returns a verified outcome} \(Y\in[0,1]\) at higher cost (e.g., \$0.20 per review), \rev{which defines the performance outcome of interest.} Our goal is to identify the configuration with the best expected audited outcome with high confidence, while keeping the total cost small. To do so, we treat proxy scores as an always-available signal and use selective human audits to learn (and continuously monitor) how the proxy deviates from the verified outcome. The resulting procedure is designed to remain valid even when the proxy is systematically wrong for some configurations or for some segments of instances, and it aims to use audits \rev{primarily} where they are most informative for the final selection decision\rev{, while maintaining a positive routine review rate}.

Such a procedure can run many low-cost LLM judge evaluations, spend audits \rev{primarily} when they can change the decision, and stop once the evidence separates the leading configuration from the rest.

Our contributions are summarized as follows.

First, we study service system design using textual evidence and formulate the problem as fixed-confidence BAI with biased proxy scores and limited human audits. \rev{We show that sharing a reference and deciding whether to purchase it after observing proxy scores can each strictly reduce the characteristic cost, the leading coefficient in the asymptotic lower bound on expected cost. We also compare each design choice with the best allocation under its corresponding restriction and quantify the cost of that restriction. Reference sharing remains valuable even with independent verified outcomes and an uninformative proxy.}

\rev{Second, we develop a joint allocation algorithm for these decisions. Its allocation program counts a shared reference once and uses the information in all outcomes it verifies. For a specified finite family of candidate probability models containing the true law and distinguishing different best configurations, the algorithm learns which model generates observations and tracks its optimal allocation. Its expected cost attains the lower bound as the error probability decreases.}

\rev{Third, when each audit evaluates one configuration output, we} adopt a simple decomposition that expresses the mean of a configuration as the proxy score plus a residual gap between the judge and the verified outcome and develop a prediction-powered estimator that combines the proxy mean with an IPW residual correction, so the resulting mean estimates of the configurations remain unbiased \rev{at fixed per-arm pull counts}. We then construct anytime-valid confidence sequences for configuration means that are valid under adaptive sampling, adaptive auditing, and optional stopping, which enable the selection algorithm to stop as soon as the evidence separates the best configuration. \rev{We use these intervals to develop the PP-LUCB (Prediction-Powered Lower and Upper Confidence Bound) procedure, which combines PPI correction and LUCB-style comparisons, with Neyman allocation as an optional audit rule \citep{angelopoulos2023prediction,neyman1934two}. We also extend this confidence guarantee to the case of delayed audit feedback.}

\begin{revblock}
Last, we conduct numerical experiments on synthetic examples and service datasets, which evaluate the cost and reliability of the proposed methods in shared-reference and delayed-feedback settings. The results show that joint evaluation and selective auditing, such as auditing disagreement cases in support routing where one intent label evaluates multiple outputs, can substantially reduce human-review costs while maintaining reliable configuration selection.
\end{revblock}

\section{Related Literature}\label{sec:related}

Our work is related to four streams of research.

\paragraph{Service system design and quality evaluation.}
Service system design is a central topic in operations management, covering dynamic pricing \rev{\citep{besbes2009dynamic}}, staffing optimization \rev{\citep{gans2003telephone}}, routing and scheduling \citep{aksin2007modern}, etc. A critical component of design is quality evaluation. The operations literature has studied quality control systems that balance inspection costs against quality assurance \rev{\citep{juran1988quality}}. In call center operations, quality monitoring programs sample a small fraction of calls for supervisor review \citep{gans2003telephone,aksin2007modern}, and in healthcare, selective case review balances thoroughness against resource constraints \citep{bates2003improving}. Our work adds to this line of literature by studying a new and important setting in which the primary evidence of performance is textual, rather than directly observed as computable scalars.

\paragraph{Best arm identification and ranking and selection.}
BAI is a pure exploration bandit problem that identifies the arm with the best mean via samples \citep{even2006action,bubeck2011pure}. Closely related ranking and selection (R\&S) methods in statistics and simulation pursue the same goal, but typically collect samples from simulation models rather than real systems \citep{chen2000,gao2017}. Two standard BAI settings have been widely studied. In the fixed-budget setting, the learner is given a sampling budget and aims to minimize the misidentification probability at the end of the budget, while in the fixed-confidence setting, the learner aims to identify the best arm with probability at least $1-\delta$ while minimizing the number of samples. Multi-fidelity bandits extend these ideas by allowing cheap approximations to an expensive target, often under assumptions such as unbiasedness, known bias, or bounded deviations across fidelities \rev{\citep{kandasamy2016gaussian,ji2025bandits}}. In this research, we adopt the fixed-confidence setting \rev{and study allocation when several configurations can share a verified reference. When each audit evaluates one output, we} use LUCB-style comparisons together with a confidence-bound separation stopping rule to focus sampling on top arms \citep{jamieson2014lilucb,kalyanakrishnan2012pac,garivier2016optimal,kaufmann2016complexity}\rev{. Both settings allow} the proxy score \rev{to have} unknown, arm- and instance-dependent bias, \rev{with} the high-fidelity outcome \rev{requested} only through an adaptive auditing decision after observing the instance and proxy score. \rev{Multi-fidelity BAI is a closely related cost-sensitive identification problem \citep{poiani2024optimal}. In the present observation setting, the learner} always observes the low-fidelity proxy and only decides whether to request a high-fidelity human audit, and does not assume the proxy bias is known or controlled. \rev{Selective auditing therefore requires an appropriate correction and time-uniform inference. The information-cost formulation follows the established allocation perspective of \citet{qin2024optimizing}. Our joint program counts the reference-acquisition cost once and uses the conditional information in the outcomes that the reference verifies for a bundle of configurations.}

\paragraph{Prediction-powered inference.}
\rev{Prediction-powered inference (PPI) combines a small verified set with model predictions to construct valid confidence intervals \citep{angelopoulos2023prediction,angelopoulos2023ppi++}. Recent work further considers adaptive collection of verified outcomes and the joint design of sampling and measurement probabilities \citep{ao2024predictionguided}. We adapt these ideas to BAI by combining always-observed proxies with an inverse propensity-weighted residual correction. This correction treats selectively audited outcomes as missing at random (MAR) conditional on observables, producing unbiased estimation despite audit decisions that depend on observed information \citep{robins1994estimation}. Our setting additionally requires anytime-valid inference under adaptive sampling and optional stopping, which we obtain using confidence sequences and time-uniform concentration tools \citep{howard2021time,kaufmann2021mixture}. This combination corrects selective-audit bias while supporting valid early stopping. Concurrent work by \citet{landesberg2025cje} considers offline ranking under deployment shift using a calibrated surrogate and a small oracle sample.} \rev{For estimator-oriented auditing in our output-specific setting, we allow a clipped Neyman allocation rule, following the allocation principle in Section~7 of \citet{zrnic2024active}, to allocate audits for residual estimation. Related work studies stratified hybrid evaluation \citep{fisch2024stratified}, prediction-powered ranking of language models \citep{chatzi2024ranking}, and BAI with LLM judges and limited human audits \citep{ao2026best}. The latter is an earlier version of this research. The present paper additionally develops shared-reference joint allocation and establishes strict cost improvements when one verified label can evaluate multiple outputs.}

\paragraph{AI-assisted decision making.}
A growing literature studies how AI systems support human decisions, in domains including healthcare, lending, and customer service \rev{\citep{topol2019high,fuster2022predictably,huang2022mixed}}. In addition, LLM-as-a-judge has emerged as a scalable alternative to human evaluation for language systems \citep{zheng2023judging}, but empirical studies document systematic issues such as position bias, self-preference, sensitivity to prompting, and drift over time \rev{\citep{wang2023large,panickssery2024llm,stureborg2024large}}. More broadly, LLMs are increasingly applied to operations research tasks such as optimization formulation and solver interaction \rev{\citep{ao2026solver,liang2026autoformulation}}. Our work connects these themes in service operations by turning LLM-based evaluation into a sequential design problem and providing a statistically grounded way to use limited human audit for confident configuration selection. \rev{The observe-then-review setting also relates to human-AI review capacity \citep{lykouris2024defer}. Our target is to identify the best configuration with fixed confidence under a declared evaluation cost model. Selecting a bundle and then deciding whether to audit based on the observed state together form a controlled-sensing action \citep{nitinawarat2013controlled}, whose information and cost depend on the reusable reference.}

The remainder of the paper is summarized as follows. Section~\ref{sec:model} formulates the problem and identifies failure modes using existing methods. Sections~\ref{sec:estimation} and~\ref{sec:theory_lower_bound} develop the IPW estimator, derive information-theoretic lower bounds, and establish structural cost comparisons and asymptotic guarantees. Section~\ref{sec:algorithm} \rev{presents the joint allocation algorithm and the PP-LUCB procedure. The E-Companion extends the PP-LUCB confidence guarantee to delayed audit feedback.} Section~\ref{sec:experiments} evaluates \rev{the proposed algorithms} on synthetic examples and \rev{service datasets}. Section~\ref{sec:conclusion} concludes this paper and discusses some managerial implications.

\section{Model and Setup}
\label{sec:model}

Consider evaluating $K$ arms, indexed by
$k\in[K]\coloneqq\{1,\dots,K\}$. At each time step, a \emph{performance
evaluation instance} $X\in\cX$ is drawn from a distribution $\cD$. \rev{Each instance contains the query or service case to be evaluated, potentially accompanied
by structured metadata (task type, language, risk tier). The selected configuration
generates the textual trajectory used to evaluate system performance.} We refer to $X$ simply as an \emph{instance} hereafter. For every arm $k$ and instance $X=x$, there is a potentially available \textit{human audit outcome} $Y(k,x)\in[0,1]$, which is expensive but is
the objective we ultimately care about, and an always-available
\textit{proxy score} $F(k,x)\in[0,1]$, which is cheap and observed at
every evaluation but can be biased.

\begin{revblock}
The joint conditional kernel $P_k(df,dy\mid x)$ describes the distribution
of $(F(k,X),Y(k,X))$ given an instance $X=x$.
Deterministic scores are included as degenerate kernels. Expectations below
average over both this conditional distribution and $X\sim\cD$. Thus the
proxy and the verified outcome may remain dependent even for a fixed instance.
\end{revblock}

\rev{We allow judge bias to vary by arm and instance.}
\[
\E\!\left[F(k,X)\mid X=x\right]
= \E\!\left[Y(k,X)\mid X=x\right] + b_k(x).
\]
\rev{Here $b_k(x)$ is an unknown bias function that may vary among arms and
instances.} Define $\theta_k \coloneqq \E_{X\sim \cD}\big[\, Y(k,X)\,\big]$ as the
expected audited outcome for arm $k$, and let $k^\star \in
\argmax_{k\in[K]} \theta_k$ be the arm that achieves the maximal value.
Our goal is to identify $k^\star$ with high confidence while keeping the
\rev{total cost} small.
\rev{When the focus is exact identification, we will impose the condition of a unique best configuration.
In a more realistic setting when multiple configurations may tie, our experiments report a best design class or an
$\epsilon$-good configuration under a stated tolerance.}

\rev{In rounds $t=1,2,\dots$, the decision maker selects arms and observes
outcomes in the following order.}

\begin{enumerate}[leftmargin=*]
\item Select an arm $k_t\in[K]$ based on historical observations.

\item Draw an instance $X_t\sim \cD$ and observe the proxy score $F_t
\coloneqq F(k_t,X_t)$ at cost $c_F>0$.

\item Decide whether to request a human audit. $A_t\in\{0,1\}$ with
probability $\pi_t$, where if $A_t=1$, we observe $Y_t\coloneqq
Y(k_t,X_t)$ at additional cost $c_Y>0$.
\end{enumerate}

\rev{If context is observed before configurations are selected, performance can be estimated within each context class and then combined according to the target service mix. This extension is discussed in Section~\ref{subsec:context-first}.}

In the single-configuration setting, the audit decision uses the recorded
history, selected configuration, and current instance and proxy score.
Its probability is
\[
\pi_t \coloneqq \Pbb\big(A_t=1\mid \mathcal{H}_{t-1}, k_t, X_t, F_t\big).
\]
Here \rev{$\mathcal{H}_{t}:=\sigma(\{X_i,F_i,A_i,A_iY_i,k_i,\pi_i\}_{i=1}^t)$} is
the history up to time $t$ (with $Y_i$ observed only when $A_i=1$). Note
that the auditing decision cannot depend on the unobserved $Y_t$.

For a selection algorithm, a (random) stopping time $\tau$ and its
selected arm $\widehat{k}$ are called \emph{$\delta$-correct} if they satisfy
\[
\Pbb(\widehat{k}=k^\star)\ge 1-\delta.
\]
\rev{Our objective is to design a selection algorithm with stopping and cost
guarantees under the stated sampling and information conditions. We define
the cumulative cost as}
\[
\mathrm{Cost}(\tau) \coloneqq c_F \cdot \tau + c_Y \cdot
\sum_{t=1}^{\tau} A_t.
\]

We make the following technical assumptions throughout the paper. \rev{They specify the range of outcomes and the information available when making audit decisions.}

\begin{assumption}\label{ass:bounded}
For all $k,x$, $Y(k,x)\in[0,1]$ and $F(k,x)\in[0,1]$ almost surely.
\end{assumption}
\begin{revblock}
The unit interval is a natural scale for normalized success scores, intent
matches and bounded quality ratings. It also bounds the residual $Y-F$, so
the confidence boundaries and stopped-cost calculations have finite increments.
Other bounded scales can be rescaled to this interval.
\end{revblock}

\begin{revblock}
For the shared-reference setting, the selected set $B_t$ produces a pre-audit
state $G_t$ containing the instance, outputs, proxy scores and metadata.
The indicator $V_t$ records whether a reference is purchased, revealing the
verified scores $Y_{B_t,t}$ for the selected configurations. The following
assumption covers both this setting, formalized below, and evaluation of one
configuration at a time.

\begin{assumption}\label{ass:audit}
For a singleton evaluation, let $\mathcal H_t^S=\sigma(\mathcal H_{t-1},k_t,X_t,F_t)$ and $\pi_t=\Pbb(A_t=1\mid\mathcal H_t^S)$. For a bundle evaluation, let $\mathcal G_{t-1}$ denote the observed
shared-reference history through round $t-1$, and let
$\mathcal H_t^B=\sigma(\mathcal G_{t-1},B_t,G_t)$ and $
\rho_t=\Pbb(V_t=1\mid\mathcal H_t^B)$. The audit and purchase indicators satisfy
$A_t\perp Y_t\mid\mathcal H_t^S$ and $V_t\perp Y_{B_t,t}\mid\mathcal H_t^B$
respectively. Their logged propensities satisfy $p\leq\pi_t\leq1$,
$p\leq\rho_t\leq1$, and $p\in(0,1]$.  Thus each review decision uses only the available pre-audit information,
and the corresponding outcome is observed with a known positive probability.
\end{assumption}

The conditional independence assumptions require randomization based on the
specified pre-audit information sets. This permits
inverse-propensity correction. The floor enforces a small routine review rate,
as service teams commonly reserve capacity for monitoring even apparently easy
cases. If some regions are never audited, their unknown residual means can
prevent distribution-free identification. The positive floor avoids this problem.
\end{revblock}

With these assumptions, we next formalize why relying on
proxy scores alone can fail under arm-dependent bias and
naively using selectively audited outcomes can still lead to biased conclusions.

The true mean of each arm decomposes as
\begin{equation}\label{eq:pp-identity}
\theta_k = \E[Y\mid k] = \E[F\mid k] + \E[Y-F\mid k].
\end{equation}
The first term is observable from proxy scores alone, and the second term (the residual mean) requires human audits to estimate. Correct selection of the best arm requires learning both components, and we show that selecting the best arm is impossible given only the proxy scores.

We define an algorithm to be proxy-only if it does not audit at all.
\begin{definition}
An algorithm is \emph{proxy-only} if it never audits, i.e., $A_t\equiv 0$, and its decision $\widehat{k}$ depends only on $\{(k_t,X_t,F_t)\}_{t\le \tau}$.
\end{definition}

\begin{theorem}\label{thm:mf-failure}
Under \Cref{ass:bounded}, consider the model class where instances are drawn i.i.d.\ $X\sim\cD$, and for each arm $k$ the joint distribution of $(Y(k,X),F(k,X))$ is otherwise unrestricted (allowing arm- and instance-dependent bias).
\begin{enumerate}[leftmargin=*]
\item For any proxy-only algorithm producing a selected arm $\widehat{k}$ after any (possibly random) number of rounds, there exist two instances $\mathcal{I}$ and $\mathcal{I}'$ in the model class such that the distribution of all observations under $\mathcal{I}$ and $\mathcal{I}'$ is identical, but the best arm is different. Consequently, for at least one instance,
\[
\Pbb(\widehat{k}\neq k^\star)\ \ge\ \tfrac{1}{2}.
\]
\item \rev{For any integrable random variable $Z$, write $\E_k[Z]:=\E[Z\mid k]$
for expectation under the arm-$k$ law.  Consider the unweighted selective-audit mean
\[
\widehat\theta_k^{\rm nai}(n)=
\frac{\sum_{t\le n:k_t=k}A_tY_t}
     {\sum_{t\le n:k_t=k}A_t},
\]
whenever the denominator is positive. This estimator ignores the logged audit probabilities.
There exist an instance and an audit policy depending only on $F$ such that, on every
sample path for which $N_k(n)\to\infty$ and the audited count also diverges,
\[
\widehat\theta_k^{\rm nai}(n)
\xrightarrow{\rm a.s.}
 m_k:=\frac{\E_k[\pi(F)Y]}{\E_k[\pi(F)]}\ne\theta_k.
\]}
\end{enumerate}
\end{theorem}

\Cref{thm:mf-failure} establishes two key limitations that motivate our
approach. Part~1 shows that correctly identifying the best arm can be impossible using only the (possibly biased) proxy scores, i.e., human audits
are essential for reliable identification when bias differs among arms. Part~2 shows that auditing alone is not enough. If audits
are triggered selectively based on observed scores, then simple
estimators that ignore the audit-selection mechanism can remain biased,
even with many audits. This motivates the IPW correction developed in the next section.
\begin{revblock}

\subsection{The shared-reference observation setting}
The model above evaluates one configuration on each instance. We now
allow several configurations to run on a common instance. A
human-verified case label can then score all their outputs, so these
evaluations share one reference. Let $\mathcal B$ be a finite collection
of nonempty subsets of $[K]$.

Let $\mathcal G_{t-1}$ denote the observed shared-reference history
through round $t-1$. At round $t$, the learner selects a bundle
$B_t\in\mathcal B$ based on $\mathcal G_{t-1}$, before the current
instance is drawn and before any verified outcome is available. The
selected configurations are run on that common instance. Let $T_{k,t}$
denote the textual output generated by configuration $k$ and let
$F_{k,t}$ denote its proxy score. They produce the pre-audit state
\[
G_t=\left(X_t,\{T_{k,t},F_{k,t}:k\in B_t\},W_t\right).
\]
Here $W_t$ contains metadata observed before verification. After
observing $G_t$, the learner purchases a reference with probability
\[
\rho_t
=
\Prob(V_t=1\mid\mathcal G_{t-1},B_t,G_t)
\in[p,1].
\]
Here $V_t$ records the purchase. If $V_t=1$, the reference returns the
verified vector $Y_{B_t,t}=\bigl(Y_{k,t}:k\in B_t\bigr).$ The reference is charged once and updates the estimates for every
configuration in $B_t$. When $V_t=0$, the vector remains unobserved.

The post-round history $\mathcal G_t$ additionally contains $B_t$,
$G_t$, $\rho_t$, $V_t$, and $Y_{B_t,t}$ when it is returned, together
with any policy randomization used through round $t$. Thus the bundle is
selected before the current instance is observed, while the purchase
decision may depend on the observed pre-audit state but not on the
unobserved verified outcomes.

Under candidate probability model $\nu$, write $\mu_{\nu,B}$ for the law of
the pre-audit state on $\mathcal G_B$ and $Q_{\nu,B}(\cdot\mid g)$ for the
conditional law of the verified vector on $\mathcal Y_B$.
\begin{assumption}
\label{ass:fresh-sampling}
Under model $\nu$, conditional on the past and $B_t=B$, $(G_t,Y_{B,t})
\sim
\mu_{\nu,B}(dg)\,Q_{\nu,B}(dy\mid g)$, and this conditional law does not otherwise depend on the past. For every
configuration $k$, the mean
\[
\theta_{\nu,k}
:=
\int_{\mathcal G_B}
\int_{\mathcal Y_B}
y_k\,Q_{\nu,B}(dy\mid g)\,\mu_{\nu,B}(dg)
\]
is invariant across all $B\in\mathcal B$ containing $k$. Singleton
evaluation corresponds to $B=\{k\}$.
\end{assumption}

The bundle-invariance condition holds, for example, when the verified
reference is an instance-level label, such as an intent, priority, or outcome
category, that is defined independently of the evaluated configurations, and
the output law of each configuration is unchanged by the other configurations
run with it. If its verified score depends only on that output and the shared
reference, its target mean is the same across bundles. An output-specific audit gives
a verified outcome only for the audited output and thus corresponds to a
singleton bundle. A pairwise preference judgment applies only to its recorded
pair and cannot be used to score arbitrary other outputs.

The total realized cost of configuration evaluation and reference acquisition is
\begin{equation}
\operatorname{Cost}_{\mathcal B}(\tau)=
\sum_{t=1}^{\tau}\left[c^F(B_t,G_t)+V_t c^Y(B_t,G_t)\right].
\label{eq:bundle-realized-cost}
\end{equation}
The configuration evaluation cost includes model calls, simulation, and
output generation. The reference acquisition cost includes expert review
and comparison with the selected outputs. Both costs may depend on the
selected configurations and the pre-audit state.

\end{revblock}

\section{Prediction-Powered Estimation}\label{sec:estimation}

We develop an estimator that combines the always-available proxy score with a correction term learned from selectively audited outcomes, and construct an anytime-valid confidence sequence that supports valid stopping. \rev{We first consider one configuration per evaluation, and then extend the estimator to shared references.}

\subsection{IPW residual estimator}

\rev{Let $N_k(t)\coloneqq \sum_{s\le t}\1\{k_s=k\}$ be the number of times an arm $k$ is pulled through time $t$.}
The key identity \eqref{eq:pp-identity} suggests estimating $\theta_k$ as the sum of the proxy mean and the residual mean. Thus, we employ the inverse propensity weighting (IPW) estimator to correct for selective auditing. We define the estimators for $\E[F\mid k]$, $\E[Y-F\mid k]$ and $\theta_k$ as follows
\begin{align}
\widehat{\mu}_{F,k}(t)
&\coloneqq \frac{1}{N_k(t)} \sum_{s\le t,\;k_s=k} F_s,\\
\widehat{\mu}_{R,k}(t)
&\coloneqq \frac{1}{N_k(t)} \sum_{s\le t,\;k_s=k}
\frac{A_s}{\pi_s}\,(Y_s - F_s),\\
\widehat{\theta}_k(t)
&\coloneqq \widehat{\mu}_{F,k}(t) + \widehat{\mu}_{R,k}(t).
\end{align}

Here, since we observe the proxy $F$ every time we pull one
arm, its mean is estimated via a simple average $\hat\mu_{F, k}(t)$. The
residual part captures the systematic gap between the proxy score and
the verified outcome. Since $Y$ is observed only when we request an audit,
we estimate the residual mean via an inverse-propensity adjustment that
re-weights audited cases by $1/\pi_s$. This is the standard correction
used in quality monitoring and program evaluation when inspection
decisions are selective but their probabilities are known and logged.

\begin{revblock}
Note that audited outcomes can also improve the proxy used in later evaluations. Specifically, replacing
$F_t$ with a prediction fitted from earlier returned audits and clipped to
$[0,1]$ gives the augmented inverse-propensity estimator in
\eqref{eq:aipw-estimator}. To establish its validity, Lemma~\ref{lem:aipw-unbiased} shows that the  estimator is unbiased at fixed per-arm pull counts. To support sequential selection and stopping, Proposition~\ref{prop:aipw-cs} further gives a
confidence sequence for the full corrected score, allowing the fitted
prediction to change as new audits return.
\end{revblock}

\begin{revblock}
\subsection{The bundle estimator}
For a bundle round, let \(I_{k,t}=\mathbf 1\{k\in B_t\}\) and let
\(M_k(t)=\sum_{s\leq t}I_{k,s}\). The proxy mean for configuration \(k\)
uses every output generated by \(k\). If a common reference
is purchased, its residual can be used for each configuration in the bundle.
The resulting inverse propensity estimator is
\begin{align}
\widehat\mu^{\mathrm{bun}}_{F,k}(t)
&=\frac{1}{M_k(t)}\sum_{s\leq t}I_{k,s}F_{k,s},\\
\widehat\mu^{\mathrm{bun}}_{R,k}(t)
&=\frac{1}{M_k(t)}\sum_{s\leq t}I_{k,s}
 \frac{V_s}{\rho_s}(Y_{k,s}-F_{k,s}),\\
\widehat\theta^{\mathrm{bun}}_k(t)
&=\widehat\mu^{\mathrm{bun}}_{F,k}(t)+
\widehat\mu^{\mathrm{bun}}_{R,k}(t).
\label{eq:bundle-aipw}
\end{align}
The same purchase indicator appears for all \(k\in B_s\), so one reference
corrects all selected configurations. Under the shared-reference observation
setting and MAR condition, propensity weighting gives each summand the correct
conditional mean under selective auditing, even when audit decisions depend on the observed pre-audit information. Section~\ref{sec:theory_lower_bound}
will use a likelihood confidence set for specified candidate models, based on the returned vector and the dependence among its components. The corrected-score confidence sequence below instead uses bounded outcomes without requiring these models to be specified.

The estimator can also use a proxy learned from earlier audits. Let $m_{k,t}\in[0,1]$ be a
prediction fitted from past returned references, evaluated at the current
state and fixed before the purchase coin is drawn. For example,
$m_{k,t}=\clip(a_{k,t}+\lambda_{k,t}F_{k,t},0,1)$ allows the intercept and
proxy weight to be fitted from earlier returned records. Then
\begin{equation}
\widehat\theta^{\mathrm{aipw}}_k(t)=
\frac{1}{M_k(t)}\sum_{s\leq t}I_{k,s}
\left[m_{k,s}+\frac{V_s}{\rho_s}(Y_{k,s}-m_{k,s})\right].
\label{eq:bundle-aipw-learned}
\end{equation}
Using the learned prediction changes the variability of the residual, while
the corrected score retains the same conditional mean. The full corrected score lies in
$[1-1/p,1/p]$, which gives the online confidence sequence in
Proposition~\ref{prop:aipw-cs}. The single-configuration estimator is obtained when all
bundles are singletons.

When comparing two configurations that may still be best, we can consider
the difference between their residuals. Let
\[
D_{ij,t}= (Y_{i,t}-F_{i,t})-(Y_{j,t}-F_{j,t})
\]
when both configurations belong to the selected bundle. Conditional on a
pre-audit state \(g\), put
\[
v_{ij}(g)=\E[D_{ij,t}^{2}\mid G_t=g].
\]
This second moment includes the covariance between the two residuals.
Adding their separate variances therefore does not give this quantity. Under a fixed
average purchase rate, the leading variance of the contrast correction is
minimized by the clipped rule
\begin{equation}
\rho_{ij}^\star(g)=\clip\left(\sqrt{v_{ij}(g)/\lambda_{ij}},p,1\right),
\qquad \E[\rho_{ij}^\star(G)]=\rho.
\label{eq:bundle-contrast-audit}
\end{equation}
For $v_{ij}>0$ almost surely and $p<\rho<1$, a positive normalizer
exists. Section~\ref{subsec:neyman-derivation} gives the zero-moment and
endpoint cases. This rule reduces the variance of an estimated mean difference.
Identifying the best configuration in a finite model family instead requires
distinguishing models with different best configurations. The joint allocation
program in Section~\ref{sec:theory_lower_bound} will measure this information by
the conditional KL divergence of the verified vector.
\end{revblock}

\subsection{Anytime-valid confidence sequences}\label{sec:anytime-ci}
\begin{revblock}
We now return to evaluating one configuration at a time. The following
confidence sequences require bounded observations and known audit probabilities,
but no specified family of candidate probability models.
{\color{black}
Let $\{\rev{\mathcal{H}_t}\}_{t\ge 0}$ be the natural filtration of the
process, where $\rev{\mathcal{H}_t}$ contains all observed quantities up to
time $t$, including $(k_s,X_s,F_s,A_s, Y_s,\pi_s)$ for $s\le t$, with
$Y_s$ missing when $A_s=0$.
} Arm choices $k_{t+1}$ are allowed to be
$\mathcal{H}_t$-measurable. The next audit probability is chosen after its
pre-audit state is observed and is measurable with respect to the history
together with that state. It is not required to be $\mathcal{H}_t$-measurable.
We construct anytime-valid confidence
sequences (CS) for the proxy mean $\mu_{F,k}\coloneqq
\E[F\mid k]$ and the residual mean $\mu_{R,k}\coloneqq \E[Y-F\mid k]$,
and then combine them for $\theta_k=\mu_{F,k}+\mu_{R,k}$.

We use an explicit dyadic stitched confidence width.  The construction below is
derived from Ville's inequality and is valid from the first positive
intrinsic time.  For $v>0$, let $j(v)=0$ when $0<v\leq1$ and
$j(v)=\lceil\log_2 v\rceil$ otherwise, and put $U_0=1$ and
$U_j=2^j$ for $j\geq1$.  With $w_j=6/[\pi^2(j+1)^2]$, define
\begin{align} \label{eq:stitched-boundary}
\psi(v;\alpha)
\coloneqq \sqrt{2U_{j(v)}\log\!\left(\frac{2}{\alpha w_{j(v)}}\right)}.
\end{align}
Set $\psi(0;\alpha)=0$.  This function is nondecreasing in $v$ and gives
time-uniform control for martingales with variance proxy $v$, including the
first positive intrinsic-time block.  Its piecewise-constant form is more
conservative than a smooth asymptotic width, but supplies the finite-time
margin used by the stopping rule.
For each arm, allocate $\delta_{F,k}=\delta_{R,k}=\delta/(2K)$ to the proxy
and residual CS, respectively. The dyadic construction already allocates its
budget among both signs, intrinsic-time blocks, and all checked times.

\begin{proposition}\label{thm:cs-proxy}
Under \Cref{ass:bounded,ass:audit} and the singleton fresh-draw condition in \Cref{ass:fresh-sampling}, fix an arm $k$ and an
arm-wise proxy error budget $\delta_{F,k}\in(0,1)$ and
define the centered sum
$S_{F,k}(t) := \sum_{s\le t,\;k_s=k}(F_s - \mu_{F,k})$.
{\color{black}
Since $F_s\in[0,1]$ and \rev{each observation is a fresh draw under the selected arm}, the terms
$(F_s-\mu_{F,k})$ form a bounded martingale-difference sequence along
pulls of arm $k$. A valid variance proxy is $V_{F,k}(t)=N_k(t)/4$. Define the confidence half-width
}
\begin{align*}
w_{F,k}(t)\ &\coloneqq\ \frac{\psi(V_{F,k}(t), \delta_{F,k})}{N_k(t)}
\ =\ \frac{\psi(N_k(t)/4, \delta_{F,k})}{N_k(t)}.
\end{align*}
Then, with probability at least $1-\delta_{F,k}$, simultaneously for all
$t$ such that $N_k(t)\ge1$,
{\color{black}
\[
\mu_{F,k}\in \big[\widehat{\mu}_{F,k}(t)\pm w_{F,k}(t)\big].
\]}
\end{proposition}

Define the raw residual $R_t=Y_t-F_t$ and corrected increment $Z_t=(A_t/\pi_t)R_t$. MAR gives $\E[Z_t\mid\mathcal H_t^S]=\E[R_t\mid\mathcal H_t^S]$. Averaging over the fresh state makes $Z_t-\mu_{R,k}$ a martingale difference on arm-$k$ pulls. Its support has length $M=2/p$, giving the sub-Gaussian variance proxy used below.

\begin{proposition}\label{thm:cs-residual}
Under \Cref{ass:bounded,ass:audit} and the singleton fresh-draw condition in \Cref{ass:fresh-sampling}, fix an arm-wise residual
error budget $\delta_{R,k}\in(0,1)$. For times with $N_k(t)\ge1$, let
{\color{black}
$\widehat{\mu}_{R,k}(t)
\coloneqq \frac{1}{N_k(t)}\sum_{s\le t:k_s=k} Z_s$ and
$S_{R,k}(t) = N_k(t)(\widehat{\mu}_{R,k}(t) - \mu_{R,k})$.
}
{\color{black}
The centered increment $Z_t - \mu_{R,k}$ is supported on an interval of length \rev{$M \coloneqq 2/p$}, so the variance process
$V_{R,k}(t) = N_k(t) \cdot M^2/4$ is a valid sub-Gaussian \rev{variance proxy}.
} Define
\begin{align*}
w_{R,k}(t)\ \coloneqq\ \frac{\psi(V_{R,k}(t);\, \delta_{R,k})}{N_k(t)}
\ =\ \frac{\psi(N_k(t)\cdot M^2/4;\, \delta_{R,k})}{N_k(t)}.
\end{align*}
Then with probability at least $1-\delta_{R,k}$, simultaneously for all
$t$ such that $N_k(t)\ge1$,
{\color{black}
\[
\mu_{R,k}\in \big[\widehat{\mu}_{R,k}(t)\pm w_{R,k}(t)\big].
\]
}
\end{proposition}

The variance proxy in Proposition~\ref{thm:cs-residual} uses the audit floor to uniformly bound the corrected residual increments.
This conservative choice gives valid intervals for every allowed audit
policy and is sufficient for the stopping rule below.
\end{revblock}

Combining the proxy and residual CS produces an anytime-valid CS for the
true mean $\theta_k$. Define
\begin{align}\label{eq:theta-ci}
L_k(t) &\coloneqq \widehat{\theta}_k(t) - w_{F,k}(t) - w_{R,k}(t), \notag\\
U_k(t) &\coloneqq \widehat{\theta}_k(t) + w_{F,k}(t) + w_{R,k}(t).
\end{align}

\begin{revblock}
\begin{theorem}\label{thm:anytime-ci}
Under \Cref{ass:bounded,ass:audit} and the singleton fresh-draw condition in \Cref{ass:fresh-sampling}, set
$\delta_{F,k}=\delta_{R,k}=\delta/(2K)$ for every arm. Using
\Cref{thm:cs-proxy,thm:cs-residual} and a union bound, and evaluating the
intervals only after each arm has been pulled at least once as in
Algorithm~\ref{alg:pp-lucb}, we have
\[
\Pbb\Big(\forall t\ge 1\ \text{with }N_j(t)\ge1\ \forall j\in[K],\ \forall k\in[K]:\ \theta_k \in
[L_k(t),U_k(t)]\Big)\ \ge\ 1-\delta.
\]
\end{theorem}

Building on these simultaneous coverage guarantees, Lemma~\ref{lem:stopping-separation} gives a sufficient condition under which the confidence interval for the best configuration becomes disjoint from those of the other configurations, which enables the separation-based stopping rule.
\end{revblock}

\Cref{thm:anytime-ci} plays two roles in our setting. First, it supports valid early stopping, allowing the system to stop
as soon as the best arm is separated from the rest.
\begin{revblock}
Second, it allows adaptive sampling and selective auditing when decisions
follow the stated order. The next arm may depend on the past history, and the audit
decision may additionally depend on the current observed state and proxy.
\end{revblock}

\begin{revblock}
\section{Joint Information and Cost Design}
\label{sec:theory_lower_bound}

With a shared reference, the selected configurations determine which outcomes
one reference can verify, while the proxy scores indicate how informative it
will be. We therefore decide which configurations to evaluate and when to
audit them in one allocation problem. For a specified finite model family, its characteristic cost is the
leading coefficient in the asymptotic lower bound on expected cost.

\subsection{A finite model family}
To develop the joint allocation algorithm, we specify a finite model family
$\mathcal M_0=\{\nu_1,\ldots,\nu_M\}$. These \(M\) models are candidate probability laws for the evaluation data instead of the \(K\) configurations being compared. For example, a service simulator can provide the joint evaluation laws under a finite set of candidate parameter settings. The laws are specified before evaluation, and the true law is assumed to be one of them, but its identity is unknown. Evaluation then learns this identity while allocating configuration evaluations and reference purchases.

Under model
$\nu$, the pre-audit state
for bundle $B$ has law $\mu_{\nu,B}$ on a finite state space $\mathcal G_B$,
and the verified vector takes values in a finite alphabet $\mathcal Y_B$ with
conditional law $Q_{\nu,B}(\cdot\mid g)$. The
verified outcomes may be dependent, and their joint law records this dependence. We omit states and outcomes of zero probability from
these alphabets, use a common support for all candidate models, and assume that
the one-round log-likelihood ratios are bounded. These conditions are used only in the likelihood analysis for the finite
model family. Section EC 7.6 further shows that a bounded-outcome procedure using shared references can maintain reliable selection under model misspecification without using the candidate likelihoods.

\begin{assumption}\label{ass:finite-library-r1}
Every $\nu\in\mathcal M_0$ has a unique best configuration. For every pair
$\nu,\nu'\in\mathcal M_0$ with different best configurations, there is a bundle
$B\in\mathcal B$ such that
\begin{equation*}
\kl(\mu_{\nu,B}\Vert\mu_{\nu',B})
+p\sum_{g\in\mathcal G_B}\mu_{\nu,B}(g)
\kl\!\left(Q_{\nu,B}(\cdot\mid g)\middle\|Q_{\nu',B}(\cdot\mid g)\right)>0.
\end{equation*}
\end{assumption}
The first term $\kl(\mu_{\nu,B}\Vert\mu_{\nu',B})$ in the assumption measures information in the pre-audit state, observed whenever
a bundle is evaluated. The second term measures information in the verified vector,
weighted by the purchase floor because the reference is observed with
probability at least $p$. Thus every two models with different best
configurations can be distinguished by an admissible bundle. The action, purchase, stopping and randomization kernels are common measurable
functions of observed history, chosen before the next verified outcome is
revealed. This information requirement does not rely on the finite-model
condition.

For an alternative model $\nu_a$ whose best configuration is different from that
of $\nu$, define
\begin{align}
d_{B0}^{a}&=\kl\left(\mu_{\nu,B}\middle\|\mu_{\nu_a,B}\right),\\
h_{Bg}^{a}&=\kl\left(Q_{\nu,B}(\cdot\mid g)\middle\|
Q_{\nu_a,B}(\cdot\mid g)\right).
\label{eq:bundle-kl-terms}
\end{align}
The second term is the KL divergence for the returned vector. Replacing it by
a sum of configuration-level divergences can either overstate or understate
the information when its components are dependent.

The shared-reference setting includes single-configuration evaluation as its singleton case. If
$\mathcal B$ contains only singleton bundles, then $G_t$ has one proxy
component, $Y_{B_t,t}$ has one verified component, and the total cost reduces to
the cumulative cost in the single-configuration model. An output-specific audit has
this form. A pairwise preference judgment instead applies only to its recorded
comparison, as in the experiment on the MT-Bench dataset (in the E-Companion).

\subsection{The joint characteristic cost}
For $B\in\mathcal B$ and $g\in\mathcal G_B$, write
$c_B^F(g)=c^F(B,g)$ for the configuration-evaluation cost and
$c_{Bg}^Y=c^Y(B,g)$ for the reference-acquisition cost from
\eqref{eq:bundle-realized-cost}. Write
\begin{align*}
d_{B0}^{a}&=\kl(\mu_{\nu,B}\Vert\mu_{\nu_a,B}),\\
h_{Bg}^{a}&=\kl\left(Q_{\nu,B}(\cdot\mid g)\middle\|
Q_{\nu_a,B}(\cdot\mid g)\right).
\end{align*}
We allocate evaluations and audits to collect one unit of information against
each alternative. Let $q_B$ be the required amount of sampling of bundle $B$
and $x_{Bg}$ be that amount multiplied by the purchase probability after
state $g$. These continuous allocation variables do not need to be integer counts.
The associated purchase probability is $x_{Bg}/q_B$. Minimizing the resulting
cost gives the joint characteristic cost
\begin{equation}
\begin{aligned}
C_{\rm bun}^\star(\nu,\mathcal A)
=\min_{q,x}\quad&
\sum_{B\in\mathcal B}\sum_{g\in\mathcal G_B}
\mu_{\nu,B}(g)\bigl[c_B^F(g)q_B+c_{Bg}^Yx_{Bg}\bigr]\\
\text{s.t.}\quad&
\sum_{B\in\mathcal B}d_{B0}^{a}q_B+
\sum_{B\in\mathcal B}\sum_{g\in\mathcal G_B}
\mu_{\nu,B}(g)h_{Bg}^{a}x_{Bg}\geq1,
\quad a\in\mathcal A,\\
&q_B\geq0,\qquad pq_B\leq x_{Bg}\leq q_B.
\end{aligned}
\label{eq:bundle-characteristic-cost}
\end{equation}
The change of variables $x_{Bg}=q_B\rho_{Bg}$ makes the program linear. It
also allows the purchase probability to depend on the pre-audit state.
Each constraint requires one unit of information to distinguish an alternative.
For confidence level $\delta$, this requirement is multiplied by the
likelihood threshold.
\rev{To state the lower bound, fix a true model $\nu\in\mathcal M_0$ and a
finite set $\mathcal A\subseteq\mathcal M_0$ whose best configuration differs
from that of $\nu$. Let $C_{\rm bun}^\star(\nu,\mathcal A)$ denote the optimal
value of the linear program in \eqref{eq:bundle-characteristic-cost}, and
let $\operatorname{Cost}_{\mathcal B}(\tau)$ be the realized total cost in \eqref{eq:bundle-realized-cost}. Assume
\Cref{ass:bounded,ass:audit,ass:fresh-sampling,ass:finite-library-r1},
take $\delta\in(0,1/2)$, and suppose that configuration-evaluation and
reference-acquisition costs are
finite and nonnegative. The alternatives in $\mathcal A$ are exactly the
models that the procedure must separate before it can stop.}

\begin{theorem}\label{thm:bundle-information-cost}
Any policy that terminates almost surely and is uniformly $\delta$-correct under
the true model and every model in $\mathcal A$ satisfies
\begin{equation}
\E_\nu[\operatorname{Cost}_{\mathcal B}(\tau)]
\geq \operatorname{kl}(1-\delta,\delta)
 C_{\rm bun}^\star(\nu,\mathcal A).
\label{eq:bundle-lower-bound}
\end{equation}
\end{theorem}

\rev{The theorem bounds the expected cost of distinguishing all alternatives
with a different best configuration, accounting for configuration-evaluation
and reference-acquisition costs. Thus, the characteristic cost $C_{\rm bun}^\star(\nu,\mathcal A)$ summarizes the least cost required per unit of statistical evidence against these alternatives.}

To characterize how this cost depends on the observed pre-audit state, we next consider the dual of the linear program in \eqref{eq:bundle-characteristic-cost}. The dual formulation shows which observed states are most informative to audit and weights the information against each alternative according to its role in the identification constraints. For
$\eta=(\eta_a\mid a\in\mathcal A)\geq0$, define
\[
H_{Bg}(\eta)=\sum_{a\in\mathcal A}\eta_a h_{Bg}^a.
\]
Linear programming duality gives
\begin{equation}
\begin{aligned}
C_{\rm bun}^\star=\max_{\eta\geq0}\quad
&\sum_{a\in\mathcal A}\eta_a\\
\text{subject to}\quad
&\sum_g\mu_{\nu,B}(g)c_B^F(g)-\sum_a\eta_a d_{B0}^a\\
&\quad+\sum_g\mu_{\nu,B}(g)
\min_{r\in[p,1]}r\left(c_{Bg}^Y-H_{Bg}(\eta)\right)\geq0,
\quad B\in\mathcal B.
\end{aligned}
\label{eq:bundle-dual}
\end{equation}
At a state with positive sampling intensity, an optimal purchase probability
is $1$ when $H_{Bg}(\eta)>c_{Bg}^Y$ and is $p$ when the inequality is
reversed. A tie is resolved by primal feasibility. This rule directs audits toward
distinguishing alternatives with different best configurations, while
variance-based allocation targets estimation accuracy.

The LP for single-configuration evaluation restricts bundles to singletons.
Its reference-acquisition cost is paid for each audited instance. A larger
bundle obtains several verified outcomes from one acquired reference. It can therefore
change the characteristic cost even when its pre-audit state carries no information.

\subsection{Value of reference sharing and audit timing}
The next theorem compares shared references with the best separate allocation.
It also compares decisions made after observing the pre-audit state with the
best constant purchase probability. Here audit timing concerns which current
information is available for the decision. Each comparison identifies the cost of imposing its restriction.

\rev{We describe the two constructions before stating the result. For
the reference-sharing construction, take two configurations with true verified means
$(0.6,0.4)$ and no informative pre-audit observations. The action menu contains both
singletons and their joint bundle. Conditional on an action, verified
components are independent. One alternative changes only the first mean and
the other changes only the second mean, with conditional KL $h>0$ for each
change. An audit of one output informs one configuration. A shared reference
informs both at one reference-acquisition cost. If one configuration costs
$c_F$ and one reference costs $c_Y$, the two information requirements are
separate under singleton evaluation and shared under the joint bundle.}

\rev{For the purchase-timing construction, use one action with two equally likely states
and true conditional means $(0.6,0.4)$ in both states. Conditional verified
components are independent Bernoulli variables. The first alternative changes
only configuration one in state one to a Bernoulli mean $a<0.2$ satisfying
$\kl(0.6\Vert a)=10$. The second changes only configuration two in state two
to a Bernoulli mean $b>0.8$ satisfying $\kl(0.4\Vert b)=1$. The state law is
the same under all models, so observing the state alone cannot distinguish
them. Instead, the state determines which alternative a reference can
distinguish. We use purchase floor $p=0.1$. In the second construction, write
$c_B^F$ for the configuration-evaluation cost of the selected bundle.}

\begin{theorem}\label{thm:bundle-public-good}
Under the two constructions just described, the following statements
hold.
\begin{enumerate}[leftmargin=*]
\item The reference-sharing costs are
$C_{\rm bun}^\star=(2c_F+c_Y)/h$ and
$C_{\rm sing}^\star=2(c_F+c_Y)/h$. Hence the joint bundle is strictly better
for every $c_Y>0$.
\item In the purchase-timing construction, the feasible statewise choice
$q=2$, $\rho_1=0.1$ and $\rho_2=1$ has expected cost
$2c_B^F+1.1c_Y$. Every constant purchase probability has cost at least
$2c_B^F+2c_Y$, so the statewise policy is strictly better for every $c_Y>0$.
\end{enumerate}
\end{theorem}

The theorem shows why these service features change the BAI allocation
problem. In the first construction,
sharing saves $c_Y/h$, or a fraction $c_Y/[2(c_F+c_Y)]$ of singleton cost.
The fraction increases toward one half as references become expensive relative
to configuration evaluation, even with independent verified components and an uninformative
proxy. Thus reference reuse can reduce cost even without improving proxy quality.
In the second construction, observing the state saves at least $.9c_Y$ at the same
amount of configuration evaluation. The state shows which alternative an audit can distinguish, and the
purchase rule uses this information. Managers should therefore assess which outputs share a reference and which
pre-audit states make a review informative. Neither loss can be removed by
changing the allocation within its restriction. This motivates the joint allocation in this research.
The constructions show strict gains under explicit laws, and the program
measures their size for a specified service model and prices.

Cost-sensitive controlled sensing already provides asymptotically optimal
tests under nonuniform control costs \citep{nitinawarat2015cost}. Here one sensing action consists of a set of configurations and a rule for
auditing after the pre-audit state is observed. The joint program represents these
choices directly, without listing every possible state-dependent audit rule.
Its information term uses the joint law of the verified outcomes, including
their dependence. Its dual identifies states that distinguish models with different best
configurations. Theorem~\ref{thm:bundle-public-good}
then quantifies the cost of restricting either reference reuse or post-proxy
purchasing, and the joint allocation algorithm uses the resulting
allocation.

\subsection{The joint allocation algorithm}
For a finite family with $M\ge2$ models, let $\ell_t(\nu)$ be the
log likelihood of the observations collected through round $t$. Set
$\beta_\delta=\log((M-1)/\delta)$ and define
\begin{equation}
\mathcal C_t=\left\{\nu\in\mathcal M_0\mid
\max_{m\in\mathcal M_0}\ell_t(m)-\ell_t(\nu)\le\beta_\delta\right\}.
\label{eq:joint-likelihood-confidence}
\end{equation}
The procedure stops when all models in $\mathcal C_t$ have the same best
configuration. The likelihoods use both the pre-audit state and the verified vector.
Thus the stopping rule uses the information collected when evaluating
configurations and purchasing references.

For each model $\nu$, let
$\mathcal A_\nu=\{\nu'\in\mathcal M_0:k^\star(\nu')\ne k^\star(\nu)\}$.
We solve \eqref{eq:bundle-characteristic-cost} over the full alternative set $\mathcal A_\nu$ and
select one optimizer $(q^\star(\nu),x^\star(\nu))$ by a fixed deterministic
rule. Thus an alternative that leaves the confidence set is still included
in the allocation problem. Put
\begin{equation}
S_\nu=\sum_B q_B^\star(\nu),\qquad
w_B(\nu)=\frac{q_B^\star(\nu)}{S_\nu},\qquad
\rho_{Bg}(\nu)=\frac{x_{Bg}^\star(\nu)}{q_B^\star(\nu)},
\label{eq:bundle-tracker-target}
\end{equation}
where $\rho_{Bg}(\nu)=p$ when $q_B^\star(\nu)=0$. If all candidate models
have the same best configuration, that configuration is returned at time
zero and no allocation is needed.

Before evaluation, models with identical observation laws for every admissible
bundle are represented by one model. Assumption~\ref{ass:finite-library-r1}
ensures that such models have the same best configuration. We assume that
the true law belongs to this family and that
\begin{equation}
\underline c_F:=\min_{\nu\in\mathcal M_0,\,B\in\mathcal B}
\sum_g\mu_{\nu,B}(g)c_B^F(g)>0.
\label{eq:bundle-positive-execution}
\end{equation}
This condition says that running any selected bundle has a positive expected
configuration-evaluation cost. The reference-acquisition cost may be zero.
Both costs are finite
and nonnegative. Together with the finite state spaces, this condition makes
the relevant LP level sets bounded. An optimizer therefore exists, and its
uniqueness is not required.

At round $t$, we find the maximum-likelihood model from previous observations
and use its allocation in \eqref{eq:bundle-tracker-target} as the target. A largest-deficit
rule tracks its bundle proportions, with a deterministic exploration share
$\varepsilon_t=t^{-\gamma}$, $0<\gamma<1$. After the state is observed,
$\rho_{Bg}$ gives the probability of purchasing the reference. Algorithm~\ref{alg:joint-bundle}
gives the updates.

\begin{theorem}\label{thm:bundle-tracker}
Under \Cref{ass:bounded,ass:audit,ass:fresh-sampling,ass:finite-library-r1},
the finite-model conditions and \eqref{eq:bundle-positive-execution},
Algorithm~\ref{alg:joint-bundle} (provided below) is uniformly $\delta$-correct for every
$\delta\in(0,1/2)$ and has finite expected stopping time and cost.
If the candidate models have different best configurations, then, for every true
model $\nu$,
\begin{equation}
\lim_{\delta\downarrow0}
\frac{\E_\nu[\operatorname{Cost}_{\mathcal B}(\tau_\delta)]}
{\log(1/\delta)}
=C_{\rm bun}^\star(\nu,\mathcal A_\nu),
\qquad
\lim_{\delta\downarrow0}\frac{\E_\nu\tau_\delta}{\log(1/\delta)}=S_\nu.
\label{eq:bundle-attainment}
\end{equation}
\end{theorem}

Theorem~\ref{thm:bundle-tracker} shows that the lower bound in Theorem~\ref{thm:bundle-information-cost} can be attained under the stated finite-model conditions. The limit of the expected cost includes all configuration evaluations and acquired references, including those used to learn which candidate model is true. It also averages over incorrect selections, rather than conditioning on the confidence set retaining the true model. Thus the characteristic cost describes an attainable leading cost of reliable identification.

This optimality statement concerns the limit as $\delta$ decreases. At a fixed confidence level, learning the true model and exploring the available configurations can still have a noticeable cost. A simpler allocation can therefore be cheaper in a finite experiment, even when its characteristic cost is higher. Section~\ref{sec:structural-study} will examine this difference between the asymptotic result and finite-sample performance.

\end{revblock}

\section{Selection Algorithm}
\label{sec:algorithm}
\rev{We now give the joint allocation algorithm for configurations sharing a reference and PP-LUCB for output-specific audits. The former uses candidate probability laws, while the latter uses the confidence sequences in Section~\ref{sec:anytime-ci}.}

\begin{revblock}
\subsection{Joint Allocation Algorithm}
Solve the LP for each candidate model before evaluation. Let $J=|\mathcal B|$ and $W_B$ be the cumulative target evaluation count for bundle $B$. Each round updates this target using the current maximum-likelihood model and exploration share, then selects the bundle furthest below its target.

\begin{algorithm}[H]
\color{MSRevisionBlue}
\caption{Joint allocation algorithm}
\label{alg:joint-bundle}
{\color{MSRevisionBlue}\begin{algorithmic}[1]\small
\REQUIRE Finite bundle family $\mathcal B$, finite model family $\mathcal M_0$,
error probability $\delta$, purchase floor $p$, and $\gamma\in(0,1)$.
\STATE Merge models with identical observation laws. If all models have
the same best configuration, \textbf{return} that configuration.
\STATE For each $\nu$, solve \eqref{eq:bundle-characteristic-cost} over
$\mathcal A_\nu$ using a fixed optimizer selector. Form $w(\nu),\rho(\nu)$
from \eqref{eq:bundle-tracker-target}.
\STATE Set $\ell_0(\nu)=0$, $N_B=0$, $W_B=0$, and
$\beta_\delta=\log((M-1)/\delta)$ for the number $M$ of distinct models.
\FOR{$t=1,2,\ldots$}
  \STATE Select $\hat\nu_{t-1}\in\argmax_\nu\ell_{t-1}(\nu)$ with a fixed tie rule.
  \STATE Set $\varepsilon_t=t^{-\gamma}$ and, for every $B$, update
  $W_B\gets W_B+(1-\varepsilon_t)w_B(\hat\nu_{t-1})+\varepsilon_t/J$.
  \STATE Select $B_t\in\argmax_B\{W_B-N_B\}$ with a fixed tie rule.
  \STATE Run every configuration in $B_t$ and observe $G_t$.
  \STATE Set $\rho_t=\rho_{B_tG_t}(\hat\nu_{t-1})$ and draw
$V_t\sim\operatorname{Bernoulli}(\rho_t)$.
  \STATE If $V_t=1$, purchase one reference and observe $Y_{B_t,t}$.
  \STATE For every $\nu$, update
  $\ell_t(\nu)\gets\ell_{t-1}(\nu)+\log\mu_{\nu,B_t}(G_t)
  +V_t\log Q_{\nu,B_t}(Y_{B_t,t}\mid G_t)$,
  omitting the last term when $V_t=0$.
  \STATE Update $N_{B_t}\gets N_{B_t}+1$ and form $\mathcal C_t$
  from \eqref{eq:joint-likelihood-confidence}.
  \STATE If all models in $\mathcal C_t$ have one best configuration,
  \textbf{return} that configuration.
\ENDFOR
\end{algorithmic}}
\end{algorithm}

For the likelihood update, put $Z_t=\dagger$ if $V_t=0$ and
$Z_t=Y_{B_t,t}$ otherwise. The observed-record density is
\begin{equation}
\mu_{\nu,B_t}(G_t)
\left[(1-\rho_t)\mathbf 1\{Z_t=\dagger\}
+\rho_t Q_{\nu,B_t}(Z_t\mid G_t)\mathbf 1\{Z_t\ne\dagger\}\right].
\label{eq:joint-observed-density}
\end{equation}
The bundle and purchase kernels are common across models and cancel in a
likelihood ratio. Thus the current pre-audit state can determine the purchase
probability without invalidating the likelihood calculation. The update uses
the joint likelihood of all outcomes verified by a shared reference.

Exploration continues to sample every bundle, allowing the maximum-likelihood estimate to identify the true model and then track its LP solution. The proof of Theorem~\ref{thm:bundle-tracker} bounds this learning time and the time to stopping, including paths where the true model leaves the confidence set.

\end{revblock}

\begin{revblock}
\subsection{PP-LUCB with output-specific audits}
PP-LUCB uses bounded outcomes and logged audit probabilities without specifying candidate laws. A Neyman audit rule is optional and targets estimation variance. Restricting Joint to individual configurations instead retains its finite model family. For arm $k$, let
\begin{equation}
\widehat\theta_k(t)=
\frac{1}{N_k(t)}\sum_{s\le t:k_s=k}
\left[F_s+\frac{A_s}{\pi_s}(Y_s-F_s)\right].
\label{eq:singleton-ipw}
\end{equation}
Under conditions of Theorem \ref{thm:anytime-ci}, PP-LUCB
uses the confidence sequences in Section~\ref{sec:anytime-ci}. It selects a
leader $\hat k_t$ and a challenger and stops when the leader's lower confidence bound exceeds the upper confidence bounds of all other configurations, so that
\begin{equation}\label{eq:stop}
L_{\hat k_t}(t)>\max_{k\ne \hat k_t}U_k(t).
\end{equation}
On the simultaneous coverage event, this separation condition guarantees that the selected leader is the best configuration. Algorithm \ref{alg:pp-lucb} summarizes the resulting procedure.

\begin{algorithm}[H]
\color{MSRevisionBlue}
\caption{PP-LUCB procedure}
\label{alg:pp-lucb}
\begin{algorithmic}[1]\small
\REQUIRE Configurations $[K]$ with $K\ge2$, error probability $\delta$, purchase floor $p$.
\STATE Initialize $s\gets0$, \rev{$r\gets0$}, and $N_k,S^F_k,R_k\gets0$ for every $k$.
\STATE Pull every configuration once, advancing $s$ at each pull. After observing
$F_s$, set $\pi_s\gets p$, draw $A_s\sim\operatorname{Bernoulli}(\pi_s)$,
and observe $Y_s$ if $A_s=1$. Add $A_s(Y_s-F_s)/\pi_s$ to $R_k$,
increment $N_k$, and add $F_s$ to $S^F_k$.
\LOOP
  \STATE Form $\widehat\theta_k=(S^F_k+R_k)/N_k$ and $[L_k,U_k]$ from
  \eqref{eq:theta-ci}. \rev{Choose $\hat k\in\argmax_k\widehat\theta_k$ and
  $c\in\argmax_{k\ne\hat k}U_k$ with fixed tie rules.}
  \IF{$L_{\hat k}>\max_{k\ne\hat k}U_k$}
    \STATE \textbf{return} the leader $\hat k$.
  \ENDIF
  \STATE \rev{Set $r\gets r+1$ and form the ordered pull list $(\hat k,c)$.
  If $r$ is a perfect square, append every configuration to this list.}
  \FOR{each configuration $k$ in the pull list}
    \STATE Advance $s$, pull $k$, observe $F_s$, choose and log $\pi_s\in[p,1]$,
    draw $A_s\sim\operatorname{Bernoulli}(\pi_s)$, and observe $Y_s$ if $A_s=1$.
    \STATE Add $A_s(Y_s-F_s)/\pi_s$ to $R_k$, increment $N_k$, and add $F_s$ to $S^F_k$.
  \ENDFOR
\ENDLOOP
\end{algorithmic}
\end{algorithm}
\end{revblock}

\begin{revblock}
The additional pulls at square-numbered iterations provide sparse exploration.
They contribute $K\lfloor\sqrt r\rfloor$ pulls through $r$ iterations and
ensure that each configuration continues to receive observations if the
procedure has not stopped.
\end{revblock}

\begin{theorem}\label{thm:delta-correct}
Under \Cref{ass:bounded,ass:audit} \rev{and the singleton fresh-draw condition in \Cref{ass:fresh-sampling}}, \rev{suppose the best configuration is
unique. Algorithm~\ref{alg:pp-lucb} stops almost surely and satisfies
$\Pbb(\widehat k=k^\star)\ge1-\delta$. More generally, any separation rule
using the simultaneous intervals satisfies
$\Pbb(\tau<\infty,\widehat k\ne k^\star)\le\delta$.}
\end{theorem}

\begin{revblock}
Theorem~\ref{thm:delta-correct} has a direct operational interpretation. On the
simultaneous coverage event, if
$\Delta_k=\theta_{k^\star}-\theta_k$, the stopping inequality holds once
\[
2\bigl(w_{F,k^\star}(t)+w_{R,k^\star}(t)
+w_{F,k}(t)+w_{R,k}(t)\bigr)<\Delta_k
\quad\text{for every }k\ne k^\star.
\]
If the procedure continues, sparse exploration makes all estimates converge
and all interval widths tend to zero. The positive gap therefore gives finite stopping. The audit
floor keeps the residual increments bounded. For a fixed floor and pull
count, increasing the actual purchase probability can reduce estimation
variance, while the displayed conservative width remains the same. When a
baseline is learned online, the intervals for the full corrected score in
Proposition~\ref{prop:aipw-cs} give the corresponding guarantee.
\end{revblock}

\section{Numerical Experiments}
\label{sec:experiments}

\rev{In this section, we conduct an experiment on a finite-model example to evaluate joint allocation and an experiment on the BANKING77 dataset to examine selective auditing with shared references. We then summarize the additional experiments in the E-Companion.}

\begin{revblock}
\subsection{Joint allocation of evaluations and audits}
\label{sec:structural-study}
\label{subsec:joint-tracker-experiment}

The experiment on the finite-model example examines how sharing references and choosing audits after observing the state affect selection cost. We use three configurations and three candidate probability models. Under $\nu^0$, verified means are $(.60,.57,.54)$. Model $\nu^1$ changes the second mean to $.64$, and $\nu^2$ changes the third to $.62$, so every configuration can be best. Each instance has two equally likely pre-audit states, and all seven nonempty bundles are admissible. The proxy law is common across models, but the information provided by an audit depends on the state. For example, $\nu^2$ changes configuration 3's success probability from $.54$ to $.70$ in state 0 and leaves it unchanged in state 1. Thus an audit in state 0 can distinguish these models, while one in state 1 cannot. We evaluate independent outcomes and correlated outcomes that use a common uniform variable with probability $.6$ and independent uniforms otherwise. The two laws have the same marginals and full support, allowing us to examine the effect of dependence. Section~\ref{ec:new-library} gives the probabilities and settings, including the audit floor $p=.1$.

For each true model and outcome law, we use 1,000 paths at $\delta=.05$, configuration-evaluation cost 1 per configuration and reference-acquisition cost 10 per vector. Singleton restricts Joint to one configuration per round. Fixed audit selects its purchase probability before observing the state. Full bundle evaluates all configurations but chooses audits after observing the state. All share Joint's candidate models and likelihood stopping rule. Singleton differs from PP-LUCB, which uses prediction-powered confidence sequences. Section~\ref{ec:new-protocol} also evaluates Full audit and Oracle.

\begin{table}[t]
\color{MSRevisionBlue}
\centering\small
\caption{Selection costs, references and errors under $\nu^0$.}
\label{tab:new-main-cost}
\begin{tabular}{llrrr}\toprule
Outcome law & Procedure & Cost (SE) & References & Errors\\\midrule
Independent & Joint & 4585.1 (81.4) & 373.3 & 47\\
 & Singleton & 5144.6 (87.5) & 434.2 & 44\\
 & Fixed audit & 5311.7 (95.1) & 467.5 & 30\\
 & Full bundle & 4316.4 (94.0) & 279.4 & 48\\\midrule
Correlated & Joint & 2286.8 (47.0) & 146.9 & 45\\
 & Singleton & 5175.6 (90.3) & 436.8 & 49\\
 & Fixed audit & 2517.4 (49.3) & 192.2 & 38\\
 & Full bundle & 2244.9 (45.5) & 145.2 & 39\\\bottomrule
\end{tabular}
\end{table}

Table~\ref{tab:new-main-cost} reports mean costs with Monte Carlo standard errors, mean reference counts and errors out of 1,000 paths. Joint improves on Singleton and Fixed audit in both cost and references, supporting the value of sharing references and choosing audits after observing the state. Full bundle is cheaper at this risk level. Thus the asymptotic advantage does not guarantee lower cost at a fixed risk level. All paths stop before 20,000 rounds across the 36 settings reported in Section~\ref{ec:new-protocol}.

To examine the limit of the expected cost in Theorem~\ref{thm:bundle-tracker}, we vary $\delta$ under $\nu^0$ with independent outcomes, using 1,000 fresh paths per procedure and risk level. Figure~\ref{fig:main-confidence-cost} divides mean costs by $C^\star_{\rm joint}\log(2/\delta)$, where $C^\star_{\rm joint}=C^\star_{\rm bun}$ is the characteristic cost in Section~\ref{sec:theory_lower_bound}. Error bars are $1.96$ Monte Carlo standard errors. Joint costs less than Full bundle at smaller risk levels. Its normalized cost falls from $1.345$ at $\delta=.01$ to $1.192$ at $10^{-9}$, remaining above the asymptotic benchmark. Exact LP calculations in Section~\ref{ec:new-prices} separate structural savings from learning costs. At configuration-evaluation cost 1, as reference-acquisition cost rises from $.1$ to $100$, savings over single-configuration evaluation increase from $1.38\%$ to $19.96\%$ for independent outcomes and from $19.18\%$ to $57.44\%$ for correlated outcomes.

\begin{figure}[t]
\color{MSRevisionBlue}
\centering
\includegraphics[width=.80\linewidth]{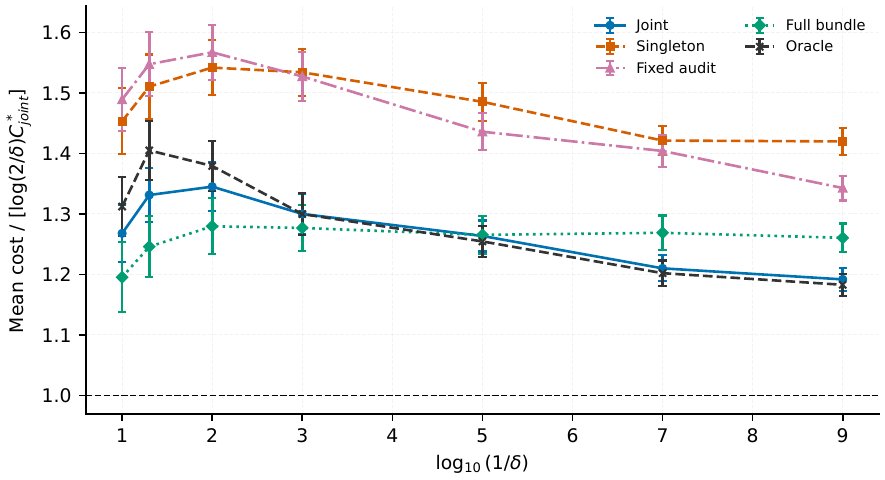}
\caption{Normalized selection costs at different confidence levels.}
\label{fig:main-confidence-cost}
\end{figure}

\subsection{Shared intent references in support routing}
\label{subsec:banking-bundle}

The next experiment is on the BANKING77 dataset \citep{casanueva2020intent}. It examines which queries need an intent label to identify an accurate configuration. We use 385 queries with five queries per intent. Each query has one published intent label. Four model and prompt configurations (Fable direct, Fable disambiguation, Astra direct and Astra disambiguation) generate predictions and proxy confidence scores without observing this label. Purchasing the label then evaluates all four predictions. Their accuracies are $85.97\%$, $86.75\%$, $94.81\%$ and $91.43\%$. Generating all 1,540 outputs costs $15.4$ normalized units, and each acquired label costs one unit. Every total includes both costs.

All methods use the same finite-population confidence rule at $\delta=.05$. A configuration is certified when the upper bound on every competitor's accuracy advantage is at most $\epsilon$. Thus success means selecting a configuration within $\epsilon$ of the best fixed-cohort accuracy.

We compare Uniform, Disagreement sampling and Adaptive sampling. Uniform samples from the full cohort. Disagreement sampling samples only the 47 queries with different predictions. The remaining 338 queries have zero accuracy difference between configurations for every possible label, so their labels are unnecessary for comparison. Adaptive sampling further groups queries using predictions and proxy categories. It selects a group according to the estimated reduction in confidence-interval widths for unresolved accuracy comparisons, then samples uniformly among its unrevealed queries. For each method and tolerance, we repeat acquisition 400 times with the cohort, predictions and proxies held fixed. Section~\ref{subsec:additional-acquisition} gives the confidence rule and additional cohort checks. This experiment compares label acquisition after output generation, while Section~\ref{sec:structural-study} evaluates the complete joint allocation algorithm.

Figure~\ref{fig:banking-acquisition} reports total costs of the compared sampling methods at the three prescribed tolerances. At $\epsilon=.01$, the mean label counts are $222.06$, $27.71$ and $24.00$, and total costs are $237.47$, $43.11$ and $39.40$. Disagreement sampling saves $81.84\%$ relative to Uniform, and Adaptive sampling saves a further $8.61\%$. Avoiding unanimous queries accounts for most of the savings, and adapting within the disagreement set adds a smaller reduction. All recorded paths stop with an $\epsilon$-good recommendation. At $\epsilon=.05$, both the best and second-best configurations satisfy this criterion because their accuracy gap is $3.38\%$.

\begin{figure}[t]
\color{MSRevisionBlue}
\centering
{\color{black}
\begin{tikzpicture}
\definecolor{plotBlue}{HTML}{0072B2}
\definecolor{plotOrange}{HTML}{D55E00}
\definecolor{plotGreen}{HTML}{009E73}
\begin{axis}[
font=\footnotesize,
axis line style={black},tick style={black},
tick label style={text=black},label style={text=black},
title style={text=black,font=\footnotesize},
legend style={font=\footnotesize,text=black,draw=none,fill=none},
grid=major,grid style={black!12},
scaled ticks=false,
clip marker paths=true
,
width=.78\linewidth,height=4.6cm,
xlabel={$\epsilon$},ylabel={Mean total cost},
xmin=.008,xmax=.052,ymin=0,ymax=260,
xtick={.01,.02,.05},xticklabels={.01,.02,.05},
ytick={0,50,100,150,200,250},
legend pos=north east,
legend cell align=left
]
\addplot[color=plotBlue,mark=*,solid,line width=.85pt,mark size=2.3pt] coordinates {(0.01,237.47) (0.02,202.91) (0.05,114.89)};
\addlegendentry{Uniform}
\addplot[color=plotOrange,mark=square*,dashed,line width=.85pt,mark size=2.3pt] coordinates {(0.01,43.11) (0.02,38.69) (0.05,27.67)};
\addlegendentry{Disagreement sampling}
\addplot[color=plotGreen,mark=diamond*,dashdotted,line width=.85pt,mark size=2.3pt] coordinates {(0.01,39.40) (0.02,36.40) (0.05,26.40)};
\addlegendentry{Adaptive sampling}
\end{axis}
\end{tikzpicture}
}
\caption{Mean selection costs on the BANKING77 dataset.}
\label{fig:banking-acquisition}
\end{figure}

\paragraph{Additional experiments.}
Section~\ref{ec:new-horizons} in the E-Companion examines the performance of the joint allocation procedure over finite evaluation horizons. As the horizon increases, Joint's recommendation regret declines and more paths satisfy the stopping criterion. In a separate four-configuration experiment, all 1,000 PP-LUCB paths retain coverage but none stops by 2,000 rounds. Section~\ref{ec:new-scale} examines larger configuration sets, where all 300 Joint paths stop at $K=128$. Section EC.7.6 examines how model misspecification affects Joint's selection reliability and evaluates a separate bounded-outcome selection rule under full auditing. The experiment on the MT-Bench dataset in Section~\ref{subsec:private-preference-details} uses pairwise human judgments, with Proxy lookahead reducing cost by $3.06\%$ relative to Labels-only lookahead. The experiment on the review queue in Section~\ref{EC-sec:queue-service} shows how congestion reduces the evidence available for selection by delaying audit completion. Sections~\ref{subsec:service-fixture-details} and~\ref{app:convergence-details} illustrate service-score construction and scoring rules for mathematical answers when references are known.
\end{revblock}

\section{Conclusions and Discussion}
\label{sec:conclusion}

Designing service systems often requires choosing among competing service
configurations when the main evidence of performance is textual. \rev{LLMs can turn
these records into low-cost proxy scores, but their judgments can be biased
across configurations and instances. Human review supplies the target
measurement, at greater cost and with possible delay. This paper combines the
two sources of evidence with a prediction-powered estimator, anytime-valid
confidence sequences and joint allocation of evaluations and audits. The resulting framework supports fixed-confidence identification of the best configuration despite biased proxies, selectively observed verified outcomes, and potentially shared references across configurations.}

\begin{revblock}
Our analysis points to several practical concerns when using textual evidence
in service system design.

The first practical point concerns the unit of information. A reference that
describes a common case can verify several configurations' outputs. If these
outputs are evaluated separately, the same reference may be purchased repeatedly.
Theorem~\ref{thm:bundle-public-good} shows that this restriction can increase cost
even under the best separate allocation. In its construction, the loss grows as
references become more expensive. Managers should therefore determine which
outputs can use the same reference before choosing an evaluation plan.
This compatibility is essential. An output-specific label cannot verify other
outputs, and a pairwise preference applies only to its recorded comparison.
Reference reuse reduces cost only when it preserves the target measurement.

A second practical point concerns the pre-audit state. Even a proxy that cannot
identify the best configuration by itself can indicate which cases are worth
auditing. The second construction in Theorem~\ref{thm:bundle-public-good} shows
that using this information can reduce cost relative to the best constant audit
probability. However, selective review also changes which outcomes are observed.
Using these outcomes without correcting for selection can produce biased
estimates and a confident but wrong choice. Managers should therefore combine
targeted review with a positive audit probability for every eligible case,
record that probability, and use it in the corrected score. The allocation
decision and the correction must follow the same audit rule.

A third practical point concerns deployment. A valid selection rule does not
ensure that enough reviews return before a service deadline. Our analysis of
delayed feedback preserves the PP-LUCB confidence guarantee, but audits that are still pending cannot yet be used to correct the estimates. The experiment on the review queue shows why this
distinction is important. Increasing audit submissions can create a backlog and delay
the evidence needed for stopping. Managers should therefore assess completion
time alongside statistical validity and choose review staffing and submission
rates together. A required confidence level becomes operationally useful only
when the review system can supply the evidence in time.

\end{revblock}

\nocite{kahneman2011thinking,aksin2007modern,angelopoulos2023prediction,angelopoulos2023ppi++,ao2026best,bates2003improving,bubeck2012regret,bubeck2011pure,casanueva2020intent,chatzi2024ranking,chen2000,even2006action,fisch2024stratified,frazier2018tutorial,gans2003telephone,garivier2016optimal,howard2021time,huang2022mixed,jamieson2014lilucb,ji2025bandits,kalyanakrishnan2012pac,kandasamy2016gaussian,kaufmann2016complexity,kaufmann2021mixture,kim2001fully,lai1976confidence,landesberg2025cje,lykouris2024defer,neyman1934two,nitinawarat2013controlled,panickssery2024llm,robins1994estimation,wang2023large,zheng2023judging}
\bibliographystyle{plainnat}
\bibliography{refs}

\clearpage
\appendix
\setcounter{section}{0}
\renewcommand{\thesection}{EC.\arabic{section}}
\renewcommand{\theHsection}{EC.\arabic{section}}
\setcounter{equation}{0}
\renewcommand{\theequation}{EC.\arabic{equation}}
\renewcommand{\theHequation}{EC.\arabic{equation}}
\setcounter{figure}{0}
\renewcommand{\thefigure}{EC.\arabic{figure}}
\renewcommand{\theHfigure}{EC.\arabic{figure}}
\setcounter{table}{0}
\renewcommand{\thetable}{EC.\arabic{table}}
\renewcommand{\theHtable}{EC.\arabic{table}}
\setcounter{theorem}{0}
\renewcommand{\thetheorem}{EC.\arabic{theorem}}
\renewcommand{\theHtheorem}{EC.\arabic{theorem}}
\setcounter{lemma}{0}
\renewcommand{\thelemma}{EC.\arabic{lemma}}
\renewcommand{\theHlemma}{EC.\arabic{lemma}}
\setcounter{proposition}{0}
\renewcommand{\theproposition}{EC.\arabic{proposition}}
\renewcommand{\theHproposition}{EC.\arabic{proposition}}
\setcounter{corollary}{0}
\renewcommand{\thecorollary}{EC.\arabic{corollary}}
\renewcommand{\theHcorollary}{EC.\arabic{corollary}}
\setcounter{assumption}{0}
\renewcommand{\theassumption}{EC.\arabic{assumption}}
\renewcommand{\theHassumption}{EC.\arabic{assumption}}
\setcounter{definition}{0}
\renewcommand{\thedefinition}{EC.\arabic{definition}}
\renewcommand{\theHdefinition}{EC.\arabic{definition}}
\setcounter{remark}{0}
\renewcommand{\theremark}{EC.\arabic{remark}}
\renewcommand{\theHremark}{EC.\arabic{remark}}
\setcounter{example}{0}
\renewcommand{\theexample}{EC.\arabic{example}}
\renewcommand{\theHexample}{EC.\arabic{example}}
\section*{Electronic Companion for ``Designing Service Systems from Textual Evidence''}

This E-Companion provides supporting  proofs, extensions, and additional numerical details for the main paper. It first consolidates the notation and modeling conditions used throughout the theoretical results, and then presents proofs of the main theorems, extensions for delayed audit feedback and contextual decisions, and complete details for the synthetic and service-system experiments.

\begin{revblock}
\section{Notation and Assumption Justification}

This section summarizes the observation sequences, histories, and model-family notation used throughout the E-Companion. For a single-configuration evaluation, the decision maker selects $k_t$, observes $(X_t,F_t)$, and then draws audit indicator $A_t$ with logged probability $\pi_t$. For shared-reference evaluation, the decision maker selects a bundle $B_t\in\mathcal B$, observes the pre-audit state $G_t$, and then draws the reference-purchase indicator $V_t$ with probability $\rho_t$. The acquired reference, such as a case label, gives verified scores $Y_{B_t,t}$ for all applicable outputs at one reference-acquisition cost. Audit probabilities lie in $[p,1]$. Given the state, each audit indicator is independent of the verified outcome. Histories $\mathcal H_t$ and $\mathcal G_t$ record observations and randomization through round $t$.

The finite family $\mathcal M_0$ specifies candidate probability models, with state law $\mu_{\nu,B}$ and conditional verified-vector law $Q_{\nu,B}(\cdot\mid g)$. Assumption~\ref{ass:fresh-sampling} keeps each configuration mean unchanged across admissible bundles. These models have common support, include the true law and distinguish different best configurations. Theorem~\ref{thm:bundle-tracker} also requires positive expected configuration-evaluation costs. We write
$c_B^F(g)=c^F(B,g)$ and $c_{Bg}^Y=c^Y(B,g)$ for the configuration-evaluation and reference-acquisition costs. Section~EC.5 defines the history of returned feedback.
\end{revblock}

\subsection{\rev{Examples of bounded audited outcomes}}
\label{subsec:bounded-outcome-examples}
\rev{This subsection gives illustrative outcome models that satisfy the bounded-outcome condition in Assumption 1. They show how verified service-quality outcomes can be represented on the unit interval when outcomes are binary, continuous ratings, or latent continuous scores that are clipped to the feasible range. These examples are relevant to the prediction-powered estimators and confidence sequences, which require bounded outcomes but do not require a fully specified parametric model. In contrast, the likelihood-based joint allocation guarantee in Section~\ref{sec:theory_lower_bound} requires a prespecified finite family of joint observation models on finite state and outcome spaces.}

\paragraph{Bernoulli logistic regression.}
Let $Y\in\{0,1\}$ and suppose that
conditional on $(k,X=x,F=f)$,
\[
\Pbb(Y=1\mid k,X=x,F=f)=\sigma(\beta_k^\top\phi(x,f)),
\qquad \sigma(u)\coloneqq\frac{1}{1+e^{-u}},
\]
for a known feature map $\phi:\cX\times[0,1]\to\mathbb R^d$ and arm parameter $\beta_k\in\mathbb R^d$. This model generates
bounded outcomes automatically and is a regular exponential family in canonical form.

\paragraph{Beta regression.}
Let $Y\in(0,1)$ and suppose that conditional on $(k,X=x,F=f)$,
\[
Y\mid(k,X=x,F=f)\sim\operatorname{Beta}(\alpha_k(x,f),\beta_k(x,f)).
\]
A common parameterization is via a conditional mean $m_k(x,f)\in(0,1)$ and precision $\varphi_k(x,f)>0$,
with $\alpha_k(x,f)=m_k(x,f)\varphi_k(x,f)$ and $\beta_k(x,f)=(1-m_k(x,f))\varphi_k(x,f)$. For example, one may
model $m_k(x,f)=\sigma(\beta_k^\top\phi(x,f))$ and let $\varphi_k(x,f)$ be either fixed or smoothly parameterized.
\rev{The symbols $\alpha_k(x,f)$ and $\beta_k(x,f)$ in this example are the two beta shape parameters. The symbol $\beta_k$ without arguments is the regression coefficient vector.}
The Beta density is
\[
p(y;\alpha,\beta)=\frac{\Gamma(\alpha+\beta)}{\Gamma(\alpha)\Gamma(\beta)}
y^{\alpha-1}(1-y)^{\beta-1},\qquad y\in(0,1).
\]

\paragraph{\rev{Clipped Gaussian.}}
Let $Z\mid(k,X=x,F=f)\sim N(\mu_k(x,f),\sigma_k^2)$
with $\sigma_k$ bounded away from 0, and define the bounded outcome by clipping\rev{,}
\[
Y\coloneqq\clip(Z,0,1).
\]
Equivalently, $Y$ has a mixed distribution consisting of point masses at 0 and 1 capturing the events
$\{Z\le0\}$ and $\{Z\ge1\}$, and a continuous density on $(0,1)$ equal to the Gaussian density of $Z$ restricted
to $(0,1)$. This construction satisfies Assumption~\ref{ass:bounded} by definition.

One convenient way to formalize this is to work with a dominating measure $m$ on $[0,1]$ that
is the sum of Lebesgue measure on $(0,1)$ plus \rev{unit point masses at} $\{0,1\}$. The likelihood $p_\mu(y)$ with
respect to $m$ is
\[
p_\mu(y)=
\begin{cases}
\Phi\!\left(\dfrac{0-\mu}{\sigma}\right),&y=0,\\[3pt]
\dfrac{1}{\sigma}\varphi\!\left(\dfrac{y-\mu}{\sigma}\right),&y\in(0,1),\\[3pt]
1-\Phi\!\left(\dfrac{1-\mu}{\sigma}\right),&y=1,
\end{cases}
\]
where $\varphi$ and $\Phi$ are the standard normal pdf and cdf.
\rev{Clipping retains endpoint masses, while conditioning on $(0,1)$ removes them and renormalizes the interior density. These examples concern bounded outcomes. The joint allocation guarantee instead uses finitely many specified models on finite observation spaces.}

\section{Proofs of Theoretical Results in Sections 3, 4, and 6}
\rev{This section provides proofs for the theoretical results stated in Sections 3, 4, and 6 of the main paper. It also includes the auxiliary results needed for these proofs.}

\subsection{Proof of Theorem~\textcolor{.}{1}}

\begin{proof}

\textit{Proof of Part 1.} Consider a setting with $K=2$ arms. We construct two problem instances where the marginal
distribution of proxy scores $F$ is identical for all arms, but the true means $\theta$ (and thus the better arm)
are different. All variables are bounded in $[0,1]$. The main text allows the algorithm to observe
$(X_t,F_t)$ in each pull. The construction below remains valid by taking the instance distribution $\cD$
to be degenerate, e.g., $X\equiv x_0$ almost surely. Then $X_t$ carries no information and the observable
history reduces to $\{(k_t,F_t)\}_{t\le\tau}$ without loss of generality.

\paragraph{Instance A (Arm 1 is better).}
For Arm 1, let $F\sim\operatorname{Bernoulli}(0.5)$ with $Y$ being deterministic given
$F$\rev{.} $Y=0.2$ if $F=0$, and $Y=1.0$ if $F=1$. The marginal mean is $\theta_1=0.5(0.2)+0.5(1.0)=0.6$, and
the marginal distribution of $F$ is $\operatorname{Bernoulli}(0.5)$.

For Arm 2, let $F\sim\operatorname{Bernoulli}(0.5)$ with $Y=0$ if $F=0$, and $Y=0.8$ if $F=1$. The marginal mean
is $\theta_2=0.5(0.0)+0.5(0.8)=0.4$, and the marginal distribution of $F$ is again $\operatorname{Bernoulli}(0.5)$.

Since $\theta_1=0.6>\theta_2=0.4$, Arm 1 is the better arm ($k^\star=1$).

\paragraph{Instance B (Arm 2 is better).}
We swap the arm distributions. Arm 1 uses the distribution of Arm 2
from Instance A ($\theta_1=0.4$), and Arm 2 uses the distribution of Arm 1 from Instance A ($\theta_2=0.6$). In
both cases, the marginal distribution of $F$ remains $\operatorname{Bernoulli}(0.5)$. Since $\theta_2=0.6>\theta_1=0.4$, Arm
2 is now better ($k^\star=2$).

A proxy-only algorithm observes only the sequence of proxy scores
(\rev{$\{(k_t,F_t)\}_{t\le\tau}$}). In both
instances, for any arm $k$, the marginal distribution of $F_t$ is $\operatorname{Bernoulli}(0.5)$. Since the algorithm
does not observe $Y_t$, the probability distribution of the observed history is identical for Instance A
and Instance B. Let $\psi$ be the algorithm's decision rule mapping history to a selected arm, and let
$p=\Pbb(\psi(\mathrm{History})=1)$. On Instance A, the error probability is $\Pbb(\mathrm{Error})=\Pbb(\mathrm{Select}\ 2)=1-p$, and
on Instance B, $\Pbb(\mathrm{Error})=\Pbb(\mathrm{Select}\ 1)=p$. The maximum error probability is $\max(1-p,p)\ge0.5$.
Thus, no proxy-only algorithm can achieve $\delta$-correctness for any $\delta<0.5$.

\rev{This equality also holds under adaptive choices and random stopping. Given any finite history, both models have the same next proxy law, policy and stopping kernels, thus the same stopped recommendation law. Here $p$ denotes only the recommendation probability.}

{\color{black}
\paragraph{Proof of Part 2.}
\rev{In this construction, $t$ indexes pulls of a fixed arm and $n$ denotes its pull count. Choose $\pi_{\min}=p\in(0,0.5)$. Audit randomization is independent of the outcome conditional on the proxy, as required by Assumption~\ref{ass:audit}.}

Consider a naive estimator that estimates $\theta_k$ by the simple average of audited outcomes, ignoring
the selectivity of auditing\rev{,}
\begin{equation}
\widehat\theta_k^{\mathrm{naive}}
=\frac{\sum_{t=1}^n A_tY_t}{\sum_{t=1}^n A_t}.
\label{eq:restored-naive-estimator}
\end{equation}
Let the audit policy depend on the proxy score while satisfying positivity (Assumption~\ref{ass:audit}), i.e., for
some small $\pi_{\min}>0$,
\begin{equation}
\pi(f)=
\begin{cases}
1&\text{if }f>0.5,\\
\pi_{\min}&\text{otherwise}.
\end{cases}
\label{eq:restored-naive-policy}
\end{equation}
This policy satisfies $\pi(f)\ge\pi_{\min}$ for all $f$, but concentrates audits in regions with high proxy scores.
Consider an arm where $F\sim\operatorname{Uniform}[0,1]$.

\paragraph{Instance A (Positive Correlation).}
Let $Y=F$. The true mean is $\theta=\E[F]=0.5$. Under the audit
policy, the audit probability is $\Pbb(A=1)=0.5(1+\pi_{\min})$. The conditional expectation of $Y$ given
audit is \rev{given by}
\begin{equation}
\begin{aligned}
\E[Y\mid A=1]
&=\frac{\Pbb(F>0.5)\cdot\E[Y\mid F>0.5]+\pi_{\min}\cdot\Pbb(F\le0.5)\cdot\E[Y\mid F\le0.5]}{\Pbb(A=1)}\\
&=\frac{0.5\cdot0.75+\pi_{\min}\cdot0.5\cdot0.25}{0.5(1+\pi_{\min})}
=\frac{0.75+0.25\pi_{\min}}{1+\pi_{\min}}.
\end{aligned}
\end{equation}
By the law of large numbers, $\widehat\theta^{\mathrm{naive}}\xrightarrow{a.s.}\frac{0.75+0.25\pi_{\min}}{1+\pi_{\min}}\ne0.5$ for any $\pi_{\min}<1$. For small $\pi_{\min}$, the
asymptotic bias is $+0.25(1-\pi_{\min})/(1+\pi_{\min})\to+0.25$ as $\pi_{\min}\to0$.

\paragraph{Instance B (Negative Correlation).}
Let $Y=1-F$. The true mean is $\theta=\E[1-F]=0.5$. A similar
calculation shows
\begin{equation}
\E[Y\mid A=1]
=\frac{0.5\cdot0.25+\pi_{\min}\cdot0.5\cdot0.75}{0.5(1+\pi_{\min})}
=\frac{0.25+0.75\pi_{\min}}{1+\pi_{\min}}.
\end{equation}
For small $\pi_{\min}$, the asymptotic bias is $-0.25(1-\pi_{\min})/(1+\pi_{\min})\to-0.25$ as $\pi_{\min}\to0$.

We embed the above inconsistency into a $K=2$ best arm identification instance. For each arm
$k\in\{1,2\}$, let the proxy score satisfy $F\sim\operatorname{Uniform}[0,1]$ on each pull. Consider the audit policy
\eqref{eq:restored-naive-policy}. Define the outcomes by
\[
Y(1):=F,\qquad
Y(2):=
\begin{cases}
1,&F\le0.5,\\
1-F,&F>0.5.
\end{cases}
\]
Both outcomes lie in $[0,1]$ almost surely, so Assumption~\ref{ass:bounded} holds.

We have
\[
\theta_1=\E[Y(1)]=\E[F]=0.5,
\]
and
\begin{equation}
\begin{aligned}
\theta_2=\E[Y(2)]
&=\E[1\cdot\mathbf1\{F\le0.5\}]+\E[(1-F)\cdot\mathbf1\{F>0.5\}]\\
&=0.5\cdot1+0.5\cdot\E[1-F\mid F>0.5]=0.5+0.5\cdot(1-0.75)=0.625.
\end{aligned}
\end{equation}
Then Arm 2 is better ($k^\star=2$).

Consider the naive selective-audit estimator \eqref{eq:restored-naive-estimator}. The conditional mean among audited samples for Arm 1 is
\begin{equation}
m_1:=\E[Y(1)\mid A=1]=\frac{0.75+0.25\pi_{\min}}{1+\pi_{\min}}>0.5.
\end{equation}
For Arm 2,
\begin{equation}
\begin{aligned}
m_2:=\E[Y(2)\mid A=1]
&=\frac{\E[Y(2)\mathbf1\{F>0.5\}]+\pi_{\min}\E[Y(2)\mathbf1\{F\le0.5\}]}{\Pbb(F>0.5)+\pi_{\min}\Pbb(F\le0.5)}\\
&=\frac{0.5\cdot\E[1-F\mid F>0.5]+\pi_{\min}\cdot0.5\cdot1}{0.5(1+\pi_{\min})}
=\frac{0.25+\pi_{\min}}{1+\pi_{\min}}<0.5,
\end{aligned}
\end{equation}
where the last inequality uses $\pi_{\min}<1/2$.

Moreover, for $\pi_{\min}<2/3$ (in particular for $\pi_{\min}<0.5$), we have $m_1>m_2$ since
\[
m_1-m_2=\frac{(0.75+0.25\pi_{\min})-(0.25+\pi_{\min})}{1+\pi_{\min}}
=\frac{0.5-0.75\pi_{\min}}{1+\pi_{\min}}>0.
\]
On any event where each arm is pulled infinitely often and receives infinitely many audits (which
holds almost surely, e.g., under any strategy that continues sampling both arms while $\pi_{\min}>0$), the
strong law of large numbers \rev{gives}
\[
\widehat\theta_1^{\mathrm{naive}}(n)\xrightarrow{a.s.}m_1,
\qquad
\widehat\theta_2^{\mathrm{naive}}(n)\xrightarrow{a.s.}m_2,
\qquad m_1>m_2,
\]
while the true ordering is $\theta_2>\theta_1$.
\rev{Consequently, whenever both arm counts diverge, the naive empirical ranking eventually favors Arm~1 almost surely. Under a strategy that samples both arms infinitely often almost surely, its probability of favoring the wrong arm therefore tends to one. \halmos}
}
\end{proof}

\begin{revblock}
\subsection{The augmented inverse-propensity identity}

Based on the prediction-powered estimator in Section 4.1, we extend the IPW residual identity to an augmented estimator whose baseline is fitted using earlier returned audits. The result establishes fixed-count unbiasedness when the current baseline is determined before the audit draw.

For a baseline $m_{k,t}$ fitted from earlier observations and evaluated at the
current pre-audit state, define
\begin{equation}
\widehat{\theta}^{\,\mathrm{AIPW}}_k(t)
=\frac{1}{N_k(t)}\sum_{s\le t:k_s=k}
\left[m_{k,s}+\frac{A_s}{\pi_s}(Y_s-m_{k,s})\right].
\label{eq:aipw-estimator}
\end{equation}
\rev{Clip the fitted value to $[0,1]$ and fix it before the audit draw, using the current instance, proxy and earlier returned audits. For example, $m_{k,t}=\clip(a_{k,t}+\lambda_{k,t}F_t,0,1)$ uses an intercept and proxy weight fitted from earlier records. A small weight reduces reliance on a weak proxy.}
\begin{lemma}\label{lem:aipw-unbiased}
Under \textcolor{.}{Assumptions~1--3} for one-configuration evaluations, suppose
$m_{k,t}\in[0,1]$ is measurable with respect to the complete pre-audit state.
Each summand in \eqref{eq:aipw-estimator} has conditional mean $\theta_k$
after averaging the current state. For every fixed positive integer $n$ whose
$n$th arm-$k$ pull occurs almost surely, the average of the first $n$
summands is unbiased for $\theta_k$.
\end{lemma}

According to the result in this lemma, using a baseline learned from earlier returned audits does not change the target mean of the corrected score. It therefore permits online baseline updates while preserving fixed-count unbiasedness, provided that the current baseline is fixed before the audit draw.

\begin{proof}[Proof of Lemma~\ref{lem:aipw-unbiased}.]
Write $\mathcal H_t^S=\sigma(\mathcal H_{t-1},k_t,X_t,F_t)$, augmented by
the fitted baseline if necessary. Conditional on this state and $Y_t$, the audit-independence condition
and logged probability give
\[
\E\!\left[m_{k,t}+\frac{A_t}{\pi_t}(Y_t-m_{k,t})
\middle|\mathcal H_t^S,Y_t\right]=Y_t.
\]
Taking expectations first over the verified outcome and then over the fresh
state gives conditional mean $\theta_k$ given the pre-round history and
$k_t=k$. The summand lies in $[1-1/p,1/p]$, so the same calculation is
integrable at each almost-surely finite arm-pull time. Applying the tower
property at the first $n$ such times and dividing by $n$ proves the claim.
Conditioning on training data covers independent fitting, and conditioning on earlier outcomes covers online fitting.
\halmos
\end{proof}
\end{revblock}

\begin{revblock}
\subsection{The bundle AIPW identity}
We next extend the augmented inverse-propensity argument to the shared-reference setting in Section 4.2, where a baseline fitted from earlier returned references can be used for every applicable configuration in a selected bundle.

\begin{lemma}\label{lem:bundle-aipw-unbiased}
Under \textcolor{.}{Assumptions~1--3}, let
$m_{k,t}\in[0,1]$ be measurable before the current purchase draw. For every
bundle round containing $k$,
\[
\E\!\left[m_{k,t}+\frac{V_t}{\rho_t}(Y_{k,t}-m_{k,t})
\middle|\mathcal G_{t-1},B_t,G_t\right]
=\E[Y_{k,t}\mid\mathcal G_{t-1},B_t,G_t].
\]
After averaging the current state, the conditional mean is $\theta_{\nu,k}$.
The average of the first $n$ appearances of $k$ is unbiased whenever that
appearance count is reached almost surely.
\end{lemma}
\begin{proof}[Proof of Lemma~\ref{lem:bundle-aipw-unbiased}.]
Condition on the history, selected bundle, current state and verified vector.
The purchase rule gives
$\E[V_t/\rho_t\mid\mathcal G_{t-1},B_t,G_t,Y_{B_t,t}]=1$. Hence the
conditional expectation of the displayed score is $Y_{k,t}$. Averaging the
verified vector and state gives $\theta_{\nu,k}$ by bundle consistency. Its
bounded range $[1-1/p,1/p]$ permits the same tower calculation at the first
$n$ appearance times. This proves the fixed-count assertion. Taking
$m_{k,t}=F_{k,t}$ gives the unlearned bundle estimator. \halmos
\end{proof}

\paragraph{Confidence sequences with a learned baseline.}
To allow updates after each audit returns, construct the confidence sequence from the corrected score
\[
H_{k,t}=m_{k,t}+\frac{V_t}{\rho_t}(Y_{k,t}-m_{k,t}),\qquad
b_p=\frac{2}{p}-1,
\]
on a bundle round containing $k$, and retain
$M_k(t)=\sum_{s\le t}\mathbf 1\{k\in B_s\}$. For singleton rounds, replace
$(V_t,\rho_t,M_k)$ by $(A_t,\pi_t,N_k)$.
\begin{proposition}\label{prop:aipw-cs}
Under the conditions of Lemma~\ref{lem:bundle-aipw-unbiased}, let
$\widehat\theta_k^{\rm AIPW}(t)$ be the average of the observed corrected
scores for configuration $k$. With $\psi$ defined in
\eqref{eq:stitched-boundary}, put
\begin{equation}
w_k^{\rm AIPW}(t)=
\frac{\psi\!\left(b_p^2M_k(t)/4;\delta/K\right)}{M_k(t)}.
\label{eq:aipw-cs-width}
\end{equation}
Then, with probability at least $1-\delta$, simultaneously for every $k$ and
every $t$ with $M_k(t)\ge1$,
\[
\theta_{\nu,k}\in
\left[\widehat\theta_k^{\rm AIPW}(t)\pm w_k^{\rm AIPW}(t)\right].
\]
The statement allows online fitting from previous returned records.
\end{proposition}
\begin{proof}[Proof of Proposition~\ref{prop:aipw-cs}.]
When $V_t=0$, $H_{k,t}=m_{k,t}\in[0,1]$. When $V_t=1$,
\[
H_{k,t}=\frac{Y_{k,t}}{\rho_t}
+\left(1-\frac{1}{\rho_t}\right)m_{k,t}
\in[1-1/p,1/p].
\]
Thus every score lies in an interval of length $b_p$. By
Lemma~\ref{lem:bundle-aipw-unbiased}, its conditional mean, given the
history and the chosen bundle before the current state is drawn, is
$\theta_{\nu,k}$. Hoeffding's lemma gives the conditional sub-Gaussian
variance proxy $b_p^2/4$. More explicitly, for
$I_{k,t}=\mathbf 1\{k\in B_t\}$, the processes
\[
\exp\!\left\{\lambda\sum_{s\le t}I_{k,s}(H_{k,s}-\theta_{\nu,k})
-\frac{\lambda^2b_p^2}{8}M_k(t)\right\},\qquad\lambda\in\mathbb R,
\]
are nonnegative supermartingales. Conditioning first on the bundle and then averaging its selection covers configurations sampled only finitely often. The
stitched-boundary proof below applies to these supermartingales with
variance clock $b_p^2M_k(t)/4$. Using error budget $\delta/K$ for each
configuration and taking a union bound proves the result.
\halmos
\end{proof}

Note that Proposition \ref{prop:aipw-cs} strengthens the fixed-count identity above to an anytime-valid confidence guarantee. It therefore supports sequential selection and stopping with an online-updated baseline in the settings of Sections 4.1 and 4.2.
\end{revblock}

\subsection{Proof of Lemma~\ref{lem:stopping-separation}}

\begin{revblock}
Next, we provide the following lemma, which gives a sufficient confidence-width condition for the separation rule in \eqref{eq:stop} to certify the unique best arm. It will be used below to establish the almost-sure stopping claim in Theorem \ref{thm:delta-correct}.

\begin{lemma}\label{lem:stopping-separation}
Suppose $k^\star$ is the unique best arm. On the simultaneous coverage event
of \textcolor{.}{Theorem~2}, if
\[
2\{w_{F,k^\star}(t)+w_{R,k^\star}(t)\}
+2\{w_{F,k}(t)+w_{R,k}(t)\}<\theta_{k^\star}-\theta_k
\]
for every competitor $k$, the stopping rule in \eqref{eq:stop} stops at $t$
and recommends $k^\star$.
\end{lemma}

\begin{proof}[Proof of Lemma~\ref{lem:stopping-separation}.]
Write $w_k(t)=w_{F,k}(t)+w_{R,k}(t)$ for the total confidence width.
On the coverage event,
$L_{k^\star}(t)\ge\theta_{k^\star}-2w_{k^\star}(t)$ and
$U_k(t)\le\theta_k+2w_k(t)$. The displayed condition therefore gives
$L_{k^\star}(t)>U_k(t)$ for every competitor. Hence $k^\star$ is the unique
leader satisfying the stopping rule.
\halmos
\end{proof}
\end{revblock}

\subsection{Proof of the dyadic stitched boundary}
\begin{revblock}

We next establish the dyadic stitched boundary used to construct the anytime-valid confidence sequences in Propositions 1 and 2. The result provides uniform control over all intrinsic times and is also used in the learned-baseline extension in Proposition~\ref{prop:aipw-cs}. Here intrinsic time is measured by the accumulated variance bound $V_t$, which depends on the sampling and audit rules.

For $j\ge0$, let $I_0=(0,1]$ and $I_j=(2^{j-1},2^j]$ for $j\ge1$.
Set $U_0=1$, $U_j=2^j$, and
$w_j=6/[\pi^2(j+1)^2]$. For $v\in I_j$ define
\[
\psi(v;\alpha)=\sqrt{2U_j\log\!\left(\frac{2}{\alpha w_j}\right)}.
\]
\begin{theorem}\label{thm:howard-subg-ref}
Let $(S_t,V_t)$ be a sub-Gaussian martingale and its predictable,
nondecreasing variance process, with $S_0=V_0=0$. If
$\log\E[\exp\{\lambda(S_t-S_{t-1})\}\mid\mathcal G_{t-1}]\le
\lambda^2(V_t-V_{t-1})/2$ for every $\lambda$, then
\[
\Prob\{\exists t:V_t>0,\ |S_t|\ge\psi(V_t;\alpha)\}\le\alpha.
\]
\end{theorem}

\begin{proof}
For each $\lambda>0$, the conditional mgf assumption makes
$L_t(\lambda)=\exp\{\lambda S_t-\lambda^2V_t/2\}$ a nonnegative
supermartingale. Fix $j$ and put
$b_j=\sqrt{2U_j\log(2/(\alpha w_j))}$ and
$\lambda_j=b_j/U_j$. On the event
$\{\exists t:V_t\in I_j, S_t\ge b_j\}$, the process
$L_t(\lambda_j)$ exceeds
$\exp\{\lambda_j b_j-\lambda_j^2U_j/2\}=
\exp\{b_j^2/(2U_j)\}=2/(\alpha w_j)$.
Ville's inequality therefore bounds this event by $\alpha w_j/2$.
Applying the same argument to $-S_t$ gives the same bound for a lower
crossing. The intervals $I_j$ partition the positive intrinsic times and
$\sum_{j\ge0}w_j=1$, so a union bound over signs and $j$ proves the claim.
\halmos
\end{proof}

\end{revblock}

\subsection{Proof of Proposition~\textcolor{.}{1}}
\begin{revblock}
We apply the stitched boundary to the proxy scores. This gives confidence sequences that remain valid when configuration choices depend on earlier observations.

\begin{proof}[Proof of Proposition~\textcolor{.}{1}.]
Fix arm $k$ and put $I_{k,t}=\mathbf 1\{k_t=k\}$. Conditional on the
history and selected arm before the fresh instance is drawn,
$\E[F_t\mid\mathcal H_{t-1},k_t=k]=\mu_{F,k}$. Since $F_t\in[0,1]$,
Hoeffding's lemma gives
\[
\E\!\left[\exp\{\lambda(F_t-\mu_{F,k})\}
\middle|\mathcal H_{t-1},k_t=k\right]\le\exp(\lambda^2/8).
\]
It follows, by first conditioning on the action and then averaging its
randomization, that
\[
\exp\!\left\{\lambda S_{F,k}(t)-\frac{\lambda^2}{8}N_k(t)\right\}
\]
is a nonnegative supermartingale for every $\lambda\in\mathbb R$. A round
that selects another arm leaves this process unchanged. The stitched-boundary
argument with variance clock $N_k(t)/4$ therefore gives
\[
\Pbb\!\left(\exists t:N_k(t)\ge1,\quad
|S_{F,k}(t)|\ge\psi(N_k(t)/4;\delta_{F,k})\right)\le\delta_{F,k}.
\]
Dividing by $N_k(t)$ proves the stated interval at every positive count,
including when an arm is sampled only finitely often. Taking
$\delta_{F,k}=\delta/(2K)$ and a union bound gives total proxy failure
probability at most $\delta/2$.
\halmos
\end{proof}
\end{revblock}

\subsection{Proof of Proposition~\textcolor{.}{2}}
\begin{revblock}
We next control the IPW residual terms. The audit floor bounds the size of each term, allowing the same stitched-boundary argument to apply after selective auditing.

\begin{proof}[Proof of Proposition~\textcolor{.}{2}.]
Write $\mathcal H_t^S=\sigma(\mathcal H_{t-1},k_t,X_t,F_t)$ for the complete
pre-audit state. For $Z_t=(A_t/\pi_t)(Y_t-F_t)$, conditional audit independence gives
\[
\E[Z_t\mid\mathcal H_t^S]
=\E[Y_t-F_t\mid\mathcal H_t^S].
\]
Averaging the current state, which is fresh conditional on the selected arm,
gives $\E[Z_t\mid\mathcal H_{t-1},k_t=k]=\mu_{R,k}$. Moreover,
$Z_t\in[-1/p,1/p]$. Subtracting $\mu_{R,k}$ translates this interval without
changing its length $M=2/p$. Hoeffding's lemma thus gives conditional
sub-Gaussian variance proxy $M^2/4$.

With $I_{k,t}=\mathbf 1\{k_t=k\}$, the same staged calculation as for the
proxy shows that
\[
\exp\!\left\{\lambda\sum_{s\le t}I_{k,s}(Z_s-\mu_{R,k})
-\frac{\lambda^2M^2}{8}N_k(t)\right\}
\]
is a nonnegative supermartingale. The stitched boundary with variance clock
$N_k(t)M^2/4$ gives
\[
\Pbb\!\left(\exists t:N_k(t)\ge1,\quad
\left|\widehat\mu_{R,k}(t)-\mu_{R,k}\right|
\ge\frac{\psi(N_k(t)M^2/4;\delta_{R,k})}{N_k(t)}\right)
\le\delta_{R,k}.
\]
The argument does not condition on an arm reaching a future sample count.
Setting $\delta_{R,k}=\delta/(2K)$ and taking a union bound over arms gives
the remaining failure budget $\delta/2$.
\halmos
\end{proof}
\end{revblock}

\subsection{Proof of Theorem~\textcolor{.}{2}}

\begin{proof}[Proof of Theorem~\textcolor{.}{2}]
\rev{We combine the proxy and residual confidence sequences to obtain simultaneous coverage for all configuration means. This coverage event will also be used to prove correctness at the stopping time.}

\begin{revblock}
Define $\mathcal{E}$ as the event that all confidence sequences hold for
all arms at every time at which every arm has been sampled\rev{,}
\[
 \mathcal{E} = \bigcap_{k=1}^K
 \bigcap_{\substack{t\ge1:\\N_j(t)\ge1\ \forall j\in[K]}}
 \{ \theta_k \in [L_k(t), U_k(t)] \}.
\]
\end{revblock}
Each interval $[L_k(t), U_k(t)]$ combines a proxy confidence sequence (width $w_{F,k}(t)$) and a residual confidence sequence (width $w_{R,k}(t)$). By \textcolor{.}{Propositions~1 and~2}, each type of sequence (proxy and residual) has total failure probability at most $\delta/2$ across all $K$ arms. \rev{Regardless of any dependence between the two sequence families, a union bound gives}
\[ \Pbb(\mathcal{E}^c) \le \frac{\delta}{2} + \frac{\delta}{2} = \delta. \]
\rev{Thus $\Pbb(\mathcal{E}) \ge 1 - \delta$.}
\halmos
\end{proof}

\subsection{Proof of Theorem~\textcolor{.}{6}}
\begin{revblock}
We use simultaneous coverage to prove correctness and continued exploration to prove finite stopping. Both steps are needed because valid intervals alone do not ensure that the stopping condition is reached.

\begin{proof}[Proof of Theorem~\textcolor{.}{6}.]
Let $\mathcal E$ be the simultaneous coverage event in
\textcolor{.}{Theorem~2}. On $\mathcal E$, if a separation rule
stops at $\tau<\infty$ and returns $\widehat k$, then, for every
$k\ne\widehat k$,
\[
\theta_{\widehat k}\ge L_{\widehat k}(\tau)
>U_k(\tau)\ge\theta_k.
\]
Consequently $\widehat k=k^\star$, and
\[
\Pbb(\tau<\infty,\widehat k\ne k^\star)
\le\Pbb(\mathcal E^c)\le\delta.
\]

Continue Algorithm~\textcolor{.}{2} with the same sampling and audit updates but without returning. Through iteration $r$, each arm has its initialization pull and at least $\lfloor\sqrt r\rfloor$ exploration pulls, so $N_k\to\infty$ for every $k$. Centered proxy and IPW residual increments are bounded martingale differences on each arm's pull clock. Their strong law gives
\begin{align*}
&\widehat\mu_{F,k}\longrightarrow\mu_{F,k},\\
&\widehat\mu_{R,k}\longrightarrow\mu_{R,k},
\end{align*}
almost surely. Dividing increments by pull indices gives summable expected squares. Martingale convergence and Kronecker's lemma then give the result.
Moreover, $\psi(v;\alpha)=O(\sqrt{v\log\log v})$ for fixed $\alpha$, so
both widths tend to zero. The unique best arm eventually leads with its lower endpoint above all competing upper endpoints. The procedure agrees with its continuation until this separation, proving $\tau<\infty$ almost surely. The finite-stop error bound then gives $\Pbb(\widehat k=k^\star)\ge1-\delta$.

When the baseline is estimated from earlier observations,
Proposition~\ref{prop:aipw-cs} gives simultaneous coverage. The corrected score has conditional mean
$\theta_k$ and bounded range, so the same strong-law and sparse-exploration
argument proves almost sure stopping.
\halmos
\end{proof}
\end{revblock}

\begin{revblock}
\section{Proofs of Theoretical Results in Section 5}
We now turn to the shared-reference setting introduced in Section 3.1 and analyzed in Section 5, where a single acquired reference can verify the outputs of multiple configurations in a selected bundle. This section establishes the information-theoretic lower bound for this setting and proves the strict cost gains from reference sharing and state-dependent audit timing. Throughout, the KL divergence of the verified outcome vector accounts for dependence among jointly verified outcomes.

\subsection{Proof of Theorem~\textcolor{.}{3}}

\begin{proof}

If $\mathcal A=\varnothing$, the zero allocation is feasible and
$C_{\rm bun}^\star(\nu,\mathcal A)=0$, so the bound follows from
nonnegative costs. We therefore assume $\mathcal A\ne\varnothing$.
Fix an alternative \(a\in\mathcal A\). Let \(N_B(n)\) be the number of
bundle evaluations up to deterministic time \(n\), and let \(N_{Bg}^Y(n)\)
be the number of purchased references for bundle \(B\) at state \(g\). Put
\(\tau_n=\tau\wedge n\) and use a common cemetery record after stopping. For
the event
\[
E_n=\{\tau\leq n,\ \widehat k=k^\star(\nu)\},
\]
an almost-surely terminating uniformly $\delta$-correct policy satisfies
$P_\nu(E_n)\uparrow P_\nu(\widehat k=k^\star(\nu))\geq1-\delta$.
For the alternative, $P_{\nu_a}(E_n)\leq\delta$ for every $n$. The data processing
inequality for binary relative entropy gives
\[
\operatorname{KL}\left(P_\nu^{\mathcal G_n}\middle\Vert
P_{\nu_a}^{\mathcal G_n}\right)
\geq \kl\bigl(P_\nu(E_n),P_{\nu_a}(E_n)\bigr).
\]
The left-hand side is the expected sum of one-step observed-record KL terms.
Since $1-\delta>\delta$, we have $P_\nu(E_n)>\delta$ for all
sufficiently large $n$. For such \(n\), put
\(\kappa_n=\min_{a\in\mathcal A}\kl(P_\nu(E_n),P_{\nu_a}(E_n))>0\).
The stopped calculation gives, for every \(a\in\mathcal A\),
\begin{equation}
\E_\nu\left[\sum_{B\in\mathcal B}N_B(\tau_n)d_{B0}^a
 +\sum_{B\in\mathcal B}\sum_{g\in\mathcal G_B}N_{Bg}^Y(\tau_n)h_{Bg}^a\right]
\geq \kappa_n,
\label{eq:proof-bundle-change-measure-truncated}
\end{equation}
The probabilities satisfy \(P_\nu(E_n)\uparrow
P_\nu(\widehat k=k^\star(\nu))\geq1-\delta\) and
\(P_{\nu_a}(E_n)\leq\delta\), so
\(\liminf_n\kappa_n\geq\kappa_\delta\), where
\(\kappa_\delta=\kl(1-\delta,\delta)\). The likelihood decomposition used
here follows by cancelling the common bundle and purchase kernels in the
one-step likelihood ratio and then conditioning on the pre-audit state.

Put \(n_{B,n}=\E_\nu[N_B(\tau_n)]\). Since the event
\(\{\tau_n\ge t\}\) and the choice of \(B_t\) are measurable before the fresh
state at round \(t\) is drawn, the fresh-state assumption gives
\[
\E_\nu\left[\sum_{t\leq\tau_n}\1\{B_t=B,G_t=g\}\right]
=\mu_{\nu,B}(g)n_{B,n}.
\]
Define \(q_{B,n}=n_{B,n}/\kappa_n\). If \(n_{B,n}>0\), define
\[
x_{Bg,n}=\frac{\E_\nu[N_{Bg}^Y(\tau_n)]}
{\kappa_n\mu_{\nu,B}(g)}.
\]
The purchase floor gives \(p q_{B,n}\leq x_{Bg,n}\leq q_{B,n}\). If
\(n_{B,n}=0\), set \(q_{B,n}=x_{Bg,n}=0\). Dividing
\eqref{eq:proof-bundle-change-measure-truncated} by \(\kappa_n\) gives every information constraint in
\textup{(\textcolor{.}{15})}. To identify the cost, note first that
$\mathbf 1\{t\leq\tau_n,B_t=B\}$ is measurable before the fresh state at
round $t$ is drawn. Hence predictable sampling and the fresh-state law give
\[
\E_\nu\!\left[\sum_{t\leq\tau_n}\mathbf 1\{B_t=B\}c_B^F(G_t)\right]
=\sum_g\mu_{\nu,B}(g)c_B^F(g)\,\E_\nu N_B(\tau_n).
\]
Similarly, conditioning on the statewise purchase coin gives
$\E_\nu\sum_{t\leq\tau_n}\mathbf 1\{B_t=B,G_t=g\}V_t c_{Bg}^Y
=c_{Bg}^Y\E_\nu N_{Bg}^Y(\tau_n)$. Therefore the truncated expected cost equals
\[
\kappa_n\left(\sum_{B\in\mathcal B}\sum_{g\in\mathcal G_B}\mu_{\nu,B}(g)c_B^F(g)q_{B,n}+
\sum_{B\in\mathcal B}\sum_{g\in\mathcal G_B}\mu_{\nu,B}(g)c_{Bg}^Yx_{Bg,n}\right).
\]
Minimizing gives
\(\E_\nu[\operatorname{Cost}_{\mathcal B}(\tau_n)]
\ge\kappa_n C_{\rm bun}^\star(\nu,\mathcal A)\).
Monotone convergence of the nonnegative costs and
\(\liminf_n\kappa_n\ge\kappa_\delta\) prove the lower bound, including
policies with infinite expected counts or zero-cost rounds. The feasible region is a finite-dimensional closed polyhedron. Nonnegative costs give a finite infimum whenever the information constraints are feasible, so the linear programming gives an optimizer. Removing zero-cost directions preserves its value.

For the dual, attach a nonnegative multiplier \(\eta_a\) to each information
constraint. For fixed \(q_B\), write \(x_{Bg}=q_Br_{Bg}\). The coefficient
of \(q_B\) in the resulting Lagrangian is
\[
\sum_g\mu_{\nu,B}(g)c_B^F(g)-\sum_a\eta_ad_{B0}^a+
\sum_g\mu_{\nu,B}(g)\min_{r\in[p,1]}
r\left(c_{Bg}^Y-\sum_a\eta_ah_{Bg}^a\right).
\]
The infimum over nonnegative \(q_B\) is finite exactly when this coefficient
is nonnegative. Finite linear programming duality gives
\textup{(\textcolor{.}{17})}. The pointwise minimizer is one when the
parenthesized coefficient is negative and \(p\) when it is positive. \halmos
\end{proof}

\subsection{Proof of Theorem~\textcolor{.}{4}}

\begin{proof}
For the first construction, take independent Bernoulli verified outcomes under the true
model with means $(0.6,0.4)$ and no information in the pre-audit
state. Let alternative $a_1$ change only the first mean to $0.2$, and let
alternative $a_2$ change only the second mean to $0.8$. The two alternatives
have the same positive conditional divergence
\[
h=\kl(0.6\Vert0.2)=\kl(0.4\Vert0.8).
\]
Evaluating configuration one alone supplies $h$ only to the
$a_1$ constraint, and evaluating configuration two alone supplies
$h$ only to the $a_2$ constraint. Evaluating both together provides $h$ to
both constraints because its returned vector has both components. Since the state is uninformative, replacing feasible $(q_B,x_B)$ by $(x_B,x_B)$ preserves information and reference-acquisition costs while weakly reducing configuration-evaluation costs. Thus an optimizer can audit every round. If
$q_{12},q_1,q_2$ denote the three evaluation allocations, feasibility is
\[
h(q_{12}+q_1)\ge1,\text{ and } h(q_{12}+q_2)\ge1.
\]
For a fixed $r=hq_{12}\in[0,1]$, the least feasible singleton intensities
are $q_1=q_2=(1-r)/h$. The resulting cost is
\[
\frac{r(2c_F+c_Y)+2(1-r)(c_F+c_Y)}{h}
=\frac{2c_F+2c_Y-r c_Y}{h},
\]
which is minimized at $r=1$. Hence the optimum allowing bundle evaluations is
$C_{\rm bun}^\star=(2c_F+c_Y)/h$. Restricting to singleton actions
requires $q_1=q_2=1/h$ and gives
$C_{\rm sing}^\star=2(c_F+c_Y)/h$. Therefore the bundle is strictly
better whenever $c_Y>0$, while adding a bundle with a weakened joint signal
would require a separate comparison of the corresponding information terms.

For the construction concerning purchase timing, let the state be Bernoulli with equal
probabilities, and let the true verified means of the two configurations be
$0.6$ and $0.4$. Let alternative $a_1$ change only configuration one in
state one to a Bernoulli mean $a<0.2$ satisfying
$\kl(0.6\Vert a)=10$, and let alternative $a_2$ change only configuration
two in state two to a Bernoulli mean $b>0.8$ satisfying
$\kl(0.4\Vert b)=1$. Such roots exist by continuity and the divergence to
$+\infty$ at the boundary. The true best configuration is one. Under $a_1$
and $a_2$, configuration two is best. The state has the same law under all three models, so observing it alone cannot distinguish them. It still indicates which alternative an audit can distinguish.
With purchase floor $p=0.1$, let $q$ be the bundle intensity and $x_i=q\rho_i$
the statewise purchase intensities. The two information constraints are
\begin{align*}
&q(1/2)\rho_1(10)=5x_1\ge1,\\
&q(1/2)\rho_2(1)=x_2/2\ge1,
\end{align*}
so $x_1\ge0.2$ and $x_2\ge2$. Since $x_i\le q$, the least evaluation
allocation is $q=2$, attained with $(x_1,x_2)=(0.2,2)$. The floor is
satisfied because $0.1q=0.2\le x_i$. If its configuration-evaluation cost is $c_B^F$, the expected cost is therefore
$2c_B^F+\tfrac12(0.2+2)c_Y=2c_B^F+1.1c_Y$.
For a constant purchase probability, $x_1=x_2=q\rho$ and the second
constraint forces $q\rho\ge2$. Since $\rho\le1$, configuration evaluation costs at least
$2c_B^F$, while the expected reference count is at least $2$. Thus every
constant-rate policy costs at least $2c_B^F+2c_Y$, which is strictly larger
when $c_Y>0$.
\halmos
\end{proof}

\end{revblock}

\begin{revblock}
\section{Proof of the Allocation Result in Section 5}
\label{sec:bundle-tracker-proof}
This section proves Theorem~\ref{thm:bundle-tracker} in Section 5.4 for the joint allocation algorithm given in Algorithm~\ref{alg:joint-bundle}. We show that Algorithm~\ref{alg:joint-bundle} identifies the best configuration with the stated confidence guarantee, has finite expected stopping time and cost, and asymptotically attains the shared-reference characteristic cost $C_{\rm bun}^\star(\nu,\mathcal A_\nu)$ as $\delta\downarrow0$. To analyze these properties, we first consider the continuation of Algorithm~\ref{alg:joint-bundle} without its stopping rule. This continuation has the same sampling and auditing decisions as the stopped procedure up to its stopping time.

Fix the true model $\nu$. Models with identical observation laws for every bundle have the same best configuration and information constraints. Retain one representative per class, preserving observations and characteristic cost. Distinct retained models differ under some bundle. Write $J=|\mathcal B|$,
$\mathcal A=\mathcal A_\nu$, $H_t=\sum_{s=1}^t\varepsilon_s$, and
$\varepsilon_s=s^{-\gamma}$, $0<\gamma<1$. The case in which every model
has the same best configuration terminates at time zero. In the remainder,
$\mathcal A\ne\varnothing$. The finite alphabets and common support imply
a finite bound $L>0$ on the absolute value of every one-round log likelihood
ratio, uniformly over models, bundles and purchase maps in $[p,1]$.
The costs also have a finite one-round bound $c_{\max}$.

\subsection{Existence and selection of an allocation}
We first show that the characteristic cost program admits an optimizer for each candidate model and specify a deterministic rule for selecting one when the optimizer is not unique.

For each model, Assumption~\textcolor{.}{4} makes the information program feasible. Full auditing and sufficiently large allocations to every bundle satisfy all constraints because finitely many alternatives have positive distinguishing information. The objective is nonnegative. Moreover,
under the positive expected configuration-evaluation cost condition, a feasible point of cost at most
$c$ satisfies $\sum_Bq_B\le c/\underline c_F$, and $0\le x_{Bg}\le q_B$.
Thus nonempty feasible level sets are compact, so the optimum is attained.
Successive coordinate minimization on the compact optimal face selects one optimizer per model. Identifying the true model then selects its optimizer, without uniqueness or continuity assumptions.

Write this optimizer as $(q^\star,x^\star)$, put $S=\sum_Bq_B^\star$,
and use $w_B=q_B^\star/S$ and $\rho_{Bg}^\star=x_{Bg}^\star/q_B^\star$
on positive coordinates, with $\rho_{Bg}^\star=p$ otherwise.
Since there is a nonempty information constraint and all coefficients are
finite, $S>0$. Positive expected configuration-evaluation cost also gives
$C_{\rm bun}^\star>0$.

\subsection{Exploration and identification of the true model}

We next show that forced exploration keeps every bundle sufficiently sampled and allows the maximum-likelihood estimate to identify the true model after a finite random time.

\begin{lemma}\label{lem:bundle-deficit}
Let $r_{t,B}=(1-\varepsilon_t)w_B(\hat\nu_{t-1})+\varepsilon_t/J$ and
$W_B(t)=\sum_{s=1}^t r_{s,B}$. Under the update in Algorithm~\textcolor{.}{1},
\begin{equation}
|N_B(t)-W_B(t)|\le J,\qquad
N_B(t)\ge H_t/J-J
\quad(B\in\mathcal B,t\ge0).
\label{eq:bundle-deficit-proof}
\end{equation}
\end{lemma}
\begin{proof}
For $J=1$ the count and target both equal $t$. For $J\ge2$, let
$e_B(t)=W_B(t)-N_B(t)$. These deficits sum to zero. Suppose every deficit
at time $t-1$ exceeds $-1$. Before the pull, the updated deficits
$e_B(t-1)+r_{t,B}$ sum to one. Their largest entry is therefore at least
$1/J$. Its decrease by one leaves it greater than $-1$. Each unselected
entry also remains greater than $-1$, since $r_{t,B}>0$. Starting from
zero, induction gives $e_B(t)>-1$. The zero sum then gives
$e_B(t)<J-1$. The first claimed bound follows, and the second follows from
$W_B(t)\ge H_t/J$.
\halmos
\end{proof}

The preceding count bound ensures continued evaluation of every bundle. The next lemma uses this information to control the probability of selecting an incorrect model and the time of its last occurrence.

\begin{lemma}\label{lem:bundle-model-lock}
There are constants $A,a>0$, depending on the fixed model family and $p$ but not
on $t$ or $\delta$, such that
\begin{equation}
\Prob_\nu(\hat\nu_t\ne\nu)\le A\exp(-aH_t).
\label{eq:bundle-model-lock-tail}
\end{equation}
The time
$T=1+\sup\{t\ge0:\hat\nu_t\ne\nu\}$, with an empty supremum set to
zero, is almost surely finite and has finite moments of every positive
order.
\end{lemma}
\begin{proof}
Fix a different model $m$. Since observationally equivalent models have
been merged, at least one bundle $B_m$ has a different state law or a
different conditional verified law. Define
\begin{align*}
a_{Bg}(m)&=\sum_y\sqrt{Q_{\nu,B}(y\mid g)Q_{m,B}(y\mid g)},\\
\bar a_B(m)&=\sum_g\sqrt{\mu_{\nu,B}(g)\mu_{m,B}(g)}
[1-p+pa_{Bg}(m)].
\end{align*}
Common support gives $0<\bar a_B(m)\le1$, and the equality condition for
Cauchy-Schwarz gives $\bar a_{B_m}(m)<1$. Conditional on the pre-round
history, the bundle and entire purchase map are fixed. For
$R_t(m)=\exp\{\ell_t(m)-\ell_t(\nu)\}$, the conditional expectation of
the square root of its one-step ratio is
\[
a_t(m)=\sum_g\sqrt{\mu_{\nu,B_t}(g)\mu_{m,B_t}(g)}
[1-\rho_t(g)+\rho_t(g)a_{B_tg}(m)]
\le\bar a_{B_t}(m).
\]
Put $\Gamma_t(m)=\prod_{s=1}^t a_s(m)$. All factors are positive, and
$U_t(m)=\sqrt{R_t(m)}/\Gamma_t(m)$ is a nonnegative martingale with
$\E_\nu U_t(m)=1$. By Lemma~\ref{lem:bundle-deficit}, pathwise,
\[
\Gamma_t(m)\le \bar a_{B_m}(m)^{N_{B_m}(t)}
\le \bar a_{B_m}(m)^{H_t/J-J}.
\]
Markov's inequality therefore gives
\[
\Prob_\nu\{\ell_t(m)\ge\ell_t(\nu)\}
\le\E_\nu\sqrt{R_t(m)}
\le\bar a_{B_m}(m)^{H_t/J-J}.
\]
A union bound over the finite model family proves
\eqref{eq:bundle-model-lock-tail}. Integral comparison gives
$H_t\ge c_\gamma t^{1-\gamma}$ for sufficiently large $t$, with
$c_\gamma>0$. Consequently
\begin{equation}
\Prob_\nu(T>n)\le\sum_{s\ge n}A\exp(-aH_s),
\qquad n\ge1.
\label{eq:bundle-last-error-tail}
\end{equation}
The right side remains summable after multiplication by any fixed power of $n$, proving the assertions about $T$.
\halmos
\end{proof}

\subsection{Likelihood and cost rates}

Having established eventual identification of the true model, we next derive the long-run likelihood-information and cost rates induced by tracking its optimal allocation.

For $a\in\mathcal A$, define the one-round information and cost under the
selected optimal purchase rule
\begin{align*}
d_{Ba}^\star&=d_{B0}^a+\sum_g\mu_{\nu,B}(g)\rho_{Bg}^\star h_{Bg}^a,\\
c_B^\star&=\sum_g\mu_{\nu,B}(g)
[c_B^F(g)+\rho_{Bg}^\star c_{Bg}^Y],
\end{align*}
and put $I_a=\sum_Bw_Bd_{Ba}^\star$ and
$\bar c=\sum_Bw_Bc_B^\star$. LP feasibility and the objective identity give
\begin{equation}
I_a\ge d:=1/S>0\quad(a\in\mathcal A),\qquad
\bar c=C_{\rm bun}^\star/S>0.
\label{eq:bundle-optimal-rates}
\end{equation}

Let $Z_t=\sum_{s=1}^t\mathbf 1\{\hat\nu_{s-1}\ne\nu\}$, so that
$Z_t\le T$. The definition of the mixed targets and
Lemma~\ref{lem:bundle-deficit} give the pathwise bound
\begin{equation}
|N_B(t)-tw_B|\le J+Z_t+H_t\le J+T+H_t.
\label{eq:bundle-target-deviation}
\end{equation}
Indeed, the target can differ from $w_B$ by at most one on a round with an
incorrect maximum-likelihood model, plus $\varepsilon_t$ on any round.
Thus $N_B(t)/t\to w_B$ almost surely.

Write $L_t^a=\ell_t(\nu)-\ell_t(\nu_a)$ and let
$D_t^a$ be the sum of its conditional one-round means. The purchase floor,
fresh-state law and common kernels give
\[
D_t^a=\sum_{s=1}^t\left[d_{B_s0}^a+
\sum_g\mu_{\nu,B_s}(g)\rho_{B_sg}(\hat\nu_{s-1})h_{B_sg}^a\right].
\]
One-round information is bounded, and audit rules differ from the true model's rule only on $Z_t$ rounds. By \eqref{eq:bundle-target-deviation}, some finite $K_0$, uniform over $\mathcal A$, satisfies
\begin{equation}
|D_t^a-tI_a|\le K_0(1+T+H_t).
\label{eq:bundle-drift-deviation}
\end{equation}
The martingale $M_t^a=L_t^a-D_t^a$ has increments of absolute value at
most $2L$. Azuma's inequality and Borel--Cantelli imply $M_t^a/t\to0$
almost surely. Since $H_t/t\to0$ and $T<\infty$ almost surely,
\begin{equation}
L_t^a/t\longrightarrow I_a\quad\hbox{almost surely for every }a\in\mathcal A.
\label{eq:bundle-llr-rate}
\end{equation}

For total cost, define its predictable compensator $K_t$ by summing the conditional expected round costs,
\[
K_t=\sum_{s=1}^t\sum_g\mu_{\nu,B_s}(g)
[c_{B_s}^F(g)+\rho_{B_sg}(\hat\nu_{s-1})c_{B_sg}^Y].
\]
Comparing the counts and purchase rules in the same way gives a finite $K_1$ such that
\begin{equation}
|K_t-t\bar c|\le K_1(1+T+H_t).
\label{eq:bundle-cost-deviation}
\end{equation}
This bound relates accumulated conditional cost to the optimal cost per round. Its dependence on $T$ and the cumulative exploration weight $H_t$ also allows us to control expected cost at the random stopping time.

\subsection{Correctness and the stopping tail}

Last, we combine the preceding allocation and rate results to prove uniform $\delta$-correctness, finite expected stopping time, and asymptotic attainment of the characteristic cost in Theorem \ref{thm:bundle-tracker}.

\begin{proof}[Proof of Theorem~\textcolor{.}{5}]
First consider validity. For each $m\ne\nu$,
$\exp\{\ell_t(m)-\ell_t(\nu)\}$ is a nonnegative martingale with initial
value one, because the policy kernels cancel and the observed densities have
common support. Ville's inequality and a union bound give
\[
\Prob_\nu\{\nu\in\mathcal C_t\text{ for every }t\ge0\}\ge1-\delta.
\]
An incorrect recommendation requires coverage failure. Finite stopping, proved below, therefore gives unconditional $\delta$-correctness.

For the unconditional stopping bound, write $\tau_\beta$ for threshold $\beta$. If $L_t^a>\beta$ for every $a\in\mathcal A$, all models with a different best configuration leave the nonempty $\mathcal C_t$, since maximum log likelihood is at least $\ell_t(\nu)$. Hence
\begin{equation}
\{\tau_\beta>t\}\subseteq
\bigcup_{a\in\mathcal A}\{L_t^a\le\beta\}.
\label{eq:bundle-stop-tail-event}
\end{equation}
Fix $\xi>0$ and set
$b_\xi=d-d/(1+\xi)>0$. Choose $\eta>0$ so that
$K_0\eta\le b_\xi/4$. Since $H_t=o(t)$, there is a deterministic
$t_\xi$ such that $K_0(1+H_t)\le b_\xi t/4$ for $t\ge t_\xi$.
For such $t$, on $\{T\le\eta t\}$, \eqref{eq:bundle-drift-deviation}
implies $D_t^a\ge (d-b_\xi/2)t$. If also
$t\ge(1+\xi)\beta/d$, then $\beta\le d t/(1+\xi)$, so
$D_t^a-\beta\ge b_\xi t/2$. Azuma's inequality applied to $M_t^a$ now
gives
\begin{equation}
\Prob_\nu(\tau_\beta>t)
\le\Prob_\nu(T>\eta t)
+|\mathcal A|\exp\left\{-\frac{b_\xi^2 t}{32L^2}\right\},
\quad t\ge\max\{t_\xi,(1+\xi)\beta/d\}.
\label{eq:bundle-stopping-tail}
\end{equation}
Constants are independent of $\beta$. By \eqref{eq:bundle-last-error-tail}, the bound is summable in $t$, giving $\E_\nu\tau_\beta<\infty$ at every finite threshold. Bounded round costs then give finite expected cost, completing the first claim.

For the limit of expectations, let
$n_\beta=\lceil\max\{t_\xi,(1+\xi)\beta/d\}\rceil$ and sum
\eqref{eq:bundle-stopping-tail} over $t\ge n_\beta$.
The tail of a fixed summable sequence tends to zero, so
\begin{align*}
&\E_\nu[\tau_\beta]\le n_\beta+o(1),\\
&\limsup_{\beta\to\infty}\frac{\E_\nu[\tau_\beta]}{\beta}
\le\frac{1+\xi}{d}.
\end{align*}
Letting $\xi\downarrow0$ proves
\begin{equation}
\limsup_{\beta\to\infty}\E_\nu[\tau_\beta]/\beta\le S.
\label{eq:bundle-stopping-mean}
\end{equation}

The cost incurred on each round is bounded and its conditional expectation
is the corresponding increment of $K_t$. Since
$\E_\nu[\tau_\beta]<\infty$, summing these conditional expectations on
$\{\tau_\beta\ge s\}$ gives
\[
\E_\nu[\operatorname{Cost}_{\mathcal B}(\tau_\beta)]
=\E_\nu [K_{\tau_\beta}].
\]
Moreover, $H_t\le1+t^{1-\gamma}/(1-\gamma)$. Jensen's inequality,
\eqref{eq:bundle-cost-deviation}, and $\E_\nu [T]<\infty$ give
\[
\left|\E_\nu[\operatorname{Cost}_{\mathcal B}(\tau_\beta)]
-\bar c\,\E_\nu[\tau_\beta]\right|
\le K_1\left[2+\E_\nu [T]+
\frac{(\E_\nu[\tau_\beta])^{1-\gamma}}{1-\gamma}\right]
=o(\beta).
\]
Equations~\eqref{eq:bundle-optimal-rates} and
\eqref{eq:bundle-stopping-mean} therefore give
$\limsup_{\beta\to\infty}\E_\nu[\operatorname{Cost}_{\mathcal B}(\tau_\beta)]/\beta
\le\bar c S=C_{\rm bun}^\star$.
The proved uniform correctness and termination under every model permit Theorem~\textcolor{.}{3} to give the matching lower bound at $\beta=\beta_\delta$, since
$\kl(1-\delta,\delta)/\beta_\delta\to1$. It follows that
\[
\frac{\E_\nu[\operatorname{Cost}_{\mathcal B}(\tau_\delta)]}{\beta_\delta}
\longrightarrow C_{\rm bun}^\star.
\]
The compensator estimate and $\bar c>0$ then also give
$\E_\nu[\tau_\delta]/\beta_\delta\to C_{\rm bun}^\star/\bar c=S$.
Last, for the fixed family after merging equivalent models,
$\beta_\delta/\log(1/\delta)\to1$. This proves both limits in the theorem. \halmos
\end{proof}

\paragraph{Other finite decision objectives.}

The proof of Theorem 5 does not rely on the decision objective being the identity of the best configuration. More generally, suppose that each candidate model
$\nu$ determines a finite-valued decision $a(\nu)$, such as the best feasible configuration, a feasibility decision, or a discretized service-level or quantile classification. Replace the alternative set $\mathcal A_\nu$ by
\begin{align*}
\mathcal A_\nu^a=\{m\in \mathcal M_0:a(m)\ne a(\nu)\},
\end{align*}
and stop when all models remaining in the likelihood confidence set imply the same decision. Models with identical observation laws must imply the same decision. With these substitutions, the likelihood, allocation, and cost arguments remain unchanged.
\end{revblock}

\begin{revblock}
\section{Delayed Audit Feedback}
\label{sec:delays}

This section extends the PP-LUCB confidence sequences to delayed audit feedback by adding a correction for audits that have been submitted but have not returned. The resulting intervals use only returned verified outcomes while accounting for the possible contribution of pending audits. Section~\ref{EC-sec:queue-service} considers a separate review-queue experiment and uses a different confidence sequence based on completed submission prefixes. It likewise relies only on returned outcomes for certification.

\subsection{Delay model and returned filtration}

We first formalize delayed audit feedback and the information available to the deployed policy at each calendar time. This distinction between submitted and returned audits is needed to construct confidence intervals that remain valid when pending outcomes are unavailable.

When an audit is submitted at round $s$, let $D_s\in\{0,1,2,\ldots\}$ be its
delay. The verified outcome is then available at round $s+D_s$. Write
\[
B_s(t)=\mathbf 1\{A_s=1,\ s+D_s\le t\},
\qquad
Z_s^{\rm del}(t)=
\begin{cases}Y_s,&B_s(t)=1,\\ \dagger,&B_s(t)=0.
\end{cases}
\]
For the filtration argument, let $\mathcal H_t$ denote the full history
that contains the latent submitted records
$\{(k_s,X_s,F_s,A_s,\pi_s,Y_s,D_s):s\le t\}$, where we set $D_s=0$
when $A_s=0$. The returned-outcome filtration $\{\mathcal G_t^{\rm del}\}$
is generated by the observable arm, instance, proxy, audit indicator and
logged propensity records through $t$, together with the return-status history
$\{(B_s(u),Z_s^{\rm del}(u)):1\le s\le u\le t\}$. Thus $A_s$ is
recorded when the audit coin is drawn, while $Y_s$ is revealed only when
$B_s(u)=1$. We have $\mathcal G_t^{\rm del}\subseteq\mathcal H_t$ and
a pending record remains in the history when it later returns.

The deployed policy is adapted to $\mathcal G_t^{\rm del}$. The arm is chosen
from $\mathcal G_{s-1}^{\rm del}$ before the current draw and, after
$(X_s,F_s)$ is observed, $\pi_s$ is measurable with respect to
$\mathcal G_{s-1}^{\rm del}\vee\sigma(X_s,F_s)$. Conditional on
$\mathcal H_{s-1}$ and the selected arm, $(X_s,F_s,Y_s)$ has the arm law
and is fresh, and $A_s$ is conditionally independent of $Y_s$ given
$\mathcal H_{s-1},X_s,F_s$ with propensity at least $p>0$. Thus workload may guide future choices while current outcomes remain unobserved at audit decisions. The pathwise validity proof allows delays to depend on past information, current state and outcome. Stopping additionally requires $D_s\le D_{\max}$ almost surely.

For arm $k$, let $P_k(t)$ count pending audits at $t$, and let $w_{F,k}(n)$ and $w_{R,k}(n)$ be the no-delay proxy and residual widths at pull count $n$. We retain all configuration evaluations in the denominator and bound the missing residual contribution from pending audits. This accounts for return times that can depend on the outcomes. The returned residual estimator and width are
\begin{align}
\widehat\mu_{R,k}^{\,\rm del}(t)&=
\frac{1}{N_k(t)}\sum_{s\le t:k_s=k}B_s(t)\frac{A_s}{\pi_s}(Y_s-F_s),\nonumber\\
w_{R,k}^{\rm del}(t)&=w_{R,k}(N_k(t))+\frac{P_k(t)}{pN_k(t)}.
\label{eq:returned-width}
\end{align}

\begin{theorem}
\label{thm:delay_cs}
Under \textcolor{.}{Assumptions~1 and~2} and the fresh-draw condition stated above, suppose the arm choice precedes
the current draw, the audit coin is conditionally independent of the current
verified outcome with $\pi_s\ge p$, and the policy is adapted to
$\{\mathcal G_t^{\rm del}\}$. Let
$\mu_{F,k}=\E[F\mid k]$ and $\mu_{R,k}=\E[Y-F\mid k]$, and, for
$N_k(t)\ge1$, define the returned center and endpoints by
\begin{align*}
\widehat\theta_k^{\rm del}(t)
   &\coloneqq \widehat\mu_{F,k}(t)+\widehat\mu_{R,k}^{\rm del}(t),\\
L_k^{\rm del}(t)
   &\coloneqq \widehat\theta_k^{\rm del}(t)-w_{F,k}(N_k(t))-w_{R,k}^{\rm del}(t),\\
U_k^{\rm del}(t)
   &\coloneqq \widehat\theta_k^{\rm del}(t)+w_{F,k}(N_k(t))+w_{R,k}^{\rm del}(t).
\end{align*}
Here the no-delay widths use error budgets
$\delta_{F,k}=\delta_{R,k}=\delta/(2K)$. Then the endpoints are
$\mathcal G_t^{\rm del}$-measurable and
\[
\Pbb\!\left(\forall t\ge1\ \text{with }N_j(t)\ge1\ \forall j\in[K],\ \forall k\in[K]:
\theta_k\in[L_k^{\rm del}(t),U_k^{\rm del}(t)]\right)\ge1-\delta.
\]
Consequently, the corresponding separation rule satisfies
$\Pbb(\tau_{\rm del}<\infty,\widehat k\ne k^\star)\le\delta$.
\end{theorem}

Note that Theorem \ref{thm:delay_cs} extends the confidence guarantee in Theorem 2 to settings in which requested audits may return only after later evaluations have occurred. The additional pending-audit term widens the residual interval to account for information that has been requested but remains unavailable for inference.

\end{revblock}

\begin{revblock}
\subsection{Proof of Theorem~\ref{thm:delay_cs}}
Fix an arm $k$ and put $I_{k,s}=\mathbf 1\{k_s=k\}$. In the full
history, the fresh-draw and audit-independence conditions give
\[
\E\!\left[I_{k,s}\left\{\frac{A_s}{\pi_s}(Y_s-F_s)-\mu_{R,k}\right\}
\middle|\mathcal H_{s-1}\right]=0.
\]
The residual increments on selected rounds have range of length $2/p$.
The calendar-time exponential supermartingale in the proof of
\textcolor{.}{Proposition~2} therefore remains valid on this full
history, with variance clock $N_k(t)/p^2$. The proxy calculation has clock
$N_k(t)/4$. Applying their confidence sequences and the union bound gives,
with probability at least $1-\delta$, simultaneously for every calendar time
with $N_k(t)\ge1$,
\begin{align*}
&\left|R_k(t)/N_k(t)-\mu_{R,k}\right|
\le w_{R,k}(N_k(t)),\\
&R_k(t)=\sum_{s\le t:k_s=k}\frac{A_s}{\pi_s}(Y_s-F_s).
\end{align*}

Write $\pi_{\min}=p$, $M=2/p$ and $\delta_k=\delta/(2K)$, and define
$S_{R,k}(N_k(t))=R_k(t)-N_k(t)\mu_{R,k}$. The preceding martingale
argument gives, with residual error probability at most $\delta_k$,
\begin{equation}
\bigl|S_{R,k}(N_k(t))\bigr|
\le\psi\!\left(\frac{N_k(t)M^2}{4};\delta_k\right)
\label{eq:delay-submission-cs}
\end{equation}
simultaneously at every $t$ with $N_k(t)\ge1$.

{\color{black}
The oracle estimator $\widehat{\mu}_{R,k}(t) = \frac{1}{N_k(t)}\sum_{s \le t:\, k_s=k} \frac{A_s}{\pi_s}(Y_s - F_s)$ uses all outcomes, while the delayed estimator $\widehat{\mu}_{R,k}^{\,\mathrm{del}}(t)$ uses only returned outcomes.
Their relationship gives the decomposition
\[
S_{R,k}(N_k(t))
= N_k(t)\bigl(\widehat{\mu}_{R,k}^{\,\mathrm{del}}(t) - \mu_{R,k}\bigr)
\;+\; \underbrace{\sum_{\substack{s \le t:\, k_s = k,\\ A_s = 1,\, B_s(t) = 0}} \frac{1}{\pi_s}(Y_s - F_s)}_{\displaystyle =:\; Q_k(t)\;\text{(pending sum)}},
\]
where $Q_k(t)$ aggregates the IPW residuals of audits \rev{requested but still pending}.

Each pending term satisfies $\bigl|\frac{1}{\pi_s}(Y_s - F_s)\bigr| \le 1/\pi_{\min}$
(since $|Y_s - F_s| \le 1$ and $\pi_s \ge \pi_{\min}$), so
\[
|Q_k(t)| \le \frac{P_k(t)}{\pi_{\min}},
\]
where $P_k(t) = \bigl|\{s \le t : k_s = k,\, A_s = 1,\, B_s(t) = 0\}\bigr|$ counts the pending audits for arm~$k$ at time~$t$. By the triangle inequality and~\eqref{eq:delay-submission-cs}, on the event that the submission-time CS holds\rev{,}
\begin{align*}
\bigl|N_k(t)\bigl(\widehat{\mu}_{R,k}^{\,\mathrm{del}}(t) - \mu_{R,k}\bigr)\bigr|
&= \bigl|S_{R,k}(N_k(t)) - Q_k(t)\bigr| \\
&\le \bigl|S_{R,k}(N_k(t))\bigr| + |Q_k(t)| \\
&\le \psi\!\left(\frac{N_k(t)\, M^2}{4};\, \rev{\delta_k}\right) + \frac{P_k(t)}{\pi_{\min}}.
\end{align*}
Dividing both sides by $N_k(t)$\rev{, we obtain}
\[
\bigl|\widehat{\mu}_{R,k}^{\,\mathrm{del}}(t) - \mu_{R,k}\bigr|
\le \underbrace{\frac{\psi(N_k(t)\, M^2/4;\, \rev{\delta_k})}{N_k(t)}}_{=\; w_{R,k}(N_k(t))}
+ \frac{P_k(t)}{\pi_{\min}\, N_k(t)}
\;=\; w_{R,k}^{\,\mathrm{del}}(t).
\]
The delayed estimator $\widehat{\mu}_{R,k}^{\,\mathrm{del}}(t)$ is computed from returned outcomes, and the pending count $P_k(t)$ is determined by which audits have been requested and which have returned. Both are $\rev{\mathcal G_t^{\rm del}}$-measurable quantities. When there are no delays ($D_s = 0$ for all $s$), every audit returns immediately, so $P_k(t) = 0$ and $\widehat{\mu}_{R,k}^{\,\mathrm{del}}(t) = \widehat{\mu}_{R,k}(t)$. This recovers \textcolor{.}{Proposition~2}.
}

The proxy confidence sequence has the same armwise error allocation
$\delta_k=\delta/(2K)$. A union bound over these $2K$ sequences gives
simultaneous coverage with probability at least $1-\delta$.
On that event, a separated returned interval recommends the unique best arm.

For a baseline learned from past returned records, retain $m_{k,s}$ at its
submission-time value and use the returned corrected center
\[
\frac{1}{N_k(t)}\sum_{s\le t:k_s=k}
\left[m_{k,s}+B_s(t)\frac{A_s}{\pi_s}(Y_s-m_{k,s})\right].
\]
Since $m_{k,s},Y_s\in[0,1]$, its difference from the complete AIPW center
is at most $P_k(t)/(pN_k(t))$. Adding this term to the width for one configuration in
\eqref{eq:aipw-cs-width} therefore gives the same returned-data guarantee for
online fitted baselines.
\end{revblock}
\halmos

\begin{revblock}
\begin{theorem}
\label{thm:delay_stop}
Suppose the intervals are the $[L_k^{\rm del}(t),U_k^{\rm del}(t)]$ from
Theorem~\ref{thm:delay_cs}, one arm is selected per calendar round, and there
is a constant $D_{\max}<\infty$ such that $D_s\le D_{\max}$ almost
surely. Suppose the continuation that omits the stopping action satisfies
$N_k(t)\to\infty$ almost surely for every $k$. The sparse-exploration
schedule in Algorithm~\textcolor{.}{2} has this property.
If the verified means have a unique best arm, so that
$\Delta_k=\theta_{k^\star}-\theta_k>0$ for $k\ne k^\star$, define
\[
\tau_{\rm del}=\inf\left\{t\ge1:
N_j(t)\ge1\ (j\in[K]),\ \exists k\in[K],\quad
L_k^{\rm del}(t)>\max_{j\ne k}U_j^{\rm del}(t)\right\}.
\]
Then $\tau_{\rm del}$ is a stopping time for
$\{\mathcal G_t^{\rm del}\}$, and the recommendation at
$\tau_{\rm del}$ is $\delta$-correct and finite almost surely.
\end{theorem}

Note that Theorem~\ref{thm:delay_stop} converts the delayed-feedback coverage result in Theorem~\ref{thm:delay_cs} into a stopping guarantee. Under bounded audit delays and continued exploration, the pending-audit correction vanishes as each arm accumulates observations, so the procedure eventually certifies the unique best arm with the stated confidence.
\end{revblock}

\begin{revblock}
\subsection{Proof of Theorem~\ref{thm:delay_stop}}
The endpoints and the event in the definition of $\tau_{\rm del}$ are
$\mathcal G_t^{\rm del}$-measurable. The first time in the displayed domain
is therefore a stopping time. On the simultaneous coverage event from
Theorem~\ref{thm:delay_cs}, if the rule stops with leader $\widehat k$, then
\[
\theta_{\widehat k}\ge L_{\widehat k}^{\rm del}(t)
>\max_{j\ne\widehat k}U_j^{\rm del}(t)
\ge\max_{j\ne\widehat k}\theta_j,
\]
so $\widehat k=k^\star$.

One arm per round and $D_s\le D_{\max}$ leave at most $D_{\max}$ pending audits, giving $P_k(t)\le D_{\max}$. Exploration gives $N_k(t)\to\infty$ for every arm, so the pending correction vanishes. No-delay widths vanish since $\psi(v;\delta)/v\to0$ as $v\to\infty$. The strong law for bounded pull-indexed martingale increments gives convergence of proxy and residual centers. Hence $\widehat\mu_{R,k}^{\rm del}(t)\to\mu_{R,k}$ almost surely. Returned centers converge to $\theta_k$, so positive gaps give finite separation almost surely. The deployed run agrees until separation, selecting the unique best arm on the coverage event with error at most $\delta$.
\end{revblock}
\halmos

\begin{revblock}
\begin{remark}
The delayed-feedback guarantees concern inference based on returned information and the resulting stopping rule. A reference-acquisition cost is incurred when an audit is submitted, while its statistical contribution becomes available only when the audit returns. Under $D_s\le D_{\max}$, the pending-audit correction is $O(D_{\max}/(pN_k(t)))$, compared with a confidence width of $O\!\left(\sqrt{\frac{\log\log N_k(t)+\log(1/\delta)}{N_k(t)}}\right)$. Thus, bounded delays do not change the cost of an individual submitted audit, but they can delay stopping and therefore change subsequent configuration choices, the number of submitted audits, and total evaluation cost.
\end{remark}
\end{revblock}

\begin{revblock}
\section{Practical Considerations for Service Evaluation}
\label{sec:design-implications}

This section draws practical implications for implementing LLM-assisted service evaluation and then describes extensions of the selection framework. We first discuss the data-recording and audit-coverage requirements for valid selective-audit inference and the use of residuals between audited outcomes and judge scores for monitoring, and then extend the framework to context-dependent configuration choices and to alternative finite decision objectives, including feasibility-constrained and quantile-based selection.

{\color{black}
From a managerial perspective, our results suggest some practical recommendations for LLM-assisted service systems.

First, managers should record the audit probability as part of the data record. Once audits are
triggered after observing the instance and proxy score, the audit probability $\pi_t$ should be logged
together with $(k_t,X_t,F_t,A_t,\rev{A_tY_t})$. This is important because audited outcomes are a selected subset, so
the selection rule must be visible in the data in order for inverse propensity weighting to correct it.
\rev{In practice, the data record should include the rule used to compute $\pi_t$ and any changes to that rule. This allows the logged probability to be checked against the information available when the audit decision was made.}

Second, managers should keep a minimum audit rate within each segment. A small minimum
audit probability (such as the $\rev{p}$ in our setup) prevents segments from becoming never audited
where the proxy-to-audit gap remains unknown. The key point is that a global average audit rate
is not enough if some segments are effectively skipped. A practical approach is to define segments
using observables available at decision time, such as task type, language, customer tier, or risk tier,
and then apply the minimum audit rate within each segment. This stabilizes the learning problem
by making residual information available throughout the system's operating regions.

Third, managers can use the residual $Y-F$ as a routine judge-audit discrepancy measure for the
LLM judge. Segment-level summaries of the residual, including its mean and variance, provide an
interpretable view of when the judge is helping and when it is misleading. More broadly, $Y-F$
can be monitored as a key health metric, much like prediction error is tracked in forecasting and
calibration error is tracked in measurement and risk scoring systems.
\rev{These summaries must correct for selective auditing using the logged purchase probabilities.}

In this research, we treat audited outcomes as ground truth, but in practice, human audits can
be noisy and may vary among reviewers. It is a promising future research direction to extend the
framework to handle multiple reviewers and disagreements, which would better match how auditing
works in reality. In addition, it would be useful to study how to model and correct for systematic
auditor preferences (e.g., favoring certain writing styles or conservative decisions) that may induce
structured deviations in human-audited outcomes.
}

\subsection{Context-first service decisions}
\label{subsec:context-first}

We first consider settings in which observable request context is available before a configuration is selected. This extension allows the policy to adapt configuration choices to context while preserving inference for a prespecified population-level performance target.

Let $C_t\in\{1,\ldots,J\}$ be iid contexts, independent of the previous
history, with declared target mix $\lambda_j=\Pbb(C_t=j)>0$. Within context
$j$, define $\theta_{k,j}=\E[Y(k,X)\mid C=j]$. The chosen bundle may depend
on the history and $C_t$. Conditional on that history, $C_t=j$ and the
selected bundle $B_t=B$, the remaining instance, proxies and verified
outcomes are a fresh draw from the declared conditional law for $(j,B)$.
The mean must remain the same across bundles within each context, so we impose
\[
 \E[Y_{k,t}\mid\mathcal G_{t-1},C_t=j,B_t=B]=\theta_{k,j},
 \qquad k\in B.
\]
Include all request information used for configuration choice in $C_t$. Auditing may also use generated text and proxies. Within each class, apply the corrected-score confidence sequence with error budget $\delta/(KJ)$ per pair $(k,j)$.
If its resulting interval is $[L_{k,j}(t),U_{k,j}(t)]$, then
\begin{equation}
\theta_k=\sum_{j=1}^J\lambda_j\theta_{k,j}
\in\left[\sum_{j=1}^J\lambda_jL_{k,j}(t),\,
\sum_{j=1}^J\lambda_jU_{k,j}(t)\right]
\label{eq:context-first-ci}
\end{equation}
simultaneously over configurations and time with probability at least
$1-\delta$. Before a pair has been sampled, use its known range $[0,1]$.
To verify the claim, write
$Z_{k,t}=F_{k,t}+V_t(Y_{k,t}-F_{k,t})/\rho_t$ when $k\in B_t$.
The audit identity first averages over the audit coin conditional on the
complete pre-audit state and verified outcome. Averaging over the fresh
conditional draw then gives
\[
 \E[Z_{k,t}\mid\mathcal G_{t-1},C_t=j,B_t=B]
 =\E[Y_{k,t}\mid\mathcal G_{t-1},C_t=j,B_t=B]
 =\theta_{k,j}.
\]
The bounded corrected-score argument used in Proposition~\ref{prop:aipw-cs} therefore applies to the observations selected for each configuration-context pair, even when the selected bundle varies with the history. A union bound
covers all $KJ$ pairs. Multiplication by the nonnegative fixed weights and
summation give \eqref{eq:context-first-ci}.

For stopping, fix finitely many admissible bundles covering every configuration in each context. Cycle through them on square-numbered arrivals of that context. Since $\lambda_j>0$, each context arrives infinitely often almost surely, and every pair count diverges on a nonstopping continuation. The bounded martingale strong
law gives consistency of every pair estimate, and the confidence widths
vanish. The weighted endpoints consequently converge to $\theta_k$ almost
surely. A unique best configuration then gives finite stopping almost
surely, while the simultaneous coverage event gives error probability at
most $\delta$.

The prespecified population weights \(\lambda_j\) preserve the intended overall performance target even when the evaluation policy samples contexts at unequal rates. In particular, for a context-dependent routing rule
\(
r:\{1,\ldots,J\}\to[K],
\)
its population value is
\(
V(r)=\sum_{j=1}^{J}\lambda_j\theta_{r(j),j}.
\)
A simultaneous confidence interval for \(V(r)\) is obtained by aggregating the corresponding context-specific endpoints,
\[
\left[
\sum_{j=1}^{J}\lambda_j L_{r(j),j}(t),\;
\sum_{j=1}^{J}\lambda_j U_{r(j),j}(t)
\right].
\]
This construction assumes a finite, prespecified context set. Treating unrestricted text itself as the context would instead require a model for conditional performance as a function of text.

\subsection{Other service selection objectives}
\label{app:service-answer-map}

We next show that our framework is not limited to selecting the configuration with the highest mean service score. By redefining the decision target over the finite model family, it can also support feasibility-constrained selection and other finite-valued objectives, such as quantile-based decisions.

We fix the target population, judge and configurations and assume that their joint evaluation law remains unchanged during the evaluation. Changing them requires a new evaluation. The candidate joint laws form $\mathcal M_0$, which must contain the true law for the likelihood guarantee. The confidence sequences for bounded means do not require this finite family.

For service-level requirements, choose the best feasible configuration or report none. Suppose verification gives a service score and compliance indicators. The acquired reference must verify all scores and indicators used in the decision, and its acquisition cost must include this verification. Write $\gamma_{\nu,kj}$ for indicator $j$'s mean, with required level $b_j$, and define
\[
 \mathcal K_{\rm feas}(\nu)=
 \{k:\gamma_{\nu,kj}\geq b_j\text{ for all }j\}.
\]
For this objective, let $a(\nu)$ denote the highest-scoring feasible
configuration, with a specified tie convention, and set $a(\nu)=\varnothing$
if none is feasible. A quantile objective similarly defines $a(\nu)$ from the
quantiles of the verified outcome distributions. The relevant alternative
set is then $\{\nu':a(\nu')\ne a(\nu)\}$.

\begin{proposition}
\label{prop:service-answer-map}
Let $a$ be a specified map from the finite model family to a finite answer
set. Suppose that models with different answers have different observation
laws for at least one admissible bundle. Under the support and cost conditions of
Theorem~\ref{thm:bundle-tracker}, replace best-configuration alternatives
by models with a different value of $a$. The resulting joint program is a
lower bound for uniformly $\delta$-correct identification of $a(\nu)$.
The corresponding joint allocation algorithm is $\delta$-correct and attains this bound as
$\delta\downarrow0$.
\end{proposition}

\begin{proof}
Fix a true model $\nu$, and define the decision-changing alternative set
\[
\mathcal{A}^a_\nu
:=
\left\{
\nu' \in \mathcal{M}_0 :
a(\nu') \neq a(\nu)
\right\}.
\]
Define $C^\star_{\mathrm{bun}}(\nu,\mathcal{A}^a_\nu)$ by the linear
program in \eqref{eq:bundle-characteristic-cost}, replacing the
best configuration alternative set by $\mathcal{A}^a_\nu$. If every model gives the same answer, this alternative set is empty and the zero allocation is optimal. Returning the common answer at time zero has zero cost and proves both claims in this case. We therefore consider a model family with at least two different answers, so every model has a nonempty alternative set. By assumption,
every $\nu' \in \mathcal{A}^a_\nu$ differs from $\nu$ in its observable law
under at least one admissible bundle. Hence the corresponding information
program is feasible under the support and cost conditions of Theorem~5.

For the lower bound, let $\widehat a$ denote the stopped answer and consider
the event
\(
E := \{\widehat a = a(\nu)\}.
\)
Uniform $\delta$-correctness gives
\[
\mathbb{P}_{\nu}(E) \geq 1-\delta
\text{  and  }
\mathbb{P}_{\nu'}(E) \leq \delta
\quad\text{for every } \nu' \in \mathcal{A}^a_\nu.
\]
Applying binary data processing to $E$, followed by the stopped
likelihood-ratio decomposition in the proof of Theorem~3, gives the same
information constraints as in \eqref{eq:bundle-characteristic-cost}, now for
every $\nu' \in \mathcal{A}^a_\nu$. Therefore,
\[
\mathbb{E}_{\nu}\!\left[\operatorname{Cost}_{\mathcal B}(\tau)\right]
\geq
\operatorname{kl}(1-\delta,\delta)\,
C^\star_{\mathrm{bun}}(\nu,\mathcal{A}^a_\nu).
\]

For achievability, first merge models with identical observation laws for every bundle. Such models give the same answer by assumption and have identical information constraints, so this step preserves the decision problem and its characteristic cost. Solve the same allocation program for each remaining candidate model
using its decision-changing alternative set, and stop at the first time
\[
\tau
:=
\inf\left\{
t \geq 0 :
\left|
\left\{
a(m) : m \in \mathcal{C}_t
\right\}
\right| = 1
\right\}.
\]
At this time, return the unique element of
$\{a(m):m\in\mathcal{C}_\tau\}$. On the event that the true model remains in
the likelihood confidence set at all times, the returned answer is necessarily
$a(\nu)$. The Ville-inequality argument in the proof of Theorem~5 gives this
event probability at least $1-\delta$.

It remains to show finite stopping and attainment of the lower bound. The finite observation spaces, common support, positive audit floor and positive expected configuration-evaluation costs are retained from Theorem~5. As in Section~EC.4.1, they give an optimal allocation $(q^\star,x^\star)$ with $S=\sum_Bq_B^\star>0$. Set $w_B=q_B^\star/S$ and $\rho_{Bg}=x_{Bg}^\star/q_B^\star$ when $q_B^\star>0$, and set $\rho_{Bg}=p$ otherwise. Feasibility and the objective give, for each $m\in\mathcal A^a_\nu$,
\[
 I_m\ge 1/S,
 \qquad
 \bar c=C^\star_{\mathrm{bun}}(\nu,\mathcal A^a_\nu)/S,
\]
where $I_m$ is the expected log likelihood ratio per round against $m$ and $\bar c$ is the expected cost per round under this allocation.

Forced exploration and the audit floor still distinguish all retained models. Lemmas~\ref{lem:bundle-deficit} and~\ref{lem:bundle-model-lock} therefore give the same tracking bound and finite moments for the last time an incorrect model is selected. The likelihood and cost bounds in Section~EC.4.3 then hold with the new optimal allocation. In particular, if the log likelihood ratio against every $m\in\mathcal A^a_\nu$ exceeds the threshold $\beta_\delta$, all remaining models give $a(\nu)$ and the displayed stopping rule has stopped. Thus the stopping-tail calculation in \eqref{eq:bundle-stopping-tail} applies to precisely this alternative set and gives
\[
 \mathbb E_\nu[\tau_\delta]<\infty,
 \qquad
 \limsup_{\delta\downarrow0}
 \frac{\mathbb E_\nu[\tau_\delta]}{\beta_\delta}\le S.
\]
The cost-compensator argument in Section~EC.4.4 consequently gives
\begin{align*}
& \mathbb E_\nu\!\left[\operatorname{Cost}_{\mathcal B}(\tau_\delta)\right]
 =\bar c\,\mathbb E_\nu[\tau_\delta]+o(\beta_\delta),\\
&\limsup_{\delta\downarrow0}
 \frac{\mathbb E_\nu[\operatorname{Cost}_{\mathcal B}(\tau_\delta)]}
 {\beta_\delta}
 \le C^\star_{\mathrm{bun}}(\nu,\mathcal A^a_\nu).
\end{align*}
Finite stopping and the coverage event establish uniform $\delta$-correctness. Combining the upper bound with the lower bound already proved, and using $\operatorname{kl}(1-\delta,\delta)/\beta_\delta\to1$, proves attainment of the characteristic cost. \halmos
\end{proof}

Proposition~\ref{prop:service-answer-map} here shows that our framework is not limited to selecting the
configuration with the largest verified mean. In particular, feasibility,
service-level, and discretized quantile decisions can be handled by defining
the relevant finite-valued decision map $a(\nu)$ and allocating evidence
against only those models that imply a different decision.

\end{revblock}

\begin{revblock}
\section{Complete Synthetic Selection Experiments}
\label{ec:new-synthetic-experiments}
This section provides the settings for the experiment on the finite-model example in Section~\ref{subsec:joint-tracker-experiment} of the main paper. We first specify the observation models and comparison procedures, including the confidence-level experiment reported there, and then examine how costs affect the value of sharing references. Additional experiments at short horizons and with larger configuration sets assess when selection can be certified and whether the cost benefit extends beyond three configurations.

\subsection{Candidate models and information available before auditing}
\label{ec:new-library}
We first specify the finite-model example used in Section~\ref{subsec:joint-tracker-experiment} of the main paper. This example examines how sharing a reference and using the observed state affect selection cost. We use $K=M=3$ service configurations and candidate models for their observations. Each model assigns equal probability to pre-audit states $G\in\{0,1\}$. The conditional Bernoulli success probabilities are
\[
\begin{array}{c|ccc}
 & k=1 & k=2 & k=3\\\hline
\nu^0 & (.60,.60)&(.57,.57)&(.54,.54)\\
\nu^1 & (.60,.60)&(.65,.63)&(.54,.54)\\
\nu^2 & (.60,.60)&(.57,.57)&(.70,.54).
\end{array}
\]
Each pair gives the probabilities in states 0 and 1. The proxy pairs $(.55,.60)$, $(.85,.75)$ and $(.80,.90)$ are common across models. Thus the proxy scores alone cannot distinguish the candidate models, but the observed state determines how informative an audit is. This separates the value of choosing when to audit from direct prediction of the best configuration.

Writing $p_{mkg}$ for a displayed probability, independent outcomes satisfy $Y_k=\mathbf1\{U_k\le p_{mkg}\}$ with independent uniform $U_k$. Correlated outcomes use a common uniform with probability $.6$ and independent uniforms with probability $.4$. The mixing indicator is unobserved, and
\[
Q_{mg}(y)=.4\prod_{k=1}^3p_{mkg}^{y_k}(1-p_{mkg})^{1-y_k}
+.6\Pr\{(\mathbf1\{U\le p_{mkg}\})_{k=1}^3=y\}.
\]
The independent component gives every binary vector positive probability. Retaining each bundle's coordinates preserves configuration marginals across bundles. Under $\nu^0$, full-vector KL against $\nu^1$ is $(.01365486,.00757378)$ for independent outcomes and $(.02632727,.01615353)$ for correlated outcomes. Against $\nu^2$, it is $(.05648820,0)$ and $(.08687090,0)$. The marginal-KL sums equal the independent values.

Order the seven nonempty bundles by cardinality, then lexicographically. Their configuration-evaluation and reference-acquisition costs are $c_F|B|$ and $c_Y$. Each LP uses exact projected probabilities and all alternatives with different best configurations. At round $t$, the current MLE supplies $w_{\widehat\nu}$. With floor $.1$, add $(1-t^{-.6})w_{\widehat\nu}+t^{-.6}/|\mathcal B|$ to the cumulative target and select the largest deficit. Ties use the first index. After observing $g$, audit with probability $x_{Bg}/q_B$ for positive $q_B$ and the floor otherwise. Stop when all models retained at likelihood threshold $\log(2/\delta)$ have the same best configuration.

\subsection{Comparisons and repeated paths}
\label{ec:new-protocol}
We next describe the comparison procedures and repeated paths for the experiment in Section~\ref{subsec:joint-tracker-experiment}. To identify which allocation choices account for cost differences, six procedures share the model family, stopping rule and costs. Joint uses the full program. Singleton restricts evaluations and exploration to one configuration per round, distinct from PP-LUCB. Fixed audit retains all bundles but sets $\rho_{B0}=\rho_{B1}$. Here its positive-mass actions audit fully and zero-mass exploration actions use the floor. Full audit always audits. Full bundle evaluates $\{1,2,3\}$ and updates audit probabilities from the current model estimate. Oracle uses the true model's LP solution with exploration and likelihood stopping.

The grid combines two outcome laws, three true models and six procedures, each using 1,000 paths at $c_F=1$, $c_Y=10$, $\delta=.05$, capped at 20,000 rounds. All paths stopped within the 20,000-round cap, and total costs include all evaluations and all requested audits, including incorrect selections. Table~\ref{tab:new-main-cost} of the main paper reports four procedures under $\nu^0$. Tables~\ref{tab:new-tracker-independent} and~\ref{tab:new-tracker-correlated} add the remaining comparisons under $\nu^0$ and results under $\nu^1$ and $\nu^2$. They report means, standard errors in parentheses and incorrect recommendations out of 1,000 paths. With correlated outcomes, all models have the same joint allocation, so Joint and Oracle give identical records. Their common results are reported once, in the main paper for $\nu^0$ and in Table~\ref{tab:new-tracker-correlated} for the other models. Evaluation and reference counts show how each procedure divides its spending, while error counts report the reliability of the resulting selection. The additional true models check whether the comparisons depend on which configuration is best. Joint costs less than Singleton and Fixed audit under both of these models. Full bundle can have lower cost than Joint, so these finite-confidence comparisons also show why asymptotic optimality does not imply the lowest cost at every confidence level.

\begin{table}[t]
\color{MSRevisionBlue}
\centering\scriptsize
\caption{Selection cost, evaluations, audits and errors with independent outcomes.}
\label{tab:new-tracker-independent}
\begin{tabular}{clrrrr}\toprule
True model & Procedure & Cost (SE) & Evaluations & References & Errors\\\midrule
$\nu^0$ & Full audit & 5141.2 (88.9) & 593.9 & 454.7 & 50\\
$\nu^0$ & Oracle & 4646.7 (80.2) & 893.2 & 375.4 & 37\\
\midrule
$\nu^1$ & Joint & 3560.8 (73.7) & 607.6 & 295.3 & 28\\
$\nu^1$ & Singleton & 3619.7 (83.1) & 569.4 & 305.0 & 27\\
$\nu^1$ & Fixed audit & 4050.3 (88.2) & 429.2 & 362.1 & 16\\
$\nu^1$ & Full audit & 4202.8 (87.6) & 423.7 & 377.9 & 15\\
$\nu^1$ & Full bundle & 4018.9 (84.2) & 1418.1 & 260.1 & 38\\
$\nu^1$ & Oracle & 3520.0 (76.5) & 574.2 & 294.6 & 0\\
\midrule
$\nu^2$ & Joint & 1966.3 (60.7) & 351.2 & 161.5 & 28\\
$\nu^2$ & Singleton & 2178.2 (72.0) & 343.9 & 183.4 & 29\\
$\nu^2$ & Fixed audit & 2541.2 (69.5) & 283.1 & 225.8 & 23\\
$\nu^2$ & Full audit & 2455.3 (68.6) & 257.7 & 219.8 & 22\\
$\nu^2$ & Full bundle & 1196.8 (25.5) & 422.0 & 77.5 & 3\\
$\nu^2$ & Oracle & 965.6 (20.7) & 167.0 & 79.9 & 0\\
\bottomrule\end{tabular}
\end{table}
\begin{table}[t]
\color{MSRevisionBlue}
\centering\scriptsize
\caption{Selection cost, evaluations, audits and errors with correlated outcomes.}
\label{tab:new-tracker-correlated}
\begin{tabular}{clrrrr}\toprule
True model & Procedure & Cost (SE) & Evaluations & References & Errors\\\midrule
$\nu^0$ & Full audit & 2550.3 (48.3) & 570.2 & 198.0 & 35\\
\midrule
$\nu^1$ & Joint / Oracle & 2174.7 (46.6) & 778.8 & 139.6 & 29\\
$\nu^1$ & Singleton & 3758.1 (80.2) & 589.9 & 316.8 & 34\\
$\nu^1$ & Fixed audit & 2262.6 (48.3) & 535.4 & 172.7 & 22\\
$\nu^1$ & Full audit & 2453.6 (52.9) & 548.4 & 190.5 & 15\\
$\nu^1$ & Full bundle & 2227.2 (48.2) & 788.0 & 143.9 & 19\\
\midrule
$\nu^2$ & Joint / Oracle & 765.8 (17.4) & 277.0 & 48.9 & 7\\
$\nu^2$ & Singleton & 1996.6 (61.5) & 315.4 & 168.1 & 29\\
$\nu^2$ & Fixed audit & 1165.3 (26.9) & 278.7 & 88.7 & 7\\
$\nu^2$ & Full audit & 1189.9 (25.8) & 261.6 & 92.8 & 13\\
$\nu^2$ & Full bundle & 733.7 (16.0) & 258.8 & 47.5 & 5\\
\bottomrule\end{tabular}
\end{table}

The confidence-level experiment reported in Figure~\ref{fig:main-confidence-cost} of the main paper examines the approach to the limit in Theorem~\ref{thm:bundle-tracker}. It uses $\nu^0$, independent outcomes and $\delta\in\{.1,.05,.01,.001,10^{-5},10^{-7},10^{-9}\}$. Joint, Singleton, Fixed audit, Full bundle and Oracle each use 1,000 fresh paths per risk level with a 50,000-round cap. All stop. In these experiments, write $C^*_{\rm joint}=C^\star_{\rm bun}$ from Theorem~\ref{thm:bundle-tracker} and $C^*_{\rm sing}$ for its singleton restriction. Each normalized cost is the mean cost divided by $C^*_{\rm joint}\log(2/\delta)$, and error bars are $1.96$ Monte Carlo standard errors. Joint error counts are $67,51,10,2,0,0,0$. The observed $51/1000$ at $\delta=.05$ has Monte Carlo uncertainty. Zero errors in 1,000 paths gives a one-sided 95\% binomial upper limit near $.003$. The cost curves in the main paper and these error counts describe different aspects of the same experiment. The curves assess the approach to the limit of the expected cost, while the counts report selection errors at the prescribed risk levels.

\subsection{Sensitivity to configuration-evaluation and reference-acquisition costs}
\label{ec:new-prices}
\label{subsec:price-sensitivity}
We now give the cost-sensitivity calculation summarized in Section~\ref{subsec:joint-tracker-experiment} of the main paper. It compares the characteristic costs with and without reference sharing, so the difference does not depend on how quickly a procedure learns the true model. This calculation examines when sharing a reference offers the greatest cost saving. The LP is solved for $c_F\in\{1,10,100\}$ and $c_Y\in\{.1,1,10,100\}$ under $\nu^0$ and both outcome laws. Saving is $1-C^*_{\rm joint}/C^*_{\rm sing}$. A common rescaling of prices changes both objectives proportionally. Table~\ref{tab:new-structural} gives the characteristic costs for $c_F=1$, and Figure~\ref{fig:new-prices} reports the relative savings for all 24 cost settings. Savings increase as reference acquisition becomes more expensive relative to configuration evaluation. Correlated outcomes provide further information from the shared reference, explaining their larger savings in this example.

\begin{table}[t]
\color{MSRevisionBlue}
\centering\small
\caption{Characteristic costs across reference-acquisition costs.}
\label{tab:new-structural}
\begin{tabular}{llrrrr}\toprule
Outcome law & Quantity & $c_Y=.1$ & $c_Y=1$ & $c_Y=10$ & $c_Y=100$\\\midrule
Independent & $C^*_{\rm joint}$ & 139.04 & 223.83 & 937.42 & 7806.58\\
Independent & $C^*_{\rm sing}$ & 140.99 & 243.30 & 1132.15 & 9753.89\\
Correlated & $C^*_{\rm joint}$ & 113.94 & 182.34 & 608.39 & 4151.36\\
Correlated & $C^*_{\rm sing}$ & 140.99 & 243.30 & 1132.15 & 9753.89\\
\bottomrule\end{tabular}
\end{table}
\begin{figure}[t]
\color{MSRevisionBlue}
\centering
{\color{black}
\begin{tikzpicture}
\definecolor{plotBlue}{HTML}{0072B2}
\definecolor{plotOrange}{HTML}{D55E00}
\definecolor{plotGreen}{HTML}{009E73}
\begin{axis}[
font=\footnotesize,
axis line style={black},tick style={black},
tick label style={text=black},label style={text=black},
title style={text=black,font=\footnotesize},
legend style={font=\footnotesize,text=black,draw=none,fill=none},
grid=major,grid style={black!12},
scaled ticks=false,
clip marker paths=true,
width=.47\linewidth,height=5.1cm,
xmode=log,log basis x=10,
xmin=.08,xmax=125,
xtick={.1,1,10,100},xticklabels={.1,1,10,100},
xlabel={$c_Y$},ylabel={Saving over Singleton (\%)},
legend pos=north west,legend cell align=left,
name=priceIndependent,ymin=0,ymax=23,ytick={0,5,10,15,20},
title={Independent outcomes}
]
\addplot[color=plotBlue,mark=*,solid,line width=.85pt,mark size=2.2pt] coordinates {(0.1,1.3812024347081509) (1.0,8.003624613636529) (10.0,17.200135720945887) (100.0,19.964443247526486)};
\addlegendentry{$c_F=1$}
\addplot[color=plotOrange,mark=square*,dashed,line width=.85pt,mark size=2.2pt] coordinates {(0.1,0.1489282841756201) (1.0,1.3812024347081842) (10.0,8.00362461363654) (100.0,17.200135720945887)};
\addlegendentry{$c_F=10$}
\addplot[color=plotGreen,mark=triangle*,dotted,line width=.85pt,mark size=2.2pt] coordinates {(0.1,0.015010285330918638) (1.0,0.1489282841756312) (10.0,1.3812024347082175) (100.0,8.00362461363655)};
\addlegendentry{$c_F=100$}
\end{axis}
\begin{axis}[
font=\footnotesize,
axis line style={black},tick style={black},
tick label style={text=black},label style={text=black},
title style={text=black,font=\footnotesize},
legend style={font=\footnotesize,text=black,draw=none,fill=none},
grid=major,grid style={black!12},
scaled ticks=false,
clip marker paths=true,
width=.47\linewidth,height=5.1cm,
xmode=log,log basis x=10,
xmin=.08,xmax=125,
xtick={.1,1,10,100},xticklabels={.1,1,10,100},
xlabel={$c_Y$},ylabel={Saving over Singleton (\%)},
legend pos=north west,legend cell align=left,
at={(priceIndependent.outer east)},anchor=outer west,xshift=10pt,
ymin=15,ymax=65,ytick={20,30,40,50,60},
title={Correlated outcomes}
]
\addplot[color=plotBlue,mark=*,solid,line width=.85pt,mark size=2.2pt] coordinates {(0.1,19.184976381770046) (1.0,25.05713105858308) (10.0,46.26235182108823) (100.0,57.438879488550974)};
\addlegendentry{$c_F=1$}
\addplot[color=plotOrange,mark=square*,dashed,line width=.85pt,mark size=2.2pt] coordinates {(0.1,18.357947513088334) (1.0,19.184976381770046) (10.0,25.05713105858307) (100.0,46.26235182108823)};
\addlegendentry{$c_F=10$}
\addplot[color=plotGreen,mark=triangle*,dotted,line width=.85pt,mark size=2.2pt] coordinates {(0.1,18.26806974438504) (1.0,18.35794751308838) (10.0,19.184976381770046) (100.0,25.05713105858308)};
\addlegendentry{$c_F=100$}
\end{axis}
\end{tikzpicture}
}

\caption{Relative savings across both cost components.}
\label{fig:new-prices}
\end{figure}

\subsection{Coverage and regret at short horizons}
\label{ec:new-horizons}
\label{sec:theoretical-guarantees}
\label{subsec:short-horizon-experiments}
We next conduct two additional experiments to examine whether valid confidence statements become informative within a limited number of evaluations. The first uses the finite-model example from Section~\ref{subsec:joint-tracker-experiment} of the main paper. The second examines the confidence sequences used by PP-LUCB in Algorithm~\ref{alg:pp-lucb}. These experiments examine how the coverage guarantees relate to interval width, recommendation quality, and the time required to certify a selection.

For the first experiment, we use the three-configuration example under $\nu^0$, independent outcomes and $\delta=.05$. At $T=100,250,500,1000,2000$, stopped paths retain their recommendation, confidence set and cost. Coverage requires true-model inclusion through $\min(T,\tau)$. Each arm interval spans its means among remaining models. We report the average maximum arm width. Simple regret is $\theta^\star-\theta_{\widehat k_T}$, using the MLE before stopping and the final recommendation afterward. The stopping fraction estimates $\Pr(\tau\le T)$ over all paths. Figure~\ref{fig:new-horizons} compares Joint, Singleton, Fixed audit and Full bundle on each of these four measures. The black dashed line marks $.95$ coverage, and regret bands are $1.96$ Monte Carlo standard errors. Joint increasingly certifies its recommendation as its intervals become narrower.

For the second experiment, we use a four-configuration example with Bernoulli means $(.70,.65,.55,.50)$, equally likely independent states $S\in\{0,1\}$ and proxy $f_k+.1(2S-1)$, where $f=(.58,.80,.60,.68)$. Conditional on the selected arm, outcomes are independent of state. The floor is $.05$. Every arm is initialized once using that purchase probability. Subsequent purchases use the fixed state-dependent probabilities $.1$ and $.4$ in states 0 and 1. These probabilities satisfy the audit floor, while the rule derived in Section~\ref{subsec:neyman-derivation} is another optional choice. Algorithm~\ref{alg:pp-lucb} selects the empirical leader and largest-upper-endpoint challenger, samples them in order, and appends all configurations at square-numbered iterations. Its proxy-plus-IPW-residual estimate uses the exact stitched widths with component error budget $.05/(2K)$.

Table~\ref{tab:new-horizons} reports the results for PP-LUCB. Its width is averaged over arms and paths before intersection with $[0,1]$, and coverage means simultaneous inclusion of all configuration means. This differs from the maximum arm width and model-inclusion coverage used in the first experiment. All 1,000 PP-LUCB paths remain covered and uncertified at these horizons. Regret decreases, but the confidence sequences remain too wide to certify a recommendation. The two experiments therefore assess the practical information in their respective confidence sets. They use different examples and do not provide a direct cost comparison.

\begin{table}[t]
\color{MSRevisionBlue}
\centering\scriptsize
\caption{PP-LUCB coverage, regret, interval widths and stopping fractions.}
\label{tab:new-horizons}
\begin{tabular}{rrrrr}\toprule
$T$ & Coverage & Simple regret & Mean raw width & Stopped\\\midrule
100 & 100\% & 0.06855 & 48.30 & 0\%\\
250 & 100\% & 0.05170 & 32.16 & 0\%\\
500 & 100\% & 0.03625 & 22.29 & 0\%\\
1000 & 100\% & 0.02600 & 15.89 & 0\%\\
2000 & 100\% & 0.01285 & 11.11 & 0\%\\
\bottomrule\end{tabular}
\end{table}
\begin{figure}[t]
\color{MSRevisionBlue}
\centering\includegraphics[width=.96\linewidth]{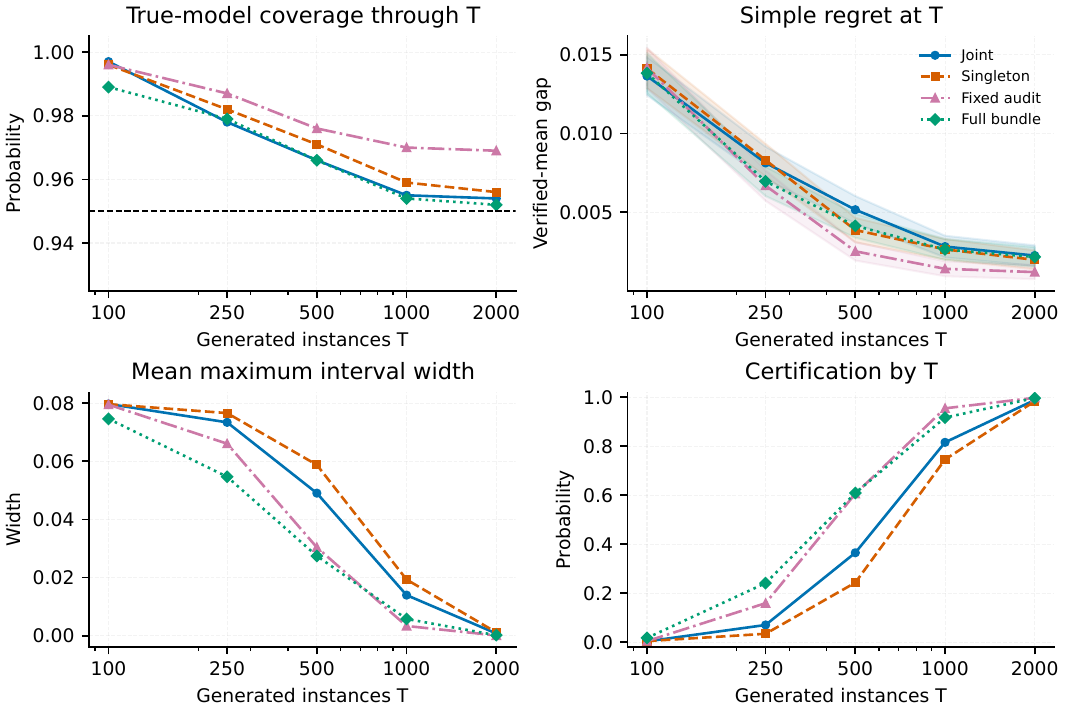}
\caption{Coverage, simple regret, interval widths and stopping fractions at short horizons.}
\label{fig:new-horizons}
\end{figure}

\subsection{Larger configuration sets}
\label{ec:new-scale}
The final experiment extends the three-configuration comparison in Section~\ref{subsec:joint-tracker-experiment} to larger configuration sets. It examines whether the cost benefit of reference sharing persists and how much computation is required when likelihood and allocation updates involve more configurations. We use $K=M\in\{32,128\}$. Under model $m$, configuration $m$ has independent Bernoulli mean $.6$ and every other configuration has mean $.5$. State and proxy are constant. The menu contains singletons, a partition into groups of eight and the full set, giving $J=K+K/8+1$. Independence makes vector KL additive. The floor is $.1$, tracking exponent $.6$, risk $.05$, threshold $\log((K-1)/.05)$, configuration-evaluation cost $.05$ and reference-acquisition cost 10.

The true zero-based index is $K/2$, and initial MLE ties select zero. Every model allocation is obtained from one solved LP by $k\mapsto k\mathbin{\mathrm{xor}}m$, which maps model zero to $m$ and preserves the declared menu. Positive-mass actions audit fully, while zero-target exploration actions use the floor.

Each size and procedure uses 300 paths, capped at 20,000 rounds. Table~\ref{tab:new-scale} includes every path's cost through stopping or the cap. The Errors column counts paths that stop with an incorrect recommendation, with all 300 paths as the denominator. A path that reaches the cap without stopping is not counted as an incorrect recommendation. The Stopped column reports the fraction that stop, so both columns are needed to assess selection performance. At $K=128$, Singleton has 15 censored paths. All paths in the remaining cells stopped before the cap. CPU seconds measure the complete vectorized 300-path batch, including likelihood and tracking updates. LP time is measured separately. Joint and Full bundle stop on every path and have similar costs, while Singleton has substantially higher capped cost at both sizes. These results support the benefit of reference reuse in this symmetric example. The reported computation times concern the stated menu and symmetry-based LP reuse, which are part of the experiment design.

\begin{table}[t]
\color{MSRevisionBlue}
\centering\scriptsize
\caption{Selection costs and computation times for larger configuration sets.}
\label{tab:new-scale}
\begin{tabular}{rrlrrrrr}\toprule
$K$ & $J$ & Procedure & Capped cost (SE) & Stopped & Errors & LP (ms) & CPU (s)\\\midrule
32 & 37 & Joint & 3980 (89) & 100\% & 2/300 & 10.8 & 0.24\\
32 & 37 & Singleton & 16091 (790) & 100\% & 20/300 & 3.0 & 1.92\\
32 & 37 & Full bundle & 3910 (92) & 100\% & 1/300 & 2.8 & 0.25\\
128 & 145 & Joint & 7873 (150) & 100\% & 0/300 & 3.7 & 0.84\\
128 & 145 & Singleton & 60968 (3051) & 95\% & 9/300 & 3.9 & 11.05\\
128 & 145 & Full bundle & 7791 (150) & 100\% & 0/300 & 4.0 & 0.80\\
\bottomrule\end{tabular}
\end{table}

\end{revblock}

\begin{revblock}
\subsection{Sensitivity to model misspecification}
\label{sec:ec-misspecification}

The preceding finite-model experiments assume that the candidate family contains the data-generating law. This subsection examines the effect of violating that condition and then considers a separate route to certification using bounded outcomes. We first report a misspecified example for Joint. The remaining comparisons retain shared references and examine whether removing resolved configurations reduces the cost of selection without using the candidate likelihoods.

For Joint, a shift of .025 in the conditional success probabilities produces 433 incorrect recommendations in 1,000 paths under the correlated \(\nu^1\) example. The shift preserves configuration means and gaps, so the best configuration is unchanged. This example concerns the specification of the conditional observation law, rather than a change in the selection target. It shows that the likelihood-based stopping rule can give unreliable recommendations when the data-generating law lies outside the candidate family.

To examine selection without the candidate likelihoods, we use 90 laws based on the six model--correlation settings above. The grid includes the matched laws; opposite statewise shifts of
\(\pm\eta(1,-1,1)\), with \(\eta\in\{.025,.05,.1\}\); correlation changes from \(c\) to \(c+\eta(1-c)\), with \(\eta\in\{.1,.2,.4\}\); probabilities compressed toward .6 by fractions .25, .5 and .75; and settings with state-0 probabilities .4 or .6. Here \(c\) is the common-uniform mixing probability. The statewise shifts and correlation changes preserve configuration means, while compression toward .6 reduces the gaps. Each retained law has a unique best configuration. We use 1,000 fresh paths per law, \(c_F=1\), \(c_Y=10\), \(\delta=.05\), and round caps of 20,000, 100,000, 500,000 and 1,500,000.

All procedures in this comparison purchase a reference on every evaluated round. The allocation decision therefore concerns which configurations to include in the bundle, rather than whether to purchase its reference after observing the state. For the active mixture, we fix the active bundle before each 1,000-round batch. For each directed pair, define
\[
E_{ij,t}
=
\frac{1}{12}
\sum_{r=1}^{12}
\prod_{\substack{s\le t:\\ \{i,j\}\subseteq B_s}}
\left\{1+2^{-r}(Y_{i,s}-Y_{j,s})\right\}.
\]
At a batch checkpoint, we remove an active configuration \(j\) when an active configuration \(i\) has
\(E_{ij,t}\ge (K-1)/\delta\). We retain the old active set if the removals would empty it, and return the last configuration when only one remains.

The error guarantee follows from the same fresh-draw and bundle-mean conditions used for bounded-outcome inference. Under Assumption~3 and \(\theta_i\le\theta_j\), each factor has conditional mean at most one, and bounded outcomes make the factors nonnegative. The products, and hence their mixtures, are nonnegative supermartingales. A wrong return requires removing a fixed true best configuration \(k^\star\). Applying Ville's inequality to its \(K-1\) incoming processes bounds this event by \(\delta\). For \(K=3\), this gives the threshold \(2/\delta\), rather than the \(6/\delta\) threshold used for simultaneous control of all six directed comparisons. This argument does not require the true law to belong to the candidate family.

We compare this active mixture with four alternatives: the same inference under full-bundle sampling, the earlier mixture using the \(6/\delta\) threshold, predictable plug-in stakes using the smaller threshold, and marginal reference confidence sequences. All procedures use the same checkpoints and paired fresh batches. Active and full-bundle counterparts share inference and stopping, so their cost difference measures the saving from no longer evaluating configurations that have been removed. Comparing the two mixture thresholds instead measures the effect of allocating the error budget specifically to best configuration identification.

Table~\ref{tab:robust-misspecification} reports the shortest and longest round caps. Costs include every stopped or capped path, and incorrect recommendations count only paths that stop with a wrong answer. At the largest cap, the active mixture with threshold \(2/\delta\) stops on all 90,000 paths with one incorrect recommendation. Its matched full-bundle counterpart and both plug-in procedures also have one incorrect recommendation each. The earlier \(6/\delta\) mixtures have none. The marginal reference confidence sequences likewise have no incorrect recommendations, but stop on 89,373 of the 90,000 paths.

At this cap, the active mixture's mean cost is 221,201, which is 17.37\% below that of the active mixture using \(6/\delta\), with paired 95\% interval \([16.90\%,17.79\%]\). Its cost is also 6.43\% below that of the matched full-bundle procedure using \(2/\delta\). The first comparison shows the benefit of the smaller valid boundary. The second shows the additional saving from dropping resolved configurations while keeping the inference and stopping rule fixed.

The stopping results also show why cost and error counts must be read together. At the 20,000-round cap, the least often completed condition stops on only 9 of 1,000 paths. Larger budgets are needed for the small-gap laws, even though the comparison processes remain valid. The reported mean costs average over laws with different gaps and therefore do not provide a matched cost comparison with Joint in Section~7.1.

The bounded-outcome procedure performs well under the model perturbations considered here. At the largest round cap, the active mixture certifies a selection on every path with only one incorrect recommendation across the 90,000 paths, while reducing cost relative to matched full-bundle sampling. These results show that shared-reference evaluation and the removal of resolved configurations can support reliable, lower-cost selection even when the candidate likelihoods do not describe the evaluation data.

\begin{table}[tbp]
\color{MSRevisionBlue}
\centering\scriptsize
\caption{Costs, stopping and incorrect recommendations under model misspecification.}
\label{tab:robust-misspecification}
\begin{tabular}{lrrr}\toprule
Procedure & Mean cost & Stopped & Incorrect\\\midrule
\multicolumn{4}{l}{(a) Round cap: 20,000}\\
Active mixture, $2/\delta$ & 97,069 & 77,902 & 1\\
Full bundle, $2/\delta$ & 102,256 & 77,902 & 1\\
Active mixture, $6/\delta$ & 108,739 & 75,413 & 0\\
Full bundle, $6/\delta$ & 114,520 & 75,413 & 0\\
Plug-in, active & 103,770 & 77,588 & 1\\
Plug-in, full bundle & 108,729 & 77,588 & 1\\
Marginal reference CS & 219,805 & 39,745 & 0\\
\midrule
\multicolumn{4}{l}{(b) Round cap: 1,500,000}\\
Active mixture, $2/\delta$ & 221,201 & 90,000 & 1\\
Full bundle, $2/\delta$ & 236,397 & 90,000 & 1\\
Active mixture, $6/\delta$ & 267,710 & 90,000 & 0\\
Full bundle, $6/\delta$ & 286,244 & 90,000 & 0\\
Plug-in, active & 242,480 & 90,000 & 1\\
Plug-in, full bundle & 258,624 & 90,000 & 1\\
Marginal reference CS & 1,152,425 & 89,373 & 0\\
\bottomrule\end{tabular}
\end{table}

\end{revblock}

\begin{revblock}
\section{Detailed Service-System Settings}\label{sec:service-case-studies}
This section provides more details of the experiment on the BANKING77 dataset in Section~\ref{subsec:banking-bundle}. We then consider an additional experiment on the MT-Bench dataset, where one human judgment concerns a pair of responses rather than all configurations on a case. These experiments fix the evaluation population and compare the order of acquiring references within that population. Last, we give an example of a service score to support the bounded-outcome formulation in Section~\ref{sec:model} of the main paper.

\subsection{Experiment on the BANKING77 dataset}
\label{subsec:additional-acquisition}
\label{subsec:tickets-retained}
This subsection provides the cohort construction, confidence rule and acquisition settings for the experiment on the BANKING77 dataset \citep{casanueva2020intent} in Section~\ref{subsec:banking-bundle}. That experiment examines how much reference acquisition can be reduced by using predictions already available for every query. Its primary comparison uses a fixed 385-query cohort with five queries per intent, error probability $\delta=.05$ and tolerance $\epsilon=.01>1/385$. Four model-prompt configurations (Fable direct, Fable disambiguation, Astra direct and Astra disambiguation) receive each query and the 77-intent taxonomy without reference labels or other outputs. The Fable configurations use Fable~5.1. All configurations answer before the label is requested, so one label evaluates all four outputs.

The cohort selects five unused texts per intent by a hash-based ordering, after removing duplicates and conflicting labels. It has no overlap with the two 77-query cohorts described below. There are 1,540 completed outputs, and 338 of the 385 queries have unanimous predictions. These queries have zero pairwise accuracy differences, so only the remaining 47 queries need labels for comparison. The configuration-evaluation cost is $1,540\times.01=15.4$ units. The reference-acquisition cost is one unit per query label. Section~\ref{subsec:banking-bundle} reports the configuration accuracies and acquisition costs. The additional cost saving from Adaptive sampling relative to Disagreement sampling has paired 95\% interval $[7.87\%,9.34\%]$. This interval concerns variation across paired reveal orders, conditional on the cohort and generated outputs.

Two disjoint 77-query cohorts provide additional comparisons to examine how the tolerance and observed disagreement affect the need for references. One balanced cohort certifies equivalence without labels at $\epsilon=.05$. The other uses prespecified $\epsilon=.01$, where disagreement sampling saves $82.03\%$ over uniform acquisition and adaptive sampling saves another $15.04\%$. Table~\ref{tab:real-selection-replacements} reports the latter comparison. The model queries use batches of four in the 77-query cohorts and 16 in the 385-query cohort, so their batch contexts differ. Reveal-order repetitions measure variation in acquisition cost conditional on the generated outputs. The target is intent accuracy in each fixed cohort, and the experiment does not assess the models' prior exposure to this public dataset.

\paragraph{Finite-population confidence rule.}
We use the following anytime-valid finite-population confidence rule to obtain valid stratum-level bounds under sequential, adaptively scheduled label acquisition.

For a stratum with $N$ binary labels and unknown total $m$, let $s$ of the
$n$ uniformly sampled labels be successes. Write
\[
 H_m(s,n)=\frac{\binom m s\binom{N-m}{n-s}}{\binom N n}.
\]
The mixture over totals $m=0,\ldots,N$ has count likelihood $1/(n+1)$. Its ratio $1/((n+1)H_m(s,n))$ is a nonnegative supermartingale under total $m$, so intersecting
\[
 \{m\in\{0,\ldots,N\}:H_m(s,n)\geq\alpha/(n+1)\}
\]
over the observed counts gives an anytime confidence set. To verify the
mixture identity, sum
$\binom m s\binom{N-m}{n-s}$ over $m$ and use Vandermonde's identity to
obtain $\binom{N+1}{n+1}$. Division by $(N+1)\binom N n$ gives
$1/(n+1)$. The likelihood-ratio argument uses the ordered sample path. The common factor $\binom n s$ cancels in its numerator and denominator.
At boundary totals, the ratio is a supermartingale because alternative paths
may lie outside the null support. Ville's inequality gives the stated coverage.

The simultaneous-coverage argument below assumes a fixed partition common to all configuration pairs. Such a partition can be obtained by taking the common refinement of the strata defined by pre-audit predictions and proxy categories, before any label is acquired. Acquired references may guide the choice of a common stratum, followed by uniform sampling among its unrevealed items. Under this sampling condition, each stratum contributes a uniformly sampled sequence even when its visit times depend on earlier labels. Pausing a stratum preserves its confidence sequence. The error budgets of all predeclared component sequences must sum to at most $\delta=.05$.

The following calculation explains how the binary-total confidence sets bound one accuracy contrast. For a fixed pair of configurations $j,k$, let $h$ index the common strata, with $N_h$ queries in a cohort of size $N$. Write $[\ell_{ah}(t),u_{ah}(t)]$ for the running confidence interval for the mean of configuration $a$'s correctness indicator in stratum $h$, for $a\in\{j,k\}$. The contrast is the difference of these two binary indicators. An upper bound on configuration $j$'s accuracy advantage over configuration $k$ is therefore
\[
 U_{jk}(t)=\sum_h\frac{N_h}{N}
       \bigl(u_{jh}(t)-\ell_{kh}(t)\bigr).
\]
If the two configurations make the same prediction throughout a stratum, its contribution is exactly zero without acquiring labels. A reference acquired on any other query updates the correctness indicators of all four configurations within the same common stratum, while its cost is charged only once. Under the common-partition and sampling conditions above, simultaneous coverage of the component sequences makes all these contrast bounds valid at every acquisition time.

Uniform acquisition samples from the full cohort. Disagreement sampling restricts it to differing predictions. Adaptive sampling uses acquired labels to choose a stratum according to the estimated reduction in the confidence-interval width of an unresolved accuracy comparison, then samples uniformly within that stratum. A configuration $k$ is certified when $U_{jk}(t)\le\epsilon$ for every competitor $j$. On simultaneous coverage, this condition implies that its fixed-cohort accuracy is within $\epsilon$ of the best. The main paper reports the primary tolerance $\epsilon=.01$ and the additional tolerances $.02$ and $.05$. Each method and tolerance uses 400 paired reveal orders with the queries, generated outputs and reference labels held fixed. Costs include the common output-generation charge and the labels acquired before certification. Figure~\ref{fig:banking-acquisition} of the main paper reports these results.

For the same acquisition paths, the total cost at reference-acquisition price $c_Y$ is $15.4+c_Y n_Y$, where $n_Y$ is the number of acquired labels. This expression changes the accounting price while holding the acquisition rule fixed. It explains how the common output-generation charge affects the relative saving as reference acquisition becomes more expensive. Deployment frequencies differing from the balanced 77-intent cohort require corresponding target-accuracy weights.

\subsection{Experiment on the MT-Bench dataset}
\label{subsec:private-preference-details}
This additional experiment examines whether judge scores help choose informative human judgments when a reference concerns a pair of responses. It complements the experiment on the BANKING77 dataset by considering references that apply only to their recorded comparisons. The purpose is to assess selective acquisition when a single label cannot evaluate all configurations. It uses published human preference records for six language models,
\texttt{gpt-4}, \texttt{claude-v1}, \texttt{gpt-3.5-turbo},
\texttt{vicuna-13b-v1.2}, \texttt{alpaca-13b} and \texttt{llama-13b}
\citep{zheng2023judging}. Human and judge records are matched by question,
turn, response text and orientation. Reversing the order reverses the
preference. All 3,355 human records are retained as separate judgments. A question-level hash split assigns 1,316 records to development and 2,039 to assessment across all 15 pairs. The development records are used to select the acquisition rule before assessment. The selected rule is then evaluated on the fixed assessment pool. The target is each model's mean preference over uniformly weighted opponents in this pool.

Encode loss, tie and win as $Y,F\in\{0,.5,1\}$. Each judgment evaluates one pair, while a BANKING77 label evaluates all predictions on its query. The finite-population
confidence rule above applies to preference scores through
$Y=(\mathbf1\{Y=1\}+\mathbf1\{Y>0\})/2$. Two binary confidence sequences therefore bound each ternary mean. Risk $.05$ is allocated to all predeclared component sequences, and the stopping tolerance is $.05$. Each acquired judgment updates both models in its pair, since reversing the comparison changes a score $Y$ to $1-Y$. Pair means are aggregated with equal weight over the five opponents of each model. The stopping rule uses the running intersection of the component confidence sets, so information from an earlier acquisition remains available when the current interval is wider.

{\color{black}
The MT-Bench documentation and related studies report several common biases in LLM-based
judges\rev{.}
\begin{itemize}[leftmargin=*]
\item Position bias\rev{.} \rev{The} judge may prefer the response shown in position A (first) more often than
warranted.
\item Verbosity bias\rev{.} \rev{The} judge may favor longer and more detailed responses even when they are not
more correct.
\item Self-preference and model-specific bias\rev{.} \rev{A} judge may systematically rate some model styles
more favorably, depending on the prompting and the judge model itself.
\end{itemize}
In our setting, these biases translate into an arm- and instance-dependent gap between the proxy
$F$ and the verified outcome $Y$. This is precisely the residual term in the decomposition
$\theta_k=\E[F\mid k]+\E[Y-F\mid k]$ from \eqref{eq:pp-identity}.
}
The finite-population acquisition rule above groups records by proxy
category and uses human labels to bound the score difference for each pair.

Purchased judgments and pre-audit metadata guide comparison selection. Active labels only omits proxies. Proxy Neyman allocates judgments using estimated residual variation, following the estimator criterion in Section~\ref{subsec:neyman-derivation}. Proxy lookahead chooses a pair and proxy stratum according to the expected reduction in an unresolved contrast's width after one additional judgment. It combines this rule with exploration at square-numbered acquisition times. After choosing a stratum, it samples uniformly from the unrevealed judgments in that stratum. Labels-only lookahead uses the same one-judgment width-reduction criterion without proxy scores. This comparison assesses the contribution of proxy information while retaining a common acquisition objective.

Obtaining a proxy score costs $.01$ per distinct comparison in the assessment pool, and acquiring a human judgment costs one. Separate human ratings remain separate purchasable judgments even when they refer to the same response comparison. A shared proxy score is charged once for that comparison, and no human judgment is purchased twice. Thus labels-only costs equal the acquired judgment count, while proxy costs also include scoring the distinct assessment comparisons. The reported costs concern acquisition on the fixed assessment pool. They use normalized prices for human judgments collected independently of this experiment.

Table~\ref{tab:real-selection-replacements} reports costs for the experiment on the MT-Bench dataset at $\epsilon=.05$ and the 77-query BANKING77 cohort at $\epsilon=.01$, each using 400 reveal orders. Costs include the applicable output-generation or proxy-scoring charges. Results for the 385-query cohort are in Figure~\ref{fig:banking-acquisition} of the main paper. The cost of Proxy lookahead is 1,098.46, compared with 1,133.11 for Labels-only
lookahead, a saving of $3.06\%$ with paired 97.5\% interval
$[1.04\%,5.00\%]$. Proxy Neyman costs 1,538.66. Within this population, the proxy rule based on variance has higher cost
than either lookahead rule. The positive saving from Proxy lookahead shows that judge scores can help target judgments in this population. The higher cost of Proxy Neyman also shows that a useful proxy does not make every proxy-based allocation more efficient. Reference choice must consider which comparisons remain unresolved.

\begin{table}[t]
\color{MSRevisionBlue}
\centering\small
\caption{Total costs on MT-Bench and the 77-query BANKING77 cohort.}
\label{tab:real-selection-replacements}
\begin{tabular}{llrr}
\toprule
Task & Acquisition & Mean labels & Mean cost \\
\midrule
MT-Bench & Active labels only & 1217.11 & 1217.11 \\
MT-Bench & Labels-only lookahead & 1133.11 & 1133.11 \\
MT-Bench & Proxy Neyman & 1527.47 & 1538.66 \\
MT-Bench & Proxy lookahead & 1087.28 & 1098.46 \\
\midrule
Banking (77) & Uniform & 49.84 & 52.91 \\
Banking (77) & Disagreement sampling & 6.43 & 9.51 \\
Banking (77) & Adaptive sampling & 5.00 & 8.08 \\
\bottomrule
\end{tabular}
\end{table}

\subsection{Service-score construction}
\label{subsec:service-fixture-details}
This example supports the definition of a bounded verified outcome in Section~\ref{sec:model}. It explains how a service objective can combine classification quality, timeliness and escalation, extending the single accuracy measure used in Section~\ref{subsec:banking-bundle}. Each configuration combines a routing rule, a prompt and a language model. Consider a ticket with subject and body text and a specified priority annotation. The annotation supplies the reference priority used to calculate the score. We describe the scoring rules conditional on this reference being available. This makes explicit which components require reference information and how they enter the service objective.
{\color{black}
For each ticket, the model outputs a predicted priority
$\widehat p\in\{\mathrm{critical},\mathrm{high},\mathrm{medium},
\mathrm{low},\mathrm{very\ low}\}$. We then simulate an operational response time
$T$ under the arm's routing policy and compute three components\rev{,}
\[
 Y=0.5\times\mathrm{accuracy}+0.3\times\mathrm{SLA\ compliance}
       +0.2\times\mathrm{efficiency},
\]
where each component is in $[0,1]$\rev{.}
\begin{itemize}[leftmargin=*]
\item Accuracy (50\% weight)\rev{.} Classification correctness measured by priority distance to the true
label $p^\star$. It is set to 1.0 for exact match, 0.8 for one-level error, 0.5 for two-level error, and 0.2
for three or more levels.
\item SLA compliance (30\% weight)\rev{.} Binary indicator $\mathbf1\{T\le\mathrm{SLA}(p^\star)\}$, using priority-dependent
SLA deadlines\rev{, namely} critical (5 min), high (15 min), medium (60 min), low (4 hr), very low (24 hr).
\item Efficiency (20\% weight)\rev{.} $1-\rev{\mathrm{escalation\ indicator}}$, where escalation is determined by a fixed observe-then-escalate rule\rev{.} \rev{A} ticket is escalated to a human specialist if the predicted priority is
critical.\footnote{This fixed escalation rule is not a separate design layer. It is held constant across arms, and provides an explicit operational channel
through which misclassification can reduce efficiency (e.g., over-predicting critical can increase escalations).}
\end{itemize}
}
These weights and deadlines define the service objective. A priority prediction affects both classification accuracy and escalation, while the response-time rule determines SLA compliance. The score therefore makes the operational consequences of a prediction explicit.
\end{revblock}
\rev{For this scoring example, a configuration specifies a routing rule, a prompt strategy and a language model. The following choices illustrate how each component enters the score.}

\paragraph{\rev{Routing policy.}}
The routing policy determines the simulated response time for a ticket
and therefore its SLA compliance. It does not change the ticket text or the model's input. We
consider three policies\rev{.}
\begin{itemize}[leftmargin=*]
\item FIFO\rev{.} Tickets are processed in arrival order without priority differentiation. This can cause
urgent tickets to wait behind non-urgent ones.
\item Priority\rev{.} Tickets are dispatched by urgency (critical $\to$ high $\to$ medium $\to$ low $\to$ very low),
so higher-priority tickets are served earlier.
\item Adaptive\rev{.} SLA-based dynamic routing that targets response times proportional to each ticket's
SLA deadline, to reduce deadline violations under heterogeneous SLAs.
\end{itemize}
\rev{These policies specify illustrative response-time rules. They can use the reference priority, which is already available in this example. The experiment on the review queue in Section~\ref{EC-sec:queue-service} instead models the time required to obtain pending references.}

\paragraph{\rev{Prompt strategy.}}
The prompt strategy controls how the model is instructed to predict
the priority\rev{.}
\begin{itemize}[leftmargin=*]
\item Standard\rev{.} Direct classification prompt\rev{.} ``Classify this support ticket into priority: critical, high,
medium, low, very low.''
\item Chain-of-Thought (CoT)\rev{.} Step-by-step reasoning prompt\rev{.} ``Analyze this support ticket: (1) What
is the main issue? (2) How urgent? (3) Business impact? (4) Classify priority.''
\end{itemize}

\paragraph{\rev{Language model.}}
\rev{The language model produces the priority prediction. This prediction enters the service score through classification accuracy and the specified escalation rule. The same score definition can therefore be applied to different models.}

\rev{The response-time rules specify the SLA component of the score.}
\begin{itemize}[leftmargin=*]
\item FIFO\rev{.} Base time $=600\times1.5=900$ seconds with noise $\sim N(0,120^2)$. Since FIFO does not
prioritize urgent tickets, critical tickets frequently miss the 5-minute SLA.
\item Priority\rev{.} Base time is scaled inversely with an urgency weight. \rev{The illustrative calibration assigns critical tickets a response time around 150 seconds and low-priority tickets a response time around 900 seconds.}
\item Adaptive\rev{.} Response time is set as $0.25\times\mathrm{SLA}(p^\star)$, \rev{which makes the SLA indicator equal to one under this deterministic rule.}
\end{itemize}
\rev{These response-time rules explain how routing enters the score. Their role is to define a service objective, rather than to estimate performance in a queue with limited review capacity.}

\paragraph{\rev{Reference-informed judge score.}}
For each instance, we also construct a \rev{judge} score $F\in[0,1]$ using the
judge model (gpt-5.2-chat). The judge is provided with the ticket text, the predicted priority
$\widehat p$, the true outcome $p^\star$, and the operational metadata needed to assess timeliness and efficiency
(simulated response time $T$ and the escalation indicator). It outputs a structured score\rev{.}
\begin{enumerate}[leftmargin=*]
\item Classification Quality (0-50 points)\rev{.} Exact match (50), one-level off (35), two-level off (20),
three+ levels off (5).
\item Timeliness (0-30 points)\rev{.} \rev{The score is 30 for $T<\mathrm{SLA}$, 20 for $\mathrm{SLA}\le T<1.5\times\mathrm{SLA}$, 10 for $1.5\times\mathrm{SLA}\le T<2\times\mathrm{SLA}$, and 0 for $T\ge2\times\mathrm{SLA}$.}
\item Operational Efficiency (0-20 points)\rev{.} Resolved without escalation (20), escalated but appropriate
(10), unnecessary escalation (0).
\end{enumerate}
The total is a 0-100 score, normalized to $F\in[0,1]$ by dividing by 100.

\begin{revblock}

Here the judge receives the same reference priority used to define $Y$. Its score can differ because its component scales differ from the service objective. For example, a one-level priority error receives $.8$ in the accuracy component of $Y$ but 35 of 50 classification points from the judge. Thus the example clarifies how two scoring rules can disagree even with the same reference information. This reference-informed score is not a pre-audit proxy. Selection with unavailable references is evaluated in the preceding experiments and in Section~\ref{EC-sec:queue-service}.

\end{revblock}

\begin{revblock}
\section{Audit Rules and Examples of Strict Cost Differences}
\label{sec:statewise-reference-policies}
This section explains why an audit rule for accurate estimation can be different from an audit rule for identifying the best configuration. Section~\ref{subsec:neyman-derivation} below derives the optional Neyman rule used with the prediction-powered estimator in Section~\ref{sec:estimation} and PP-LUCB in Section~6.2 of the main paper. The remaining subsections use the information program in Section~\ref{sec:theory_lower_bound} to examine the cost of restricting audit decisions. We first give conditions under which these restrictions leave the characteristic cost unchanged, and then give a finite example in which a rule based on residual second moments has strictly higher cost. They explain the different roles of variance and information in the two procedures.
\end{revblock}

\subsection{\rev{The estimator-oriented auditing rule}}
\label{subsec:neyman-derivation}
\rev{The clipped Neyman rule follows Section~7 of \citet{zrnic2024active}, with audit floor $\pi_{\min}=p$. It allocates a fixed average audit rate to reduce the variance of the residual correction for one configuration. This supplies the estimator-oriented rule discussed in Section~6.2. We first treat $\pi_{\min}<\rho<1$ and positive conditional residual second moments, and then address zero moments and endpoints.}

We derive the auditing policy that minimizes the asymptotic variance of the IPW residual estimator
under an average audit-rate constraint.

Fix arm $k$. On each pull of arm $k$, we observe an instance $X_t$ and a proxy score $F_t\in[0,1]$, and
define the residual
\[
R_t\coloneqq Y_t-F_t\in[-1,1].
\]
An audit is requested with probability $\pi_t$, where
\[
\pi_t=\pi_k(X_t,F_t)\in[\pi_{\min},1],\qquad
A_t\mid(\rev{\mathcal H_{t-1}},k_t=k,X_t,F_t)\sim\operatorname{Bernoulli}(\pi_t).
\]
The IPW residual increment used in the estimator is $Z_t=(A_t/\pi_t)R_t$. Define the conditional first and
second moments of the residual given the observables at audit-decision time\rev{,}
\begin{align*}
m_k(x,f)&\coloneqq\E[R_t\mid k_t=k,X_t=x,F_t=f],\\
g_k(x,f)&\coloneqq\E[R_t^2\mid k_t=k,X_t=x,F_t=f].
\end{align*}

Under Assumption~\ref{ass:audit}, conditional on the history and current observables
$(\rev{\mathcal H_{t-1}},k_t,X_t,F_t)$, the
audit decision $A_t$ is independent of the potential outcome $Y_t$, and thus independent of $R_t=Y_t-F_t$.
Therefore, conditioning on $(k_t=k,X_t,F_t)$,
\begin{align*}
\E[Z_t\mid k_t=k,X_t,F_t]
&=\E\!\left[\frac{A_t}{\pi_t}R_t\middle|k_t=k,X_t,F_t\right]\\
&=\E[R_t\mid k_t=k,X_t,F_t]=m_k(X_t,F_t),
\end{align*}
and, using $A_t^2=A_t$ for Bernoulli variables,
\begin{align*}
\E[Z_t^2\mid k_t=k,X_t,F_t]
&=\E\!\left[\frac{A_t}{\pi_t^2}R_t^2\middle|k_t=k,X_t,F_t\right]\\
&=\frac{1}{\pi_t}\E[R_t^2\mid k_t=k,X_t,F_t]
=\frac{g_k(X_t,F_t)}{\pi_t}.
\end{align*}
Then the conditional variance is
\[
\Var(Z_t\mid k_t=k,X_t,F_t)=\frac{g_k(X_t,F_t)}{\pi_t}-m_k(X_t,F_t)^2.
\]
Taking expectations, and noting that $\E[Z_t\mid k_t=k]=\E[m_k(X_t,F_t)\mid k_t=k]=\mu_{R,k}$, we obtain
\[
\Var(Z_t\mid k_t=k)=\E\!\left[\frac{g_k(X_t,F_t)}{\pi_k(X_t,F_t)}\middle|k_t=k\right]-\mu_{R,k}^2.
\]
Since $-\mu_{R,k}^2$ does not depend on the audit policy, minimizing $\Var(Z_t\mid k_t=k)$ is equivalent to
minimizing
\[
J(\pi_k)\coloneqq\E\!\left[\frac{g_k(X_t,F_t)}{\pi_k(X_t,F_t)}\middle|k_t=k\right]
\]
over measurable policies $\pi_k:\cX\times[0,1]\to[\pi_{\min},1]$ that satisfy the average audit-rate constraint
$\E[\pi_k(X_t,F_t)\mid k_t=k]=\rho$.

Consider the Lagrangian functional with multiplier $\lambda\in\mathbb R$\rev{,}
\[
\mathcal L(\pi_k,\lambda)=\E\!\left[\frac{g_k(X_t,F_t)}{\pi_k(X_t,F_t)}+\lambda\pi_k(X_t,F_t)\middle|k_t=k\right]-\lambda\rho.
\]
By linearity of expectation, minimizing $\mathcal L$ over functions $\pi_k(\cdot)\in[\pi_{\min},1]$ reduces to minimizing
the integrand pointwise. Fix $(x,f)$ and write $g=g_k(x,f)$. We seek
\[
\min_{\rev{u}\in[\pi_{\min},1]}\left\{\frac{g}{\rev{u}}+\lambda\rev{u}\right\}.
\]
For \rev{$u>0$}, the objective is convex since \rev{$2g/u^3\ge0$}. The unconstrained minimizer satisfies
\[
\rev{-\frac{g}{u^2}+\lambda=0\quad\Longrightarrow\quad u^\circ(x,f)=\sqrt{\frac{g_k(x,f)}{\lambda}}.}
\]
Projecting onto the feasible interval $[\pi_{\min},1]$ gives
\[
\pi_k^\star(x,f)=\clip\!\left(\sqrt{\frac{g_k(x,f)}{\lambda}},\pi_{\min},1\right).
\]
The normalizing constant $\lambda>0$ is chosen so that the budget binds
$\E[\pi_k^\star(X_t,F_t)\mid k_t=k]=\rho$. In
particular, before clipping, $\pi_k^\star(x,f)\propto\sqrt{g_k(x,f)}$\rev{, which gives the estimator-oriented rule.}

\begin{revblock}
For the stated interior case, the mean of the clipped map is continuous in
$\lambda>0$ by dominated convergence. Its limits as $\lambda\downarrow0$
and $\lambda\uparrow\infty$ are $1$ and $\pi_{\min}$, respectively, so a
normalizer exists. If $g_k$ can vanish, let $s_k=\Pbb(g_k(X,F)>0\mid k)$.
For $\rho<\pi_{\min}+(1-\pi_{\min})s_k$, the same argument supplies a
positive normalizer. At or above that value, set the audit probability to
one on $\{g_k>0\}$ and allocate the remaining required rate on $\{g_k=0\}$. These zero-moment states contribute zero to $J$. The endpoint rates
$\rho=\pi_{\min}$ and $\rho=1$ use the corresponding constant maps.
If $g_k=0$ almost surely, every feasible policy has $J=0$.
\end{revblock}

Next, we give a simple two-segment construction showing that ignoring heterogeneity in $g_k(x,f)$
can incur a large variance penalty.

Let the conditional second moment $g_k(X_t,F_t)$ take two values\rev{,}
\[
g_k(X_t,F_t)=
\begin{cases}
G&\text{with probability }\rev{q},\\
\varepsilon&\text{with probability }\rev{1-q},
\end{cases}
\qquad\rev{0<\varepsilon<G\le1.}
\]
\rev{These moments are attained by taking $F=0$ and $Y=\sqrt{g_k(X,F)}$ in each observable segment, so both scores remain in $[0,1]$.}
Consider the feasible clipped policy
\[
\pi_{\mathrm{seg}}(X_t,F_t)=
\begin{cases}
1&\text{on the }g=G\text{ segment},\\
\pi_{\min}&\text{on the }g=\varepsilon\text{ segment},
\end{cases}
\]
and choose \rev{$q\coloneqq(\rho-\pi_{\min})/(1-\pi_{\min})$} so that
$\E[\pi_{\mathrm{seg}}]=\rho$. \rev{If $G/\varepsilon\ge\pi_{\min}^{-2}$, choose $\lambda\in[\varepsilon/\pi_{\min}^2,G]$. The clipped Neyman rule then assigns probability one to the $G$ segment and probability $\pi_{\min}$ to the $\varepsilon$ segment, giving $\pi_{\mathrm{seg}}$.}

The objective under uniform auditing $\pi_k\equiv\rho$ is
\[
J(\pi_k\equiv\rho)=\E\!\left[\frac{g_k(X_t,F_t)}{\rho}\right]
=\rev{\frac{qG+(1-q)\varepsilon}{\rho}}.
\]
Under $\pi_{\mathrm{seg}}$,
\[
J(\pi_{\mathrm{seg}})=\E\!\left[\frac{g_k(X_t,F_t)}{\pi_{\mathrm{seg}}(X_t,F_t)}\right]
=\rev{qG+(1-q)\frac{\varepsilon}{\pi_{\min}}}.
\]
\rev{Keeping $G\in(0,1]$ and $\pi_{\min}<\rho<1$ fixed and letting
$\varepsilon\downarrow0$, we have}
\[
\frac{J(\pi_k\equiv\rho)}{J(\pi_{\mathrm{seg}})}\longrightarrow\frac{1}{\rho}.
\]
Thus, when $\rho$ is small, a uniform audit policy can be worse than a policy guided by residual
\rev{second moments} (clipped Neyman) by a large factor (on the order of $1/\rho$). \rev{This is a comparison of the policy-dependent second-moment objective $J$. The identification allocation is determined by the information program.}

\begin{revblock}
\subsection{Policy classes and equality conditions}
The preceding calculation concerns estimation for one arm. We now use the singleton restriction of \eqref{eq:bundle-characteristic-cost} to compare audit rules for identifying the best configuration. This comparison makes precise the distinction following the dual program in Section~\ref{sec:theory_lower_bound}. An audit is valuable for identification when its outcome helps distinguish models with different best configurations, which do not need to be the states with the largest residual second moments.

Let $S$ denote the finite pre-audit state when one configuration is evaluated. The unrestricted class $\Pi_{\rm cont}$ allows a separate audit probability in $[p,1]$ for each arm and state. The constant class $\Pi_{\rm ex}$ uses one probability per arm, regardless of the current state. The score-based class $\Pi_{\rm AF}(\mathsf R)$ uses $\pi_{ks}=\psi_k(\mathsf R_k(s))$, where $\psi_k$ is nondecreasing and valued in $[p,1]$. Thus states with larger scores cannot receive smaller audit probabilities. The score $\mathsf R_k(s)$ is fixed from arm $k$ alone, without using competing arms or different-best alternatives. Constants belong to all three classes. The example below uses $g_k(s)=\E[(Y-F)^2\mid k,S=s]$ as the score. With positive state-dependent reference-acquisition costs, a score can instead be specified as $g_k(s)/c_{ks}^Y$.

For each class, minimize \eqref{eq:bundle-characteristic-cost} over singleton allocations $(q,x)$ satisfying its restrictions, with $x_{ks}=q_k\pi_{ks}$. Assume the finite-model identifiability condition holds for this singleton action set. This gives $C_{\rm cont}^\star$, $C_{\rm ex}^\star$ and $C_{\rm AF(\mathsf R)}^\star$, while holding alternatives, costs, state laws and the audit floor fixed. Let $\mathcal O$ be the set of unrestricted optimizers. For a dual optimum $\eta^\star$, put $H_{ks}=\sum_a\eta_a^\star h_{ks}^a$, where $h_{ks}^a$ is the conditional outcome KL divergence against alternative $a$ in \eqref{eq:bundle-kl-terms}. Thus $H_{ks}$ weights the information obtained from an audit according to the identification constraints. Comparing it with $c_{ks}^Y$ determines whether that state should receive more audits.

\begin{proposition}\label{prop:r1-contingency-value}
For the finite-state program,
$C_{\rm cont}^\star\le C_{\rm AF(\mathsf R)}^\star\le C_{\rm ex}^\star$.
A restricted class has the full value if and only if its feasible
allocations intersect $\mathcal O$. For an arm with $q_k>0$,
an optimum has $\pi_{ks}=1$ when $H_{ks}>c_{ks}^Y$ and $\pi_{ks}=p$
when $H_{ks}<c_{ks}^Y$. At equality the chosen probabilities must satisfy
the primal constraints. In particular, if $p<1$, the full optimizer is
unique, and two positive-mass states of an arm with $q_k>0$ have
opposite strict bound choices, then
$C_{\rm cont}^\star<C_{\rm ex}^\star$. At $p=1$ all three values coincide.
\end{proposition}
\begin{proof}
Since every constant map is score-based and every score-based map is state-dependent, their feasible allocation sets are nested in the stated order. Minimizing the same objective over these sets gives the inequalities. Constant rules impose $x_{ks}=x_{ks'}$ for every pair of states of arm $k$. Score-based rules impose $x_{ks}\le x_{ks'}$ whenever $\mathsf R_k(s)\le\mathsf R_k(s')$. In particular, equal scores require equal probabilities. These are linear constraints. Each restricted problem is feasible under full auditing by the finite-model identifiability condition, and its nonnegative objective has an attained finite minimum. If its value equals the unrestricted value, a restricted minimizer belongs to $\mathcal O$. Conversely, an element of $\mathcal O$ satisfying the restriction attains both values.

For the threshold statement, attach the dual multipliers $\eta_a^\star$ to the information constraints. The coefficient of $x_{ks}$ in the Lagrangian is $\mu_{\nu,k}(s)(c_{ks}^Y-H_{ks})$. By strong duality, a primal optimum minimizes this Lagrangian over $pq_k\le x_{ks}\le q_k$. On a positive-mass state with $q_k>0$, a negative coefficient requires $x_{ks}=q_k$, and a positive coefficient requires $x_{ks}=pq_k$. A zero coefficient leaves the probability to be determined by primal feasibility. These are the stated audit probabilities.

Under uniqueness and opposite strict choices, the only unrestricted optimum has a nonconstant audit map on a sampled arm. It therefore cannot satisfy the constant-rule constraints. The intersection criterion just proved gives strict inequality. At $p=1$, all audit probabilities equal one, so the three feasible sets and their values coincide.
\halmos
\end{proof}

A rule independent of the proxy can still depend on the current instance.
Let $\Pi_X$ contain maps of the form $\pi_k(X)$, with the same floor and
costs. Then $\Pi_{\rm ex}\subseteq\Pi_X\subseteq\Pi_{\rm cont}$, and
\[
 C_{\rm cont}^\star\le C_X^\star\le C_{\rm ex}^\star.
\]
The class $\Pi_X$ equates audit probabilities at states sharing $X$. These equalities are linear constraints on $x_{ks}$, and full auditing remains feasible. Its minimum therefore equals the unrestricted value exactly when a minimizer belongs to $\mathcal O$. Equivalently, observing proxies leaves optimal cost unchanged when some unrestricted optimizer depends only on $X$. If no such optimizer exists, allowing the rule to use $F$ strictly reduces cost. Using this reduction in an allocation rule requires the state-specific information and costs.

The estimator counterpart depends on which observables the rule uses.
For a fixed rule $\pi_k(X)$ and a diverging arm count, the unweighted
mean of audited outcomes converges almost surely to
\[
 \frac{\E[\pi_k(X)Y\mid k]}{\E[\pi_k(X)\mid k]}.
\]
This equals $\theta_k$ for a constant arm-specific audit rate. Dependence
on $X$ can change the limit even when the audit rule ignores $F$. For
example, if $\Pbb(X=0)=\Pbb(X=1)=1/2$, $Y=X$, and
$\pi_k(0)=p<1$, $\pi_k(1)=1$, the limit is $1/(1+p)$ and
$\theta_k=1/2$. Logged propensity correction recovers the target mean
for both classes, since $\E[AY/\pi_k(X)\mid k]=\theta_k$, and the corrected score has the same expectation. These are fixed-count identities. The confidence sequence supplies the guarantee at adaptive stopping times.

\subsection{A finite example with strict cost differences}
\label{subsec:finite-separating-instance}

This example gives a strict comparison between the two objectives discussed above. It supplements the audit-timing comparison in Theorem~\ref{thm:bundle-public-good} by restricting audits to probabilities that increase with residual second moments. We specify a finite model family and certify its optimal costs directly from feasible allocations and dual weights. The calculation shows why the estimator-oriented rule in Section~6.2 does not need to minimize the characteristic cost in Section~\ref{sec:theory_lower_bound}.

Consider \(\{\nu^0,\nu^1,\nu^2\}\), sharing two equally likely states per arm. Under \(\nu^0\), verified outcomes are Bernoulli with probabilities \(0.7\) for arm 1 and \(0.2\) for arm 2 in both states. Alternatives leave arm 1 unchanged. Under \(\nu^1\) and \(\nu^2\), arm-2 success probabilities are
\[
\begin{aligned}
\bigl(r_{1,0},r_{1,1}\bigr)&=(0.955583522015,\ 0.489186109759),\\
\bigl(r_{2,0},r_{2,1}\bigr)&=(0.590989164409,\ 0.928053798128).
\end{aligned}
\]
Each value rounds to 12 decimals the unique root in the stated interval of
\[
\begin{aligned}
\kl(\operatorname{Ber}(0.2)\Vert\operatorname{Ber}(r_{1,0}))&=2.00,
&r_{1,0}&\in(0.95,0.96),\\
\kl(\operatorname{Ber}(0.2)\Vert\operatorname{Ber}(r_{1,1}))&=0.18,
&r_{1,1}&\in(0.48,0.50),\\
\kl(\operatorname{Ber}(0.2)\Vert\operatorname{Ber}(r_{2,0}))&=0.32,
&r_{2,0}&\in(0.59,0.60),\\
\kl(\operatorname{Ber}(0.2)\Vert\operatorname{Ber}(r_{2,1}))&=1.62,
&r_{2,1}&\in(0.92,0.94).
\end{aligned}
\]
Thus all outcomes lie in \([0,1]\), while arm 2 has mean approximately
\(0.722384815887\) under \(\nu^1\) and \(0.759521481269\) under \(\nu^2\),
so both alternatives change the unique best arm from arm 1 to arm 2. The
state-law KL terms are zero, and the conditional KL table is exactly
\begin{equation}
\begin{array}{c|cc}
 &s=0&s=1\\ \hline
a=1&2.00&0.18\\
a=2&0.32&1.62
\end{array}
\label{eq:r1-kl-table}
\end{equation}
with configuration-evaluation cost $0.1$, reference-acquisition cost $1$, and floor $p=0.1$. The exact roots
above give finite, bounded log-likelihood ratios, and solving the finite LP
gives the state-dependent policy $\pi^\star=(6/7,1)$ and characteristic cost
$240/221\approx1.08597$. The optimum among constant audit rules has cost
$110/97\approx1.13402$. Thus conditioning on the state saves approximately
$4.24\%$ relative to this constant-audit optimum.
Take the observed proxy $F=1-S$. The specified Bernoulli outcome laws and
common state distribution define a joint binary $(Y,F)$ model in which
auditing follows the observation of $F$. Under the true model,
the specified score for arm 2 is the conditional second moment of its residual,
$g_2(0)=\mathbb{E}[(Y-F)^2\mid S=0]=4/5$ and
$g_2(1)=\mathbb{E}[(Y-F)^2\mid S=1]=1/5$. Hence every nondecreasing
$g_2$-score map satisfies $\pi_0\geq\pi_1$, while
$\pi^\star_2=(6/7,1)$ has $\pi_0<\pi_1$. For an allocation using this score,
write $q=q_2$, $x_s=q\pi_s$, $c(q,x)=q/10+(x_0+x_1)/2$, and
$I_2(x)=(16x_0+81x_1)/100$. Then
\[
c(q,x)-\frac{110}{97}I_2(x)
=\frac{q-x_0}{10}+\frac{203}{485}(x_0-x_1)\geq0.
\]
Full auditing at $q=x_0=x_1=100/97$ attains the bound, giving
\[
C_{\mathrm{cont}}^\star=240/221
 < 110/97=C_{\mathrm{AF}(g)}^\star=C_{\mathrm{ex}}^\star.
\]
For the same instance, the floor sensitivity check gives state-dependent and constant-audit
costs $(1.08597,1.13402)$ at each
$p\in\{0.01,0.05,0.10,0.25,0.50\}$, while the two costs coincide at
$p=1$. Thus the floor restricts feasible audit rules, without a universal $1/p$ multiplier on characteristic cost.
Scaling all four conditional KL entries by $1.25$ and $1.50$ gives alternative arm-2 means $(0.75105,0.79686)$ and $(0.77288,0.82520)$, respectively. The state-dependent and constant-audit costs become $(0.86878,0.90722)$ and $(0.72398,0.75601)$, while the relative saving remains $4.24\%$ in this calibrated family. This sensitivity changes the information scale, not the policy-class ordering.
The dual weights are
\[
 \eta^\star=\left(\frac{515}{1326},\frac{925}{1326}\right).
\]
Arm 1 provides no information about either alternative, so the optimum has
$q_1=0$. The unrestricted value is certified by the displayed dual weights.
With $I_1(x)=x_0+9x_1/100$,
\[
c(q,x)-\eta_1^\star I_1(x)-\eta_2^\star I_2(x)
=\frac{q-x_1}{10}\ge0.
\]
The feasible point $q=x_1=700/663$, $x_0=200/221$ has
$I_1=I_2=1$ and attains equality, giving cost $240/221$.
\end{revblock}

\section{\rev{Finite-Model Selection with Output-Specific Audits}}
\label{sec:private-reference-selection}
\rev{This section specializes the joint information and cost analysis in Section~\ref{sec:theory_lower_bound} to audits that verify one configuration output. We derive the information available from proxies and acquired references, and then relate the resulting lower bound to Algorithm~\ref{alg:joint-bundle}. This gives the finite-model benchmark for output-specific audits. PP-LUCB instead uses the bounded-outcome confidence sequence without specifying a finite model family, so its guarantee is the correctness result in Theorem~\ref{thm:delta-correct}.}
\subsection{\rev{Joint allocation when evaluating one configuration at a time}}
\label{subsec:jpta-finite-ec}
\rev{We first express the expected information and cost of an adaptive procedure through its configuration counts and average state-dependent audit probabilities. These identities allow a direct comparison with the singleton restriction of the main paper's allocation problem.}
Fix the true instance $\nu$ and an alternative $\nu'\in\operatorname{Alt}(\nu)$
such that $k^\star(\nu')\ne k^\star(\nu)$. Consider any $\delta$-correct
algorithm with stopping time $\tau$ (with respect to the natural filtration
defined below) and finite expected cost.

\rev{Here $\operatorname{Alt}(\nu)=\mathcal A_\nu$ in the specified finite family. Each round chooses one configuration before a fresh instance and scores are observed. The common instance law $P(X_t)$ is $\mathcal D$. Finite common support bounds the absolute round log-likelihood ratio by $L$. Positive expected configuration-evaluation cost gives
$\E_\nu[\tau]\le\E_\nu[\operatorname{Cost}(\tau)]/\underline c_F<\infty$,
so log-likelihood increments are absolutely summable in expectation through $\tau$.}

Let the per-pull observation be $O_t=(X_t,F_t,A_t,\rev{A_tY_t})$, and the
filtration $\rev{\mathcal H}_t=\sigma\bigl((k_s,O_s,\pi_s)_{s\le t}\bigr)$.
Here $k_t$ is $\rev{\mathcal H}_{t-1}$-measurable, and
$\pi_t=\Pbb(A_t=1\mid\rev{\mathcal H}_{t-1},k_t,X_t,F_t)$ is measurable
with respect to $\sigma(\rev{\mathcal H}_{t-1},k_t,X_t,F_t)$. By
Assumption~\ref{ass:audit}, $A_t$ cannot depend on the unobserved $Y_t$ given
$(\rev{\mathcal H}_{t-1},k_t,X_t,F_t)$.

Under Assumption~\ref{ass:audit}, conditional on
$(\rev{\mathcal H}_{t-1},k_t)$, the conditional distribution of $O_t$ factors as
\begin{align*}
P_\nu(O_t\mid\rev{\mathcal H}_{t-1},k_t)
={}&P(X_t)P_\nu(F_t\mid X_t,k_t)\,
P(A_t\mid\rev{\mathcal H}_{t-1},k_t,X_t,F_t)\,
P_\nu(Y_t\mid X_t,F_t,k_t)^{A_t}.
\end{align*}
\rev{The display uses densities on the common observation support. When
$A_t=0$, the recorded last coordinate equals zero with probability one, and
the last factor is defined to be one. When $A_t=1$, that coordinate is
$Y_t$ and the last factor is its conditional density.}

The audit mechanism $P(A_t\mid\cdot)$ is part of the algorithm, so it is
identical under $\nu$ and $\nu'$. Therefore the one-step log-likelihood ratio is
\begin{align}
\ell_t(\nu,\nu')
&:=\log\frac{dP_\nu(O_t\mid\rev{\mathcal H}_{t-1},k_t)}
{dP_{\nu'}(O_t\mid\rev{\mathcal H}_{t-1},k_t)}\notag\\
&=\underbrace{\log\frac{dP_\nu(F_t\mid X_t,k_t)}
{dP_{\nu'}(F_t\mid X_t,k_t)}}_{\displaystyle =:\ell_t^{(F)}(\nu,\nu')}
+A_t\underbrace{\log\frac{dP_\nu(Y_t\mid X_t,F_t,k_t)}
{dP_{\nu'}(Y_t\mid X_t,F_t,k_t)}}_{\displaystyle =:\ell_t^{(Y\mid X,F)}(\nu,\nu')}.
\label{eq:jpta-one-step-ratio}
\end{align}
Let
\[
L_\tau(\nu,\nu'):=\sum_{t=1}^{\tau}\ell_t(\nu,\nu')
\]
denote the cumulative log-likelihood ratio.

By the KL chain rule (see, e.g., \citealp{kaufmann2016complexity}),
\begin{align}
\E_\nu[L_\tau(\nu,\nu')]
=\E_\nu\!\left[\sum_{t=1}^{\tau}
\operatorname{KL}\!\left(
P_\nu(O_t\mid\rev{\mathcal H}_{t-1},k_t)
\,\middle\|\,
P_{\nu'}(O_t\mid\rev{\mathcal H}_{t-1},k_t)
\right)\right],
\label{eq:singleton-stopped-kl-chain}
\end{align}
whenever the displayed conditional KL terms are integrable.
\rev{Under the conditions stated above, the identity follows by writing each
stopped sum as a sum of increments multiplied by
$\mathbf 1\{\tau\ge t\}$. This indicator is measurable before round $t$. The absolute expected sum is bounded by $L\E_\nu[\tau]$, so conditional
expectation and summation can be interchanged.}
Fix $t$. Conditional on $(\rev{\mathcal H}_{t-1},k_t,X_t,F_t)$, we have
$A_t\sim\operatorname{Bernoulli}(\pi_t)$ and $A_t$ is independent of $Y_t$
under Assumption~\ref{ass:audit}. Since $A_t$'s conditional distribution is
the same under $\nu$ and $\nu'$, the conditional KL splits into a proxy
part and an audited outcome part revealed only when $A_t=1$\rev{,}
\begin{align}
&\operatorname{KL}\!\left(
P_\nu(O_t\mid\rev{\mathcal H}_{t-1},k_t)
\,\middle\|\,
P_{\nu'}(O_t\mid\rev{\mathcal H}_{t-1},k_t)\right)\notag\\
=&\E_\nu\!\left[\left.\begin{aligned}
&\operatorname{KL}\!\left(P_\nu(F_t\mid X_t,k_t)
\,\middle\|\,P_{\nu'}(F_t\mid X_t,k_t)\right)\\
&+\pi_t\operatorname{KL}\!\left(P_\nu(Y_t\mid X_t,F_t,k_t)
\,\middle\|\,P_{\nu'}(Y_t\mid X_t,F_t,k_t)\right)
\end{aligned}\,\right|\,\rev{\mathcal H}_{t-1},k_t\right].
\label{eq:singleton-conditional-kl-split}
\end{align}

\rev{The audit probability may depend on the full history. For the lower bound, we summarize these probabilities by a state-dependent average that preserves the expected information and cost up to stopping.} Fix arm $k$.
Define two finite measures on $\mathcal X\times[0,1]$\rev{.} \rev{For} any measurable
set $B\subseteq\mathcal X\times[0,1]$,
\begin{align*}
\Lambda_k(B)&:=\E_\nu\!\left[\sum_{t=1}^{\tau}
\mathbf 1\{k_t=k\}\mathbf 1\{(X_t,F_t)\in B\}\right],\\
\Gamma_k(B)&:=\E_\nu\!\left[\sum_{t=1}^{\tau}
\mathbf 1\{k_t=k\}\pi_t\mathbf 1\{(X_t,F_t)\in B\}\right].
\end{align*}
\rev{Here $\Lambda_k(B)$ is the expected number of evaluations whose observables lie in $B$, and $\Gamma_k(B)$ is the expected number of requested audits among those evaluations.}
Since $\rev{p\le}\pi_t\le1$, we have
$\rev{p\Lambda_k(B)\le}\Gamma_k(B)\le\Lambda_k(B)$, so
$\Gamma_k\ll\Lambda_k$. By the Radon--Nikodym Theorem, there exists a
measurable $\bar\pi_k:\mathcal X\times[0,1]\to[\rev{p},1]$ such that
for all measurable $B$,
\[
\Gamma_k(B)=\int_B\bar\pi_k(x,f)\,d\Lambda_k(x,f).
\]
Equivalently, for any integrable measurable $h(x,f)$,
\begin{align}
\E_\nu\!\left[\sum_{t=1}^{\tau}\mathbf 1\{k_t=k\}\pi_t h(X_t,F_t)\right]
&=\E_\nu\!\left[\sum_{t=1}^{\tau}\mathbf 1\{k_t=k\}
\bar\pi_k(X_t,F_t)h(X_t,F_t)\right].
\label{eq:singleton-audit-average}
\end{align}
Moreover, by Assumption~\ref{ass:audit} and the fact that
$\pi_t\ge\rev{p}$ whenever auditing is eligible, the averaged kernel
satisfies $\bar\pi_k(x,f)\in[\rev{p},1]$ on the support of $(X,F)\mid k$.

\begin{revblock}
The fresh-state law makes this support statement explicit. Since
$\mathbf 1\{t\le\tau,k_t=k\}$ is measurable before the current instance
and proxy are drawn, Tonelli's theorem and conditional expectation give
\begin{align*}
\Lambda_k(B)
&=\sum_{t\ge1}\E_\nu\!\left[
\mathbf 1\{t\le\tau,k_t=k\}
\Pbb_\nu((X_t,F_t)\in B\mid\mathcal H_{t-1},k_t=k)\right]\\
&=\E_\nu[N_k(\tau)]\,P_\nu((X,F)\in B\mid k).
\end{align*}
When $\E_\nu N_k(\tau)>0$, the two measures have the same null sets, and
the Radon-Nikodym derivative can be chosen in $[p,1]$ throughout the
arm's support. When $\E_\nu [N_k(\tau)]=0$, set $\bar\pi_k\equiv p$. That arm contributes zero to the following identities.
\end{revblock}

Combining \eqref{eq:singleton-stopped-kl-chain} and
\eqref{eq:singleton-conditional-kl-split} and using
\eqref{eq:singleton-audit-average}, we have
\begin{equation}
\E_\nu L_\tau(\nu,\nu')
=\sum_{k=1}^K\E_\nu[N_k(\tau)]\cdot I_k(\nu,\nu';\bar\pi_k),
\label{eq:singleton-stopped-information}
\end{equation}
where $N_k(\tau):=\sum_{t=1}^{\tau}\mathbf 1\{k_t=k\}$ and
\begin{align}
I_k(\nu,\nu';\pi_k)
:={}&\E_\nu\!\left[\left.
\operatorname{KL}\!\left(P_\nu(F\mid X,k)\,
\middle\|\,P_{\nu'}(F\mid X,k)\right)\right.\right.\notag\\[-2pt]
&\left.\left.{}+\pi_k(X,F)\operatorname{KL}\!\left(
P_\nu(Y\mid X,F,k)\,\middle\|\,P_{\nu'}(Y\mid X,F,k)\right)
\,\right|\,k\right].
\label{eq:singleton-information-rate}
\end{align}
\rev{The first term is available on every evaluation. The second is obtained only when a reference is acquired, so the audit probability directly changes the information obtained per evaluation.}

Similarly, for the cost, \rev{using configuration-evaluation cost $c_F$ and reference-acquisition cost $c_Y$ in the configuration-first model,}
\[
\E_\nu[\operatorname{Cost}(\tau)]
=c_F\E_\nu[\tau]+c_Y\E_\nu\!\left[\sum_{t=1}^{\tau}A_t\right].
\]
Since $\E_\nu[A_t\mid\rev{\mathcal H}_{t-1},k_t,X_t,F_t]=\pi_t$ and
by the averaging identity \eqref{eq:singleton-audit-average},
\[
\E_\nu\!\left[\sum_{t=1}^{\tau}A_t\right]
=\sum_{k=1}^K\E_\nu[N_k(\tau)]\cdot
\E_\nu[\bar\pi_k(X,F)\mid k].
\]
Then
\begin{equation}
\E_\nu[\operatorname{Cost}(\tau)]
=\sum_{k=1}^K\E_\nu[N_k(\tau)]\cdot
\bigl(c_F+c_Y\E_\nu[\bar\pi_k(X,F)\mid k]\bigr).
\label{eq:singleton-stopped-cost}
\end{equation}

By the standard change-of-measure arguments for BAI
\citep{kaufmann2016complexity}, for any $\nu'\in\operatorname{Alt}(\nu)$,
\begin{equation}
\E_\nu L_\tau(\nu,\nu')\ge
\rev{\kappa_\delta:=\kl(1-\delta,\delta)}.
\label{eq:singleton-binary-information}
\end{equation}
\rev{Here $0<\delta<1/2$ and correctness holds uniformly over the declared
model family. Apply binary data processing to the event that the recommendation
equals $k^\star(\nu)$\rev{.} \rev{Its} probabilities are at least $1-\delta$ under
$\nu$ and at most $\delta$ under $\nu'$.}
Combining \eqref{eq:singleton-stopped-information}-\eqref{eq:singleton-binary-information}
and dividing by $\E_\nu[\operatorname{Cost}(\tau)]$ using
\eqref{eq:singleton-stopped-cost}, we have
\[
\frac{\rev{\kappa_\delta}}{\E_\nu[\operatorname{Cost}(\tau)]}
\le\frac{\sum_{k=1}^K w_k I_k(\nu,\nu';\bar\pi_k)}
{c_F+c_Y\sum_{k=1}^K w_k\E_\nu[\bar\pi_k(X,F)\mid k]},
\]
where $w_k:=\E_\nu[N_k(\tau)]/\E_\nu[\tau]$ and
$w\in\Delta_K$. Taking $\inf_{\nu'\in\operatorname{Alt}(\nu)}$ and then
$\sup_{w\in\Delta_K,\pi}$, \rev{the reciprocal of the optimal
information-to-cost ratio is the singleton characteristic cost. Thus
$\E_\nu[\operatorname{Cost}(\tau)]\ge
\kappa_\delta C_{\rm sing}^\star(\nu,\mathcal A_\nu)$. Since
$\kappa_\delta/\log(1/\delta)\to1$,} letting $\delta\to0$, we obtain
\[
\liminf_{\delta\to0}
\frac{\E_\nu[\operatorname{Cost}(\tau)]}{\log(1/\delta)}
\ge\rev{C_{\rm sing}^\star(\nu,\mathcal A_\nu)},
\]
where \rev{$C_{\rm sing}^\star$ is the singleton restriction of
\eqref{eq:bundle-characteristic-cost} for these costs. This gives the lower bound for one-configuration evaluations.}

\rev{To apply Algorithm~\ref{alg:joint-bundle}, take $\mathcal B=\{\{1\},\ldots,\{K\}\}$ and pre-audit state $(X,F)$. The state-law KL term then equals the expected proxy KL in \eqref{eq:singleton-information-rate}, because the instance law is common to the models. The conditional verified-vector KL becomes the scalar-outcome KL in the same equation. The cost terms are $c_Fq_k$ and $c_Y\sum_s\mu_{\nu,k}(s)x_{ks}$. Thus the likelihoods, information constraints and costs are exactly those of the joint allocation problem restricted to singleton bundles. Under the finite-support, identifiability and positive configuration-evaluation cost conditions of Theorem~\ref{thm:bundle-tracker} for this action set, that theorem gives uniform correctness, finite expected stopping and attainment of $C_{\rm sing}^\star$. When all models have the same best configuration, the procedure returns it at time zero as in the main paper.}

\begin{revblock}
\section{Service Evaluation and Delayed Verification}
\label{sec:service-evaluation-delays}
This section supports the discussion of review capacity in Section~\ref{sec:conclusion} of the main paper. The experiment on the review queue examines how pending references affect the evidence available for selection. It complements the validity results for delayed feedback in Section~\ref{sec:delays} by reporting costs and completion times under limited review capacity. We also give a separate example of scoring mathematical answers to clarify the distinction between a verified outcome and a judge score in Section~\ref{sec:model} of the main paper. The final subsection defines completion time when requests remain pending at stopping.
\end{revblock}

\begin{revblock}
\subsection{Experiment on the review queue}
\label{EC-sec:queue-service}

This additional experiment examines whether purchasing more references produces usable evidence sooner when review capacity is limited. It addresses the practical distinction between statistical validity and timely certification discussed in Section~\ref{sec:conclusion}. Using the four configurations in Section~\ref{ec:new-horizons}, we vary the audit rate and reviewer capacity under a common submission rule. The confidence sequence uses completed submission prefixes, so a pending reference cannot help certify a configuration. Keeping the submission order fixed makes the effect of review workload explicit.

The configuration means are $(.70,.65,.55,.50)$. On every submission of
configuration $k$, a fresh state $S\in\{0,1\}$ is equiprobable, the proxy is
$F=f_k+.1(2S-1)$ with $f=(.58,.80,.60,.68)$, and
$Y\sim\mathrm{Bernoulli}(\theta_k)$ is independent of $S$ and earlier
submissions. Configurations are submitted in the fixed cyclic order
$1,2,3,4$. Submission $t$ occurs at calendar time $t/r$, where the submission
rate is $r=.6$ or $1.1$. The workload offered to the auditors also depends on the audit rule.

We compare 40\% audit, 70\% audit, and full audit. Each policy purchases a
reference with its fixed probability $p=.4,.7$, or $1$, independently of the
current state and outcome. A purchased reference joins a common FIFO queue
with $m=1$ or $2$ identical auditors. Its service duration is $1+4(1-Y)$.
Requests start with the first available auditor, breaking ties by server index. Failures take five time units and successes one, making duration outcome dependent. Cyclic submissions give mean service demand $2.6rp$ auditor-time units per calendar-time unit. Configuration-evaluation and reference-acquisition costs are $c_F=1$ and $c_Y=10$, separate from calendar time.

Index the submissions of a given configuration by $n$. The corrected score
is $Z_n=F_n+A_n(Y_n-F_n)/p$, with mean $\theta_k$ and support
$[1-1/p,1/p]$. We fix the center $c=.5$, the envelope $B=1/p-.5$, and
\[
 \lambda_n=\frac{.04}{\sqrt{1+(n-1)/10000}},\qquad
 \phi(u)=-\log(1-u)-u.
\]
These quantities precede the current state and outcome, and $0<\lambda_nB\le .08<1$ for all three policies. A resolved prefix contains all corrected scores from the first submission onward. Unaudited scores are immediate. Audited scores wait for their references and any earlier pending submissions of the same configuration. For prefix length $n\ge1$, define
\begin{align*}
 W_n&=\sum_{j=1}^n\lambda_j,&
 \widehat\theta_{k,n}&=\frac{\sum_{j=1}^n\lambda_jZ_j}{W_n},\\
 v_n&=\sum_{j=1}^n\phi(\lambda_jB)\frac{(Z_j-.5)^2}{B^2},&
 w_{k,n}&=\frac{\log(2K/\delta)+v_n}{W_n}.
\end{align*}
The interval is $[\widehat\theta_{k,n}-w_{k,n},
\widehat\theta_{k,n}+w_{k,n}]\cap[0,1]$, with $[0,1]$ at $n=0$.
We use $K=4$ and $\delta=.05$. The tuning is common to all cells and is
fixed before simulation. Since $\sum_{j\le n}\lambda_j$ grows at order
$\sqrt n$ and $\sum_{j\le n}\lambda_j^2$ at order $\log n$, the bounded-score
half-width tends to zero as the resolved prefix grows.

For completeness, this is the empirical-Bernstein betting construction
under the submission-order filtration. For $|x|\le u<1$,
$\log(1+x)\ge x-\phi(u)x^2/u^2$. Apply it with
$x=\pm\lambda_n(Z_n-.5)$ and $u=\lambda_nB$. Exponentiating gives
\[
 \exp\!\left\{\pm\lambda_n(Z_n-\theta_k)
       -\phi(\lambda_nB)(Z_n-.5)^2/B^2\right\}
 \le e^{\pm\lambda_n(.5-\theta_k)}
       \{1\pm\lambda_n(Z_n-.5)\}.
\]
Conditional expectation of the right side is at most one, because
$e^{-a}(1+a)\le1$ for $a=\pm\lambda_n(\theta_k-.5)$.
The products over $n$ are therefore nonnegative supermartingales starting
at one. Ville's inequality and a union bound over two signs and four
configurations give simultaneous coverage of every submission prefix with
probability at least $1-\delta$.

Each decision uses the longest resolved prefix per configuration. Simultaneous coverage permits prefix lengths and service durations to depend on $Y$, while score tuning remains predictable in submission order.

After processing returns at each submission, let $b$ maximize the lower endpoint, breaking ties by index. Stop when $L_b\ge\max_{j\ne b}U_j-\epsilon$, with $\epsilon=.05$, implying $\theta_b\ge\max_j\theta_j-\epsilon$ on simultaneous coverage. Paths end at stopping or 20,000 submissions. Charge all evaluations and audits, including pending requests. The completion tail measures time from this endpoint until the final return.

Each condition uses 200 independent paths. Corresponding paths across conditions share their state, outcome and audit-uniform streams.  Table~\ref{tab:focused-service} gives mean costs with standard errors in parentheses, pending counts and completion tails, including all paths capped at 20,000 submissions.

\begin{table}[t]
\color{MSRevisionBlue}
\centering\small
\caption{Selection, total costs and completion times in the review queue.}
\label{tab:focused-service}
\begin{tabular}{lllrrrr}\toprule
Rate & Auditors & Rule & Certified & Cost (SE) & Pending & Tail \\ \midrule
0.6 & 1 & 40\% audit & 183/200 & 72543 (1082) & 1.3 & 3.5 \\
0.6 & 1 & 70\% audit & 199/200 & 111862 (1274) & 832.2 & 2161.1 \\
0.6 & 1 & Full audit & 133/200 & 201976 (1311) & 6593.3 & 17133.8 \\
0.6 & 2 & 40\% audit & 183/200 & 72536 (1082) & 0.7 & 1.5 \\
0.6 & 2 & 70\% audit & 200/200 & 102363 (1168) & 1.3 & 2.4 \\
0.6 & 2 & Full audit & 200/200 & 133946 (1277) & 2.2 & 3.6 \\
1.1 & 1 & 40\% audit & 165/200 & 81265 (1055) & 821.2 & 2135.3 \\
1.1 & 1 & 70\% audit & 16/200 & 159295 (210) & 6975.8 & 18129.2 \\
1.1 & 1 & Full audit & 0/200 & 220000 (0) & 13005.2 & 33822.0 \\
1.1 & 2 & 40\% audit & 183/200 & 72557 (1082) & 1.7 & 2.7 \\
1.1 & 2 & 70\% audit & 200/200 & 103281 (1180) & 77.0 & 100.5 \\
1.1 & 2 & Full audit & 169/200 & 189258 (1544) & 5176.5 & 6725.6 \\
\bottomrule\end{tabular}
\end{table}

Across the 2,400 paths in Table~\ref{tab:focused-service}, 1,831 stop and all select an $\epsilon$-good configuration. Coverage at all
submission decisions is $.995$ for 40\% audit and $1$ for the other two
rules in each rate--capacity condition. Because stopping fractions differ, the table includes both counts and all capped means.

These results measure total costs and delays in obtaining usable evidence. At rate $.6$ with two auditors, both 70\% audit and full
audit certify all 200 paths. Their mean total costs are 102,363 and
133,946. At rate $1.1$ with one auditor, 40\% audit certifies 165 paths,
while full audit reaches the submission cap on every path with an average
of 13,005 pending references. A second auditor raises full audit's
certification count to 169. The stopping counts answer whether enough usable evidence arrives within the submission budget, while pending counts and completion tails measure the resulting review burden. They show that the audit rate must be chosen with review capacity in mind. Purchasing more references can delay certification when it creates a backlog.
\end{revblock}

\subsection{\rev{Scoring rules for mathematical answers}}\label{app:convergence-details}

\rev{This example supports the definition of a bounded verified outcome in Section~\ref{sec:model}. We specify two ways to score a solver response when the correct answer is known. The first compares the response with that answer, while the second asks a judge to assign a score. Their difference explains why the target outcome must be defined separately from the judge score.}

\rev{The scoring rules apply to problems with a specified correct answer. The following question categories illustrate the kinds of reference information used.}
\begin{enumerate}[leftmargin=*,itemsep=2pt]
\item \textbf{Date calculations}\rev{.} Require calendar
arithmetic, e.g., ``Jan 15 is Monday. What day is Feb 1?''
\item \textbf{Complex arithmetic}\rev{.} Large multiplication or
exponentiation, e.g., ``What is $37 \times 89$?''
\item \textbf{Classic trick questions}\rev{.} Problems with
counterintuitive answers, e.g., the bat-and-ball problem from
\citet{kahneman2011thinking}.
\item \textbf{Function composition}\rev{.} Nested function
evaluation, algebraic simplification, or summation formulas.
\item \textbf{Combinatorics or probability}\rev{.} Counting and
probability calculations.
\end{enumerate}

\rev{A configuration consists of a solver model and its instructions, applied to a common set of questions.} For each evaluation instance, we first obtain the solver
response, then obtain a \rev{score} from a judge
model, \rev{and compare the response with the known correct answer.}
Each problem has a unique correct answer. We encode the verified outcome
as \(Y \in \{0,1\}\) (binary correctness) to satisfy
\textcolor{.}{Assumption~1}. \rev{The rule assigns $Y=1$ if and only if the}
correct answer appears in the model's response (case-insensitive, with
numeric tolerance \(\epsilon = 0.01\) for floating-point comparisons).
\rev{This binary outcome records agreement with the answer-matching rule. It does not evaluate the reasoning in the response.}

\rev{For the second scoring rule,} we use a separate judge model to score
the solver response on a \([0,1]\) scale. Specifically, the judge model
(\texttt{gpt-4.1-mini}) receives the question, the correct answer, and
the solver response, and returns a numeric score \(F \in [0,1]\). \rev{The prompt assigns scores 0, .5 and 1 to incorrect, partly correct and correct responses, respectively, and permits any number in this interval.} This judge score can differ from the
verified binary outcome, \rev{so the residual \(Y-F\) measures disagreement between the two scoring rules.}

\rev{For questions with numeric answers, the solver prompt takes the following form.}
\begin{lstlisting}[basicstyle=\ttfamily\small,breaklines=true,frame=single,caption={Solver
System Prompt}]
You are a math assistant. Answer with ONLY the numerical answer. No
explanation, no units, just the number.
\end{lstlisting}

\begin{lstlisting}[basicstyle=\ttfamily\small,breaklines=true,frame=single,caption={Solver
User Prompt}]
Question: {question}
\end{lstlisting}

The judge model (\texttt{gpt-4.1-mini}) evaluates solver responses on a
\([0,1]\) scale\rev{.}
\begin{lstlisting}[basicstyle=\ttfamily\small,breaklines=true,frame=single,caption={Judge
Evaluation Prompt}]
You are evaluating a math answer.

Question: {question}

Correct answer: {correct_answer}

Student response: {model_response}

Rate the response from 0 to 1:

- 1.0: Exactly correct

- 0.5: Partially correct or unclear

- 0.0: Wrong or no answer

Reply with ONLY a number between 0 and 1.
\end{lstlisting}

\rev{For a date question that asks for a weekday, the answer format instruction instead requests only the day of the week. In both cases, the required format follows the question, and the answer-matching rule uses its specified reference answer.}

\rev{The two rules can differ even after a correct answer is supplied. The answer-matching rule gives a binary outcome, while the judge prompt allows partial credit. Consequently, $Y-F$ records a difference between the stated scoring rules rather than an error measured against an independently collected human assessment. Since the judge already sees the reference, its score is also not the pre-audit proxy in our selective-auditing model. Experiments on acquisition cost and selection reliability are reported in Sections~\ref{ec:new-synthetic-experiments} and~\ref{sec:service-case-studies}.}

\begin{revblock}
\subsection{Completion-Time Accounting}
\label{subsec:delayed-verification-experiments}
This subsection defines the distinction between stopping and completion used in the experiment on the review queue in Section~\ref{EC-sec:queue-service}. It also connects that distinction to the delay model in Section~\ref{sec:delays}. A valid recommendation can be available while previously requested references remain pending. These requests no longer affect the stopping decision, but their review still has to be completed.
\paragraph{Completion-time accounting.}
In the round-based delay model, an episode is complete only after the stopping decision
and all references submitted before that decision have returned. Let $\tau$ be
the stopping submission index and $D_s$ the delay of request $s$. The calendar
completion time is
\[
  T_{\rm comp}=\max\!\left\{\tau,\;\max_{s\le\tau:A_s=1}(s+D_s)\right\}.
\]
If no audit has been requested, the inner maximum is omitted and $T_{\rm comp}=\tau$.
Configuration-evaluation and reference-acquisition costs are included in the stopped total
\(\sum_{s\le\tau}(c_F+c_YA_s)\), and $T_{\rm comp}-\tau$ defines the additional return time. Thus pending labels are not used as evidence, but they still require staff
time and must be completed.
\paragraph{Completion under bounded delay.}
If every submitted audit returns within $D_{\max}$ rounds, the definition gives
$0\le T_{\rm comp}-\tau\le D_{\max}$ pathwise. Under unbounded delays, its distribution
also depends on the number and ages of requests pending at the stopping time.
\end{revblock}

\end{document}